\documentclass[10pt,twocolumn,letterpaper]{article}

\usepackage[accsupp]{axessibility}  
\usepackage{iccv}
\usepackage{times}
\usepackage{epsfig}
\usepackage{graphicx}
\usepackage{amsmath}
\usepackage{amssymb}
\usepackage{microtype}
\usepackage{gensymb}
\usepackage{multirow}
\usepackage{booktabs}
\usepackage{bm}
\usepackage[caption=false,labelformat=empty,captionskip=1pt,farskip=0pt]{subfig}
\usepackage{makecell}
\usepackage{float}
\usepackage{caption}

\usepackage[pagebackref=true,breaklinks=true,letterpaper=true,colorlinks,bookmarks=false]{hyperref}

\iccvfinalcopy 

\def\iccvPaperID{****} 

\ificcvfinal\fi

\begin{document}

\newcommand{\ours}{FIPT}
\newcommand{\neilf}{NeILF~\cite{yao2022neilf}}
\newcommand{\fvp}{FVP~\cite{philip2021free}}
\newcommand{\milo}{MILO~\cite{yu2023milo}}
\newcommand{\lieccv}{Li22~\cite{li2022physically}}
\newcommand{\ipt}{IPT~\cite{azinovic2019inverse}}

\makeatletter
\newcommand{\printfnsymbol}[1]{%
        \textsuperscript{\@fnsymbol{#1}}%
}
\makeatother

\title{Factorized Inverse Path Tracing \\for Efficient and Accurate Material-Lighting Estimation}



\author{
Liwen Wu$^{1}$\thanks{Equal contribution}
\quad  
Rui Zhu$^{1}$\printfnsymbol{1}
\quad  
Mustafa B. Yaldiz$^1$ 
\quad  
Yinhao Zhu $^2$
\quad  
Hong Cai $^2$
\quad  
Janarbek Matai  $^2$
\\
Fatih Porikli $^2$
\quad  
Tzu-Mao Li $^1$
\quad  
Manmohan Chandraker $^1$
\quad  
Ravi Ramamoorthi $^1$
\\
$^1$UC San Diego \quad $^2$Qualcomm AI Research
\\
{\tt\small \{liw026,rzhu,myaldiz,tzli,mkchandraker,ravir\}@ucsd.edu}
\\
{\tt\small \{yinhaoz,hongcai,jmatai,fporikli\}@qti.qualcomm.com}
}

\maketitle

\ificcvfinal\thispagestyle{empty}\fi

\begin{abstract}
Inverse path tracing has recently been applied to joint material and lighting estimation, given geometry and multi-view HDR observations of an indoor scene.
However, it has two major limitations:
path tracing is expensive to compute,
and 
ambiguities exist between reflection and emission.
Our Factorized Inverse Path Tracing (FIPT) addresses these challenges by using a factored light transport formulation 
and finds emitters driven by rendering errors. 
Our algorithm enables accurate material and lighting optimization faster than previous work, and is more effective at resolving ambiguities.
The exhaustive experiments on synthetic scenes show that our method 
(1) outperforms state-of-the-art indoor inverse rendering and relighting methods particularly in the presence of complex illumination effects; 
(2) speeds up inverse path tracing optimization 
to less than an hour.  
We further demonstrate robustness to noisy inputs through material and lighting estimates that allow plausible relighting in a real scene.
The source code is available at: \url{https://github.com/lwwu2/fipt}
\end{abstract}

\section{Introduction}
\label{sec:intro}

\begin{figure}
    \centering
    \includegraphics[width=0.99\linewidth]{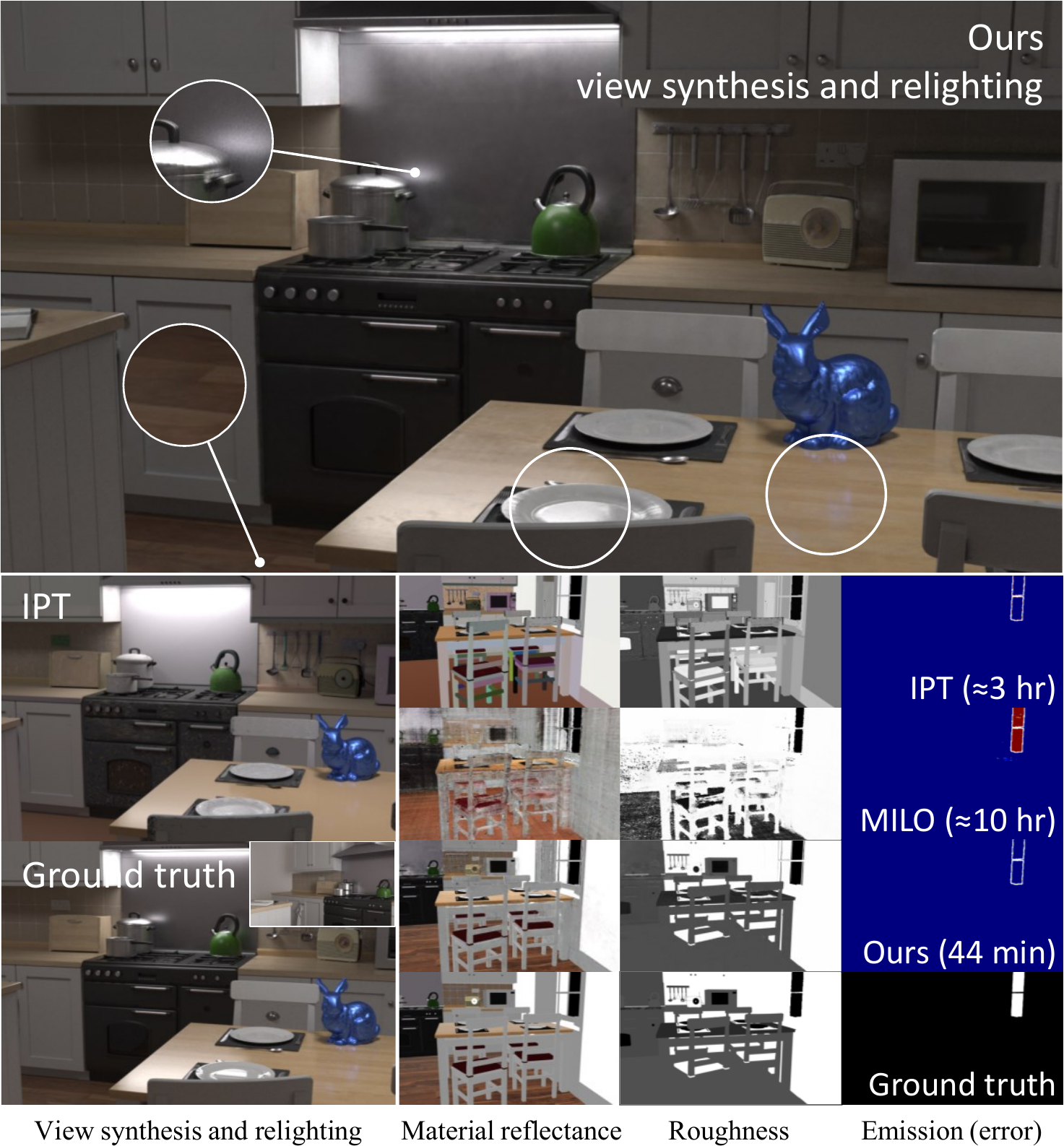}
    \caption{
    \textbf{Ours vs standard IPT.} 
        \ipt~takes a piecewise constant parameterization of material to reduce Monte Carlo variance and ambiguity for inverse rendering,
        losing fine spatial details as a result. 
        Directly extending it to complex material representation (\eg\milo) shows very slow convergence.
        In contrast,
        we propose Factorized Inverse Path Tracing (FIPT) to get rid of variance and reduce ambiguities,
        yielding efficient and high quality BRDF and emission (4th row),
        appealing relighting (1st row), and object insertion (the bunny on the table).
        The presented scene is synthetic with the inset showing the input (lower-left sub-figure). We further showcase results on real scenes in Fig.~\ref{fig:real-world} and ~\ref{fig:real-world-relight}.
    }
    \label{fig:teaser}
\end{figure}

We address the task of estimating the materials and lighting of an indoor scene based on image observations (Fig.~\ref{fig:teaser}). Recent work has shown that optimizing per-scene material and emission profiles through photometric loss and a differentiable renderer, with geometry reconstructed with the existing 3D reconstruction algorithms~\cite{schoenberger2016mvs,mildenhall2020nerf,Yu2022MonoSDF}, can lead to promising results~\cite{azinovic2019inverse,nimierdavid2021material,yu2023milo}. However, key challenges remain unsolved in these methods: (1) they require expensive Monte Carlo estimation for both the loss and derivative evaluations; (2) inherent ambiguity exists between material and lighting, and this ill-posed inverse problem hinders the optimization.
We present an alternative inverse rendering algorithm that outperforms the state-of-the-art in terms of both efficiency and accuracy.

Optimizing scene parameters with Monte Carlo differentiable rendering can suffer from high variance and lead to slow convergence. 
Inspired by classical 
irradiance caching literature~\cite{Ward:1988:RTS},
our key idea to address this challenge is to factorize the material term out of the rendering integral and bake the incoming radiance to significantly speed up inverse rendering.
Unlike prior work which also applies a similar factorization (\eg~\cite{philip2021free,li2022multi}) but does not consider view-dependent reflections, our method extends to general specular materials and both local and global illumination.

To address the ill-posed nature of joint optimization of material and lighting, we observe that by taking out the emission term in the rendering equation for the first bounce, only emissive surfaces will have high rendering loss. This observation allows us to design an effective way to detect emitters. 
We incorporate our emitter detection method into a full inverse rendering pipeline 
and independently estimate the emission after emitter detection.

Overall,
our method achieves fast convergence over the material-lighting estimation task thanks to our factorized light transport formulation and emitter extraction strategy (Fig.~\ref{fig:teaser}).
To demonstrate accurate BRDF-emission comparison,
we perform exhaustive experiments on synthetic scenes (Sec.~\ref{subsec:synthetic-brdf},~\ref{subsec:synthetic-relight}) 
while also validating on noisy data of captured real scenes (Sec.~\ref{subsec:realworld}).
The results show our method is able to obtain high-quality reconstruction for complicated indoor scenes that can easily fail for the state-of-the-art (Tab.~\ref{tab:synthetic-brdf}),
yet the training speed is 4-10 times faster (Tab.~\ref{tab:profiling}).
\section{Related Work}
\label{sec:related-work}
\paragraph{Inverse rendering.} 
Inverse rendering aims to estimate the intrinsic properties of an observed scene, via decoupling material, geometry and lighting which jointly contribute to image appearance. Given the inherent ambiguity between the aforementioned high-dimensional factors, classical methods seek to regularize the solution with a surface rendering objective. Approaches include a low-dimensional surface reflectance representation~\cite{yu1999inverse}, sparsity priors for intrinsic images~\cite{barron2013intrinsic}, and spherical-harmonics-based lighting representation~\cite{maier2017intrinsic3d}.
These methods rely on simplified representation of material or lighting, and their regression-based nature calls for heuristic-based priors which may not be appropriate for a wide variety of scenes.

Earlier work can already photorealistically render synthetic objects in a photograph by estimating lighting and geometry~\cite{debevec1998rendering,karsch2011rendering,karsch2014automatic}. 
These methods do not retrieve the materials of the scene, and thus cannot show the reflection of the object on a specular surface in the scene.

\vspace{-8pt}
\paragraph{Learning-based methods.}
Learning-based approaches leverage priors learned from datasets. These methods typically take a single image~\cite{sengupta2019neural,li2020inverse,zhu2022irisformer,wang2021learning,li2022physically} or a pair of stereo images~\cite{srinivasan2020lighthouse}, and apply deep learning models to predict spatially-varying materials and lighting.
Although learned priors help to regularize individual components, these methods do not explicitly model the physics of global light transport and have to rely on approximated inference~\cite{li2018learning}.

Philip~\etal~\cite{philip2021free} take multiple images and aggregate multiview irradiance and albedo information to a pre-trained network to synthesize the relit image. The network takes physically rendered shading using light sources that are semi-automatically estimated as inputs, and outputs an image after relighting. We show in the results that in our synthetic scenes, their method's reliance on the network to render the final image can lead to undesired artifacts, while our use of a physically-based renderer delivers more realistic images.

\vspace{-8pt}
\paragraph{Local or distant lighting.}
Many recent methods aim to model a specific form of light transport. Some methods focus on a single object or distant illumination (environment map)~\cite{gardner2017learning,zhang2021physg,munkberg2022extracting,boss2021nerd,boss2021neural,zhang2021nerfactor}. Srinivasan~\etal~\cite{srinivasan2021nerv} model two-bounce volumetric lighting with known light sources, and Yao~\etal~\cite{yao2022neilf} represent incident radiance as a 5D network. However, optimization of spatially-varying lighting without physically-based constraints is extremely ill-posed especially without abundant observation of light sources. Moreover, object-centric methods do not trivially generalize to indoor settings, where complex lighting effects including occlusion, inter-reflections, and directional highlights call for modeling of long-range interactions of lighting and scene properties.

\vspace{-8pt}
\paragraph{Global light transport.}
Most related to our work, to model general global light transport, recent methods~\cite{azinovic2019inverse,nimierdavid2021material,yu2023milo} build on a per-scene optimization pipeline using a differentiable path tracer~\cite{li2018differentiable,bangaru2020unbiased,nimier2019mitsuba,zeltner2021montecarlo}. 
These methods jointly optimize material and lighting along extensively sampled light paths, and thus are subject to incorrect and slow convergence and high variance due to expensive path queries, gradient propagation, and Monte Carlo sampling, as well as the inherent ambiguity between materials and lighting.
We propose an inverse rendering pipeline that models the global light transport, but converges significantly faster and more accurately than existing methods.
Our variance reduction technique using light baking is inspired by classical rendering methods~\cite{Ward:1988:RTS,krivanek2007practical,seyb20uberbake}, and we tightly integrate the technique in an inverse rendering pipeline. A concurrent work TexIR~\cite{li2022texir} adopts similar ideas to ours by using a pre-baked irradiance as HDR texture map to recover scene materials.
However, they do not model view-dependent light transport and do not estimate emission.
\section{Background}
\label{sec:background}
\begin{table}[t]
    \centering
    \resizebox{0.98\linewidth}{!}{
    \begin{tabular}{|l|l|}
    \hline
    $(\cdot)_+$ & dot product clamped to positive value\\
    $\bm{\omega}_i$ & incident (light) direction\\
    $\bm{\omega}_o$ & outgoing (viewing) direction\\
    $\mathbf{h}$ & half vector: $(\bm{\omega}_i+\bm{\omega}_o)/\lVert\bm{\omega}_i+\bm{\omega}_o\lVert_2$\\
    $\mathbf{n}$ & surface normal\\
    $\mathbf{a}(\mathbf{x})$ & surface base color\\
    $m(\mathbf{x})$ & surface metallic\\
    $\sigma(\mathbf{x})$ & surface roughness\\
    $\mathbf{k}_d(\mathbf{x})$ & diffuse reflectance: $\mathbf{a}(\mathbf{x})(1-m(\mathbf{x}))$\\
    $\mathbf{k}_s(\mathbf{x})$ & specular reflectance: $\mathbf{a(\mathbf{x})}m(\mathbf{x})+0.04(1-m(\mathbf{x}))$\\
    $D(\cdot)$ & GGX normal distribution~\cite{walter2007microfacet}\\
    $F(\cdot)$ & Schlick's approximation of Fresnel coefficient~\cite{schlick1994inexpensive}\\
    $G(\cdot)$ & Geometry (Shadow-Masking) term~\cite{walter2007microfacet}\\
    \hline
    \end{tabular}
    }
    \caption{\textbf{Notations}}
    \label{tab:notations}
\end{table}
Given posed HDR image captures of an indoor scene, our method builds upon input mesh or existing 3D reconstruction algorithms (\eg MonoSDF~\cite{Yu2022MonoSDF}) to further estimate the material and lighting of the scene. 
To ensure the problem is well-constrained, 
we make similar assumptions about scene acquisition as in previous works~\cite{azinovic2019inverse,philip2021free,yu2023milo}
that the dominant light sources and most of the scene geometry are observed in input images.

The material is described as a spatially varying BRDF~\cite{Karis:2013} (including the cosine term) with notations specified in Tab.~\ref{tab:notations}:
\begin{gather}
\begin{gathered}
f(\mathbf{x},\bm{\omega}_i,\bm{\omega}_o)=
\frac{\mathbf{k}_d(\mathbf{x})}{\pi}\left(\mathbf{n}\cdot\bm{\omega}_i\right)_+\\
+\frac{
F(\bm{\omega}_i,\mathbf{h},\mathbf{k}_s(\mathbf{x}))
D(\mathbf{h},\mathbf{n},\sigma(\mathbf{x}))
G(\bm{\omega}_i,\bm{\omega}_o,\mathbf{n},\sigma(\mathbf{x}))
}{4(\mathbf{n}\cdot \bm{\omega}_o)},
\label{eq:brdf}
\end{gathered}
\end{gather}
where $\mathbf{k}_d=\mathbf{a}(1-m)$ and $\mathbf{k}_s=0.04(1-m)+\mathbf{a}m$ are the diffuse and specular reflectance with base color $\mathbf{a}$ and metallic $m$ controlling the two coefficients.
The emitted light is assumed to be view-independent across the surface: $\mathbf{L}_e(\mathbf{x},\bm{\omega}_o)=\mathbf{L}_e(\mathbf{x})$, 
which generalizes well to the emission profile of the indoor scene 
(extension to more complex emitters is possible; 
see Sec.~\ref{subsec:ablation}).

With the parameterization above, our goal is to find $\mathbf{a},\sigma,m,\mathbf{L}_e$ that minimize the difference of renderings with respect to the ground truth over the training images:
\begin{gather}
    \min_{\mathbf{a},\sigma,m,\mathbf{L}_e}\sum_{\mathbf{x},\bm{\omega}_o}
    \left\Vert 
    \mathbf{L}_o(\mathbf{x},\bm{\omega}_o)-\mathbf{L}_{gt}(\mathbf{x},\bm{\omega}_o)
    \right\Vert_2^2
    \label{eq:loss}\\
    \mathbf{L}_o(\mathbf{x},\bm{\omega}_o)=\mathbf{L}_e(\mathbf{x},\bm{\omega}_o)+
    \mathbf{L}_r(\mathbf{x},\bm{\omega}_o)
    \label{eq:rendering-equation}\\
    \mathbf{L}_r(\mathbf{x},\bm{\omega}_o)
    = \int_{\Omega^+} \mathbf{L}_i(\mathbf{x},\bm{\omega}_i)
    f(\mathbf{x},\bm{\omega}_i,\bm{\omega}_o)d\bm{\omega}_i.
    \label{eq:rendering-equation-no-emit}
\end{gather}
\noindent $\mathbf{L}_{gt}$ is a ground truth RGB pixel obtained from camera ray $(\mathbf{x},\bm{\omega}_o)$. $\mathbf{L}_o$ denotes the synthesized rendering following the rendering equation~\cite{kajiya1986rendering}, where $\mathbf{L}_e$ is the surface emitted radiance and $\mathbf{L}_r$ is the reflected radiance, given by integrating incident radiance $\mathbf{L}_i$ times the BRDF response (Eq.~\ref{eq:rendering-equation-no-emit}). 
Note, $\mathbf{L}_i$ is defined in a recursive manner with multi-bounce illumination naturally taken into account of.

\section{Factorized Inverse Path Tracing}
\label{sec:pipt}
To optimize re-rendering error (Eq.~\ref{eq:loss}), previous works~\cite{azinovic2019inverse,nimierdavid2021material} apply differentiable path tracing to solve Eq.~\ref{eq:rendering-equation} and update BRDF and emission jointly with gradient descent.
This approach can be unstable and inefficient:
(1) gradient descent optimization is computationally intensive, which limits the number of path tracing samples and therefore increases the estimation variance;
(2) fundamental ambiguities exist between BRDF and emission, making emission optimization difficult to regularize or converge. To reduce variance in optimization, we propose a factorized light transport representation (Sec.~\ref{subsec:brdf1}) which utilizes pre-baking of diffuse and specular shading maps (Eq.~\ref{eq:shading}) to separate the BRDF coefficients out of the rendering integral. 

\begin{figure}[t]
    \centering
    \setlength\tabcolsep{1.0pt}
    \includegraphics[width=0.99\linewidth]{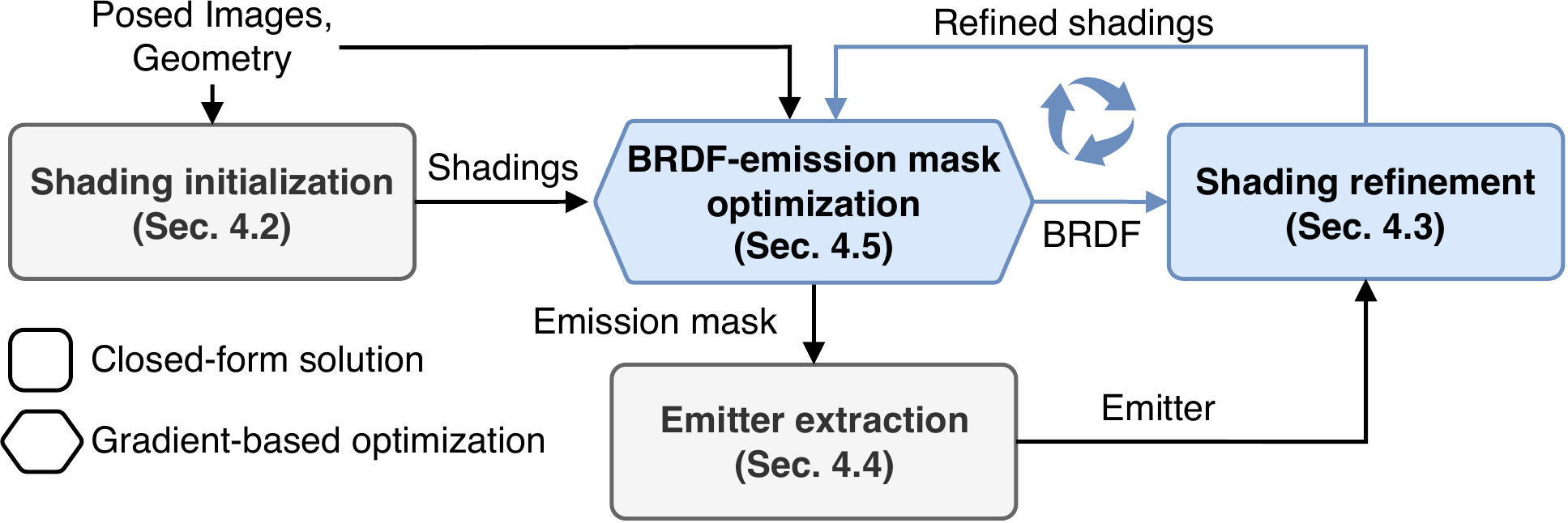}
    \caption{
    \textbf{Our inverse rendering pipeline} approximates diffuse and specular shadings from input images and geometry,
    which are used for efficient renderings during the BRDF-emitter optimization.
    The optimized BRDF and emitters are then passed into a path tracer to refine shadings. 
    The BRDF and the shadings are then updated alternatively.
    }
    \label{fig:pipeline}
\end{figure}

Our full pipeline is demonstrated in Fig.~\ref{fig:pipeline}, 
which optimizes dense BRDF and emission from posed images and scene geometry. 
The pipeline consists of 3 stages: 
(1) first, the factorized diffuse and specular shadings are initialized (baked) as described in Sec.~\ref{subsec:brdf2}; 
(2) given baked shadings, BRDF and emission mask are then optimized (Sec.~\ref{subsec:method}), 
followed by emitter extraction  (Sec.~\ref{subsec:emitter-search}); 
(3) given current BRDF-emission estimation, 
the shadings are refined (Sec.~\ref{subsec:brdf3}), and the algorithm alternates between (2) and (3) until convergence. 

\subsection{Factorized light transport}
\label{subsec:brdf1}
A common way to speed up path tracing in the rendering literature~\cite{Ward:1988:RTS,krivanek2007practical,seyb20uberbake} is to factor the BRDF from the rendering integral, and then to pre-bake and reuse the integral parts. 
Employing a similar idea, we rewrite the reflection equation (Eq.~\ref{eq:rendering-equation-no-emit}) as:
\begin{gather}
\begin{split}
    \mathbf{L}_r(\mathbf{x},\bm{\omega}_o)
    &=\mathbf{k}_d \mathbf{L}_d(\mathbf{x})\\
    &+\mathbf{k}_s\mathbf{L}_s^0(\mathbf{x},\bm{\omega}_o,\sigma)
    +\mathbf{L}_s^1(\mathbf{x},\bm{\omega}_o,\sigma)
    \label{eq:factorized-rendering-equation}
\end{split}\\
\begin{split}
    \mathbf{L}_d(\mathbf{x})&=
    \int_{\Omega^+} \mathbf{L}_i(\mathbf{x},\bm{\omega}_i)
    \frac{(\mathbf{n}\cdot\bm{\omega}_i)_+}{\pi}d\bm{\omega}_i\\
    \mathbf{L}_s^0(\mathbf{x},\bm{\omega}_o,\sigma)&=
    \int_{\Omega^+} \mathbf{L}_i(\mathbf{x},\bm{\omega}_i)
    \frac{F_0DG}{4(\mathbf{n}\cdot \bm{\omega}_o)}
    d\bm{\omega}_i\\
    \mathbf{L}_s^1(\mathbf{x},\bm{\omega}_o,\sigma)&=
    \int_{\Omega^+} \mathbf{L}_i(\mathbf{x},\bm{\omega}_i)
    \frac{F_1DG}{4(\mathbf{n}\cdot \bm{\omega}_o)}
    d\bm{\omega}_i,
    \label{eq:shading}
\end{split}
\end{gather}
\noindent where $\mathbf{L}_d$ is the diffuse shading; $\mathbf{L}_s^0,\mathbf{L}_s^1$ are the two specular shadings associated with two Fresnel components~\cite{schlick1994inexpensive}:
\begin{equation}
\begin{gathered}
    F(\mathbf{h},\bm{\omega}_i,\mathbf{k}_s(\mathbf{x}))=
    \mathbf{k}_s(\mathbf{x})
    F_0+F_1\\
F_0=\left(1-(1-\mathbf{h}\cdot \bm{\omega}_i)^5\right),F_1=(1-\mathbf{h}\cdot \bm{\omega}_i)^5.
\end{gathered}
\end{equation}
\noindent The specular shadings are further approximated by linear interpolation of 6 pre-defined roughness levels:
\begin{equation}
    \mathbf{L}_s^*(\cdot,\sigma)\approx
    \text{lerp}(\{\mathbf{L}_s^*(\cdot,\sigma_k)|\sigma_k \in \text{linspace(0,1,6)}\},\sigma),
\end{equation}
\noindent such that $\mathbf{k}_d,\mathbf{k}_s,\sigma$ are all separated out of the integral.

With this factorization, we can bake the shadings $\mathbf{L}_d,\mathbf{L}_s^0,\mathbf{L}_s^1$ offline for each input view into image buffers, then query the shading pixels at training time to speed up rendering.
Baking shadings offline allows us to use a large sampling rate for variance reduction.
Owing to its linear formulation, we also empirically found this factorized rendering handles mirror-like objects much better (Fig.~\ref{fig:synthetic_render}), 
while standard Monte Carlo integration can easily obtain unstable gradients caused by the large BRDF value.

\begin{figure}[t]
    \centering
    \setlength\tabcolsep{0.5pt}
    \resizebox{0.99\linewidth}{!}{
    \begin{tabular}{ccccc}
    \multicolumn{2}{c}{}&
    \multicolumn{3}{c}{Estimated}\\
    \multicolumn{2}{c}{
    \begin{tabular}{c}
    \includegraphics[width=0.58\linewidth]{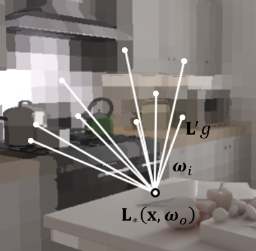}\\
    \multicolumn{1}{c}{}
    \end{tabular}
    }
    &
    \multicolumn{3}{c}{
    \begin{tabular}{ccc}
    \includegraphics[width=0.25\linewidth]{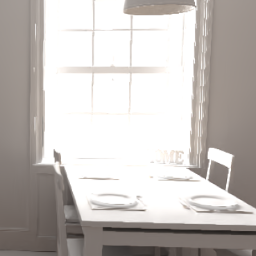}&
    \includegraphics[width=0.25\linewidth]{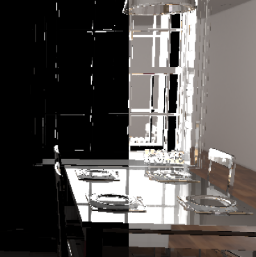}&
    \includegraphics[width=0.25\linewidth]{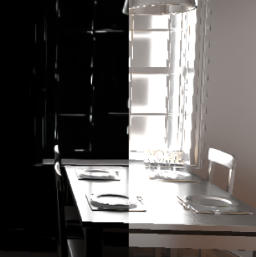}\\
    \multicolumn{3}{c}{Ground truth}\\
    \includegraphics[width=0.25\linewidth]{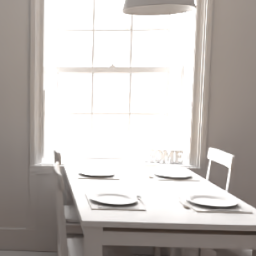}&
    \includegraphics[width=0.25\linewidth]{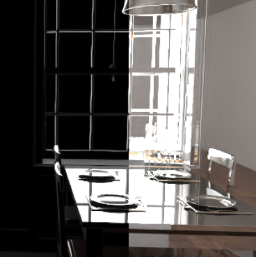}&
    \includegraphics[width=0.25\linewidth]{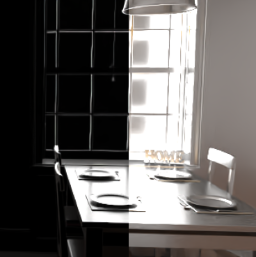}\\
    $\mathbf{L}_d$ & $\mathbf{L}_s^*(\sigma=0.02)$ & $\mathbf{L}_s^*(\sigma=0.2)$
    \end{tabular}
    }
    \end{tabular}
    }
    \caption{\textbf{Diffuse and specular shadings are initialized} 
    by tracing a voxel representation of the surface light field $\mathbf{L}'$ (left),
    which gives approximations (top row on right) close to the ground truth (bottom row on the right; obtained by path tracing).
    }
    \label{fig:shading-comparison}
\end{figure}
\subsection{Image-based shading initialization}
\label{subsec:brdf2}
As we pre-bake the shadings $\mathbf{L}_d,\mathbf{L}_s^0,\mathbf{L}_s^1$, they are fixed during BRDF-lighting estimation. 
They need to be properly initialized; otherwise, the optimization will not converge.
For simplicity, we abstract the integrands in Eq.~\ref{eq:shading} to the form: $\mathbf{L}_*=\mathbf{L}_i g$,
where $g$ denotes the factorized BRDF term.
In path tracing notation, the shading integral can be initialized by querying a surface light field approximation:
\begin{equation}
\begin{gathered}
\mathbf{L}_*(\mathbf{x})=
\mathbf{L}(\mathbf{x}_1 \rightarrow \mathbf{x})
g(\mathbf{x}_1 \rightarrow \mathbf{x})\\
\approx \mathbf{L}'(\mathbf{x}_1)g(\mathbf{x}_1\rightarrow \mathbf{x}),
\end{gathered}
\label{eq:shading-pathtrace}
\end{equation}
where $\mathbf{L}$ is the exact surface light field at sampled location $\mathbf{x}_1$ towards $\mathbf{x}$,
and $\mathbf{L}'$ is its approximation obtained by average pooling all the input pixels onto a voxel grid spanned on the scene geometry (Fig.~\ref{fig:shading-comparison} left).
Since objects in an indoor scene are often near-diffuse, 
and the renderings are essentially low pass filtering the incident light field~\cite{ramamoorthi2001efficient} that blurs the detail, 
we find that using a $256^3$ voxel grid with nearest neighbor radiance query gives good shading approximations (Fig.~\ref{fig:shading-comparison} right).

\subsection{Path-traced shading refinement}
\label{subsec:brdf3}
\begin{figure}
    \centering
    \setlength\tabcolsep{1.0pt}
    \resizebox{0.99\linewidth}{!}{
    \begin{tabular}{cc}
    \begin{tabular}{c}
    \includegraphics[width=0.6\linewidth]{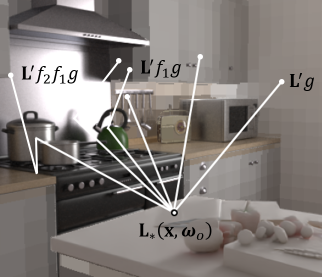}\\
    \multicolumn{1}{c}{}
    \end{tabular}
    &
    \begin{tabular}{ccc}
    \includegraphics[width=0.25\linewidth]{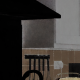}&
    \includegraphics[width=0.25\linewidth]{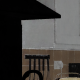}&
    \includegraphics[width=0.25\linewidth]{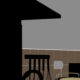}\\
    \includegraphics[width=0.25\linewidth]{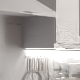}&
    \includegraphics[width=0.25\linewidth]{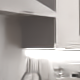}&
    \includegraphics[width=0.25\linewidth]{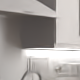}\\
    Origin & Refined & Ground truth
    \end{tabular}
    \end{tabular}
    }
    \caption{
    \textbf{Shading refinement:} The cabinet's diffuse reflectance estimation is initially darker than ground truth, 
    owing to the excessive incident light received from the range hood that reflects non-diffuse light (2nd column).
    The artifacts are reduced by growing the path for the specular surface according to the optimized BRDF (1st column),
    which gives more accurate shadings that can be used to further refine the BRDF (3rd column).
    }
    \label{fig:shading-update}
\end{figure}

\begin{figure}
    \centering
    \setlength\tabcolsep{1.0pt}
    \resizebox{0.99\linewidth}{!}{
    \begin{tabular}{cccc}
    \includegraphics[width=0.28\linewidth]{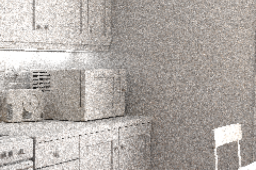}&
    \includegraphics[width=0.28\linewidth]{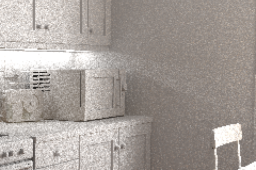}&
    \includegraphics[width=0.28\linewidth]{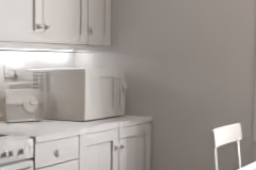}&
    \includegraphics[width=0.28\linewidth]{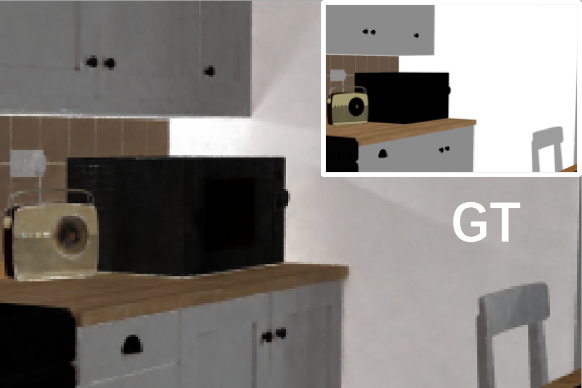}\\
    w/o cache & w/ cache & GT & $\mathbf{k}_d$ w/o cache 
    \end{tabular}
    }
    \caption{\textbf{Diffuse radiance cache} 
    from $\mathbf{L}'$ helps reduce variance and error for shading estimation (2nd image).
    Without it, sampling the tiny emitters below the cabinet will be difficult (1st image),
    which leads to incorrect shading and albedo (4th image).
    }
    \label{fig:diffuse-radiance-cache}
\end{figure}

Eq.~\ref{eq:shading-pathtrace} gives incorrect shading if the surface light field is sampled at locations that are mainly specular ($\mathbf{L}$ is view dependent), 
which subsequently leads to incorrect BRDF estimation.
Given BRDF-emitter estimations optimized from Eq.~\ref{eq:loss} under current shading estimations, 
we re-estimate the light transport on specular surfaces by growing the path in Eq.~\ref{eq:shading-pathtrace} until the ray either hits an emitter or intersects with a near diffuse surface (identified by $\sigma>0.6$; Fig.~\ref{fig:shading-update}):
\begin{gather}
\begin{gathered}
\mathbf{L}_*(\mathbf{x})=
\mathbf{R}(\mathbf{x}_n)\prod_{i=1}^{n-1}f(\mathbf{x}_{i+1}\rightarrow \mathbf{x}_i)
g(\mathbf{x}_1 \rightarrow \mathbf{x}),\\
\text{s.t. } \sigma(\mathbf{x}_i)\leq 0.6 \text{, } \forall i < n
\label{eq:shading-refine}
\end{gathered}\\
\mathbf{R}(\mathbf{x}_n)=
\begin{cases}
\mathbf{L}'(\mathbf{x}_n) & \sigma(\mathbf{x}_n)>0.6\\
\mathbf{L}_e(\mathbf{x}_n) & \mathbf{L}_e(\mathbf{x}_n) > 0
\end{cases},
\end{gather}
\noindent where $n$ is the length of a certain path before it terminates. The above equation essentially estimates the shadings by multi-bounced path tracing with $\mathbf{L}'$ being a diffuse radiance cache,
which helps speed up the evaluation and also reduce error: 
initial estimations of $f$s may retain large error but $\mathbf{L}'$ is very close to a diffuse surface light field (as it is also view-independent). Most of the path hits a diffuse surface within one to two bounces,
such that the errors from the BRDF will not be magnified (Fig.~\ref{fig:diffuse-radiance-cache}).

Substituting shadings in factorized rendering by their refinements makes
Eq.~\ref{eq:factorized-rendering-equation} more closely match the ground truth light transport, such that BRDF can be re-estimated with fewer artifacts (Fig.~\ref{fig:shading-update}: `Origin' VS `Refined').
The re-estimated BRDF in turn is applied to further improve the shadings,
and this BRDF and shading refinement is performed alternatively until convergence.

\subsection{Error-driven emitter estimation}
\label{subsec:emitter-search}
\begin{figure}[t]
    \centering
    \setlength\tabcolsep{1.0pt}
    \resizebox{0.9\linewidth}{!}{
    \begin{tabular}{cccc}
    \includegraphics[width=0.3\linewidth]{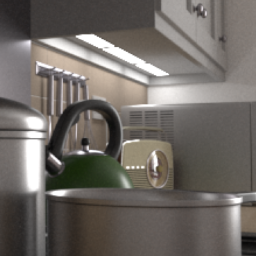}&
    \includegraphics[width=0.3\linewidth]{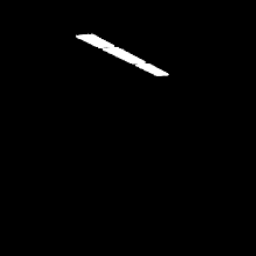}&
    \includegraphics[width=0.3\linewidth]{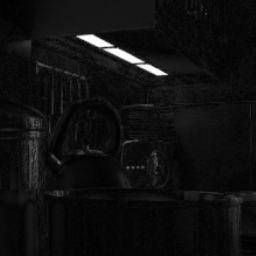}&
    \includegraphics[width=0.3\linewidth]{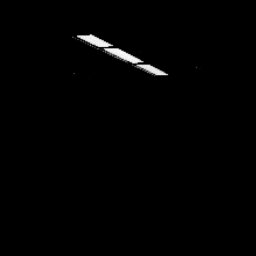}\\
    RGB & GT emission & Rendering error & Emission mask
    \end{tabular}
    }
    \caption{
    \textbf{Rendering images without emission terms} produces distinctive error near emissive surfaces (3rd image). By jointly optimizing an emission mask (4th image) to cancel this error, the emitter can be found by checking the mask's response, which is robust even for tiny emitters (2nd image for ground truth).
    }
    \label{fig:emission-error}
\end{figure}

If we replace rendering equation Eq.~\ref{eq:rendering-equation} by Eq.~\ref{eq:rendering-equation-no-emit} that excludes the emission,
the objective Eq.~\ref{eq:loss} still converges for non-emissive surfaces (as their $\mathbf{L}_e=0$),
but regions with emission will present large errors,
which is a good indicator of emitters (Fig.~\ref{fig:emission-error}).
With this intuition, we introduce an emission mask (encouraged to be small) $\alpha\in[0,1]$ to the rendering loss:
\begin{gather}
\begin{gathered}
\min_{\mathbf{a},\sigma,m,\alpha}\sum_{\mathbf{x},\bm{\omega}_o}
    \left\Vert 
    (1-\alpha)\mathbf{L}_r+\alpha\mathbf{L}_{gt}
    -\mathbf{L}_{gt}
    \right\Vert_2^2\\
    \text{s.t. }
    \alpha \rightarrow 0.
\end{gathered}
\label{eq:loss-masked}
\end{gather}

When a surface is non-emissive, $\alpha$ will stay small owing to the regularization
and the loss is minimized by adjusting $\mathbf{L}_r$ towards $\mathbf{L}_{gt}$;
but $\mathbf{L}_r$ cannot model the emission,
so $\alpha$ for an emissive surface has to become large to accommodate the error.
In practice we apply a L1 sparsity loss to $\alpha$. 
By changing the optimization objective to Eq.~\ref{eq:loss-masked},
we first jointly estimate the BRDF-emission mask, and then threshold the mask to find the emitter ($\alpha>0.01$).
Afterward, each emitter's emission $\mathbf{L}_e$ is estimated independently from BRDF:
\begin{equation}
\mathbf{L}_e=
\begin{cases}
\underset{\mathbf{L}_e}{\arg\min} 
\underset{\mathbf{x},\bm{\omega}_o}{\sum}
\left\Vert 
\mathbf{L}_e + \mathbf{L}_r
-\mathbf{L}_{gt}
\right\Vert_1 & \alpha > 0.01\\
0 & \text{otherwise}
\end{cases}.
\label{eq:loss-emission}
\end{equation}
Our formulation is found to be more stable than joint optimization 
(demonstrated in ablations in Sec.~\ref{subsec:ablation}),
because the emission mask value is in the same range as BRDF coefficients, 
such that the gradient update is balanced between the BRDF and the emission mask.
In contrast, surface emission can be much larger than BRDF coefficients, 
making it more difficult to directly fit or regularize.

\vspace{-8pt}
\paragraph{Emitter extraction.}
We assume emission is constant for each mesh triangle.
After $\alpha$ is optimized, we uniformly sample 100 locations on each triangle and find their corresponding $\alpha$ value.
A triangle is then classified as an emitter if the mean of its $\alpha$s is above 0.01.
Eq.~\ref{eq:loss-emission} in general is ill-posed (\eg $\mathbf{k}_d$ can be increased by decreasing $\mathbf{L}_e$),
so we make the assumption that an emitter reflects zero light ($f=0$).
In such a situation, $\mathbf{L}_e$ for a triangle has the closed-form solution as the median of RGBs from all input pixels it intersects,
which does not require any gradient descent optimization,
so it can be estimated efficiently and accurately.

\subsection{Optimization}
\label{subsec:method}
 
Given either initial or refined shadings,
the BRDF and emission mask are optimized using the objective in Eq.~\ref{eq:loss-masked}.
We encode BRDF and the emission mask with two MLPs:
\begin{equation}
\begin{gathered}
(\mathbf{a},m,\sigma)
=\text{Sigmoid}\left(\text{MLP}_\text{brdf}(\mathbf{x})\right)\\
\alpha = 1-\exp{\left(
    -\text{ReLU}(\text{MLP}_\text{emit}(\mathbf{x}))
\right),}
\end{gathered}
\label{eq:mlp}
\end{equation}
where $\text{MLP}_\text{brdf}$ uses hash encoding~\cite{muller2022instant}
and $\text{MLP}_\text{emit}$ is a positional encoded MLP~\cite{mildenhall2020nerf}. 
The objective Eq.~\ref{eq:loss-masked} is converted to a gradient descent loss function as a tone-mapped L2 loss $l$ plus a L1 regularization term $l_e$:
\begin{gather}
    l=\sum_{\mathbf{x,o}}
    \left\Vert 
    \Gamma((1-\alpha)\mathbf{L}_r+\alpha\mathbf{L}_{gt})
    -\Gamma(\mathbf{L}_{gt})
    \right\Vert_2^2
    \label{eq:loss-tonemapped}
    \\
    l_e=\lambda_e\sum_\mathbf{x}\Vert\text{MLP}_\text{emit}(\mathbf{x})\Vert_1,\lambda_e=1,
    \label{eq:reg-emission}
\end{gather}
where $\Gamma$ is the tone-mapping function by Munkberg \etal~\cite{munkberg2022extracting} to help suppress noise from high dynamic range values.
We prefer neural networks rather than a textured mesh (as in~\cite{ azinovic2019inverse,nimierdavid2021material}) as scenes with complex geometries can create degenerate UVs, which reduces the BRDF quality.

\vspace{-8pt}
\paragraph{Roughness-metallic regularization.}
\begin{figure}[t]
    \centering
    \setlength\tabcolsep{0.5pt}
    \resizebox{0.99\linewidth}{!}{
    \begin{tabular}{ccccc}
    \includegraphics[width=0.22\linewidth]{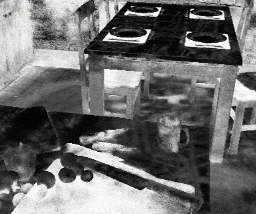}&
    \includegraphics[width=0.22\linewidth]{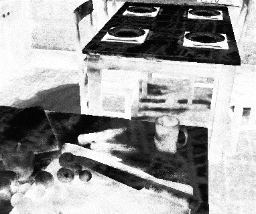}&
    \includegraphics[width=0.22\linewidth]{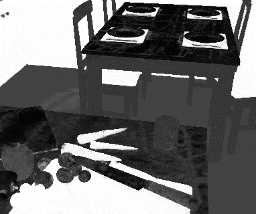}&
    \includegraphics[width=0.22\linewidth]{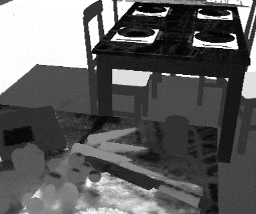}&
    \includegraphics[width=0.22\linewidth]{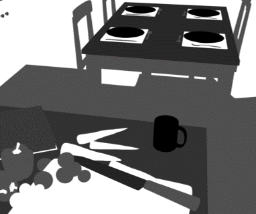}\\
    w/o $l_d,l_p$&w/o $l_p$&
    w/ part& w/ semantic&
    GT
    \end{tabular}
    }
    \caption{
    \textbf{Roughness optimization} can be ambiguous without any regularization (1st image). 
    By encouraging a surface to be diffuse, specular surfaces still get an incorrect roughness value if no highlights are observed (2nd image).
    The roughness can be more reasonably estimated with part segmentation for guidance (3rd image).
    Semantic segmentation (4th image) shows similar results except the roughness for small objects get blurred. 
    }
    \label{fig:roughness-regularization}
\end{figure}
Surface roughness and metallic can take arbitrary values if there are no highlights ($\mathbf{L}_s^0, \mathbf{L}_s^1 \approx 0$), which leads to ambiguity.
We prevent this by encouraging surfaces to be diffuse:
\begin{equation}
    l_d=\lambda_d \sum_{\mathbf{x}}
    \left(
    \Vert 1-\sigma(\mathbf{x}) \Vert_1
    + \Vert m(\mathbf{x}) \Vert_1 \right),
    \lambda_d = \text{5e-4},
\end{equation}
such that a diffuse surface will not be misinterpreted as a specular surface with weak reflection.
To get valid roughness-metallic for input pixels that do not observe highlights,
we further assume they stay constant inside each material part, and utilize image-level part segmentation to group input pixels.
The roughness-metallic from pixels with highlights are propagated to their corresponding group by another regularization loss:
\begin{gather}
    l_p=\lambda_p \sum_\mathbf{x}
    \left\Vert
    \begin{bmatrix}
    \sigma(\mathbf{x}) \\
    m(\mathbf{x})
    \end{bmatrix}
    -
    \begin{bmatrix}
    \sigma'(\mathbf{x}) \\
    m'(\mathbf{x})
    \end{bmatrix}
    \right\Vert_1,
    \lambda_p=\text{5e-3}
    \\
    \begin{gathered}
    \begin{bmatrix}
    \sigma'(\mathbf{x}) \\
    m'(\mathbf{x})
    \end{bmatrix}
    =
    \sum_{\text{Seg}(\mathbf{x'})=\text{Seg}(\mathbf{x})}
    \frac{w(\mathbf{x'})}
    {\sum_{\mathbf{x}'}w(\mathbf{x}')}
    \begin{bmatrix}
    \sigma(\mathbf{x'})\\
    m(\mathbf{x'})
    \end{bmatrix}
     \\
    w(\mathbf{x'})=\text{sg}\left(
    \Vert
    \mathbf{k_s}\mathbf{L}_s^0+\mathbf{L}_s^1
    \Vert_1\right),
    \end{gathered}
    \label{eq:part-loss}
\end{gather}
where $\text{Seg}(\mathbf{x})$ gives the segmentation ID for $\mathbf{x}$,
 $\text{sg}(\cdot)$ denotes stop the gradient,
and $w(\mathbf{x})$ is a propagation kernel that weights the pixel by the amount of highlights.
While part segmentation in practice can be hard to obtain,
semantic segmentation is readily available from pre-trained model \eg Mask2Former~\cite{cheng2021mask2former}, 
where multiple material parts may stay inside the same semantic label.
To account for such detail loss, we consider two pixels belong to the same material part only if: (1) they share the same semantic ID; (2) have similar albedo value; (3) and are close to each other,
which suggests an alternative propagation kernel $w(\mathbf{x,x'})$:
\begin{equation}
\begin{gathered}
    w(\mathbf{x}',\mathbf{x})=
\text{sg}\left(
e^{-\frac{\Vert
\mathbf{a}(\mathbf{x})-\mathbf{a}(\mathbf{x'})
\Vert_2^2}{2\sigma_a^2}
}
e^{-\frac{\Vert
\mathbf{x}-\mathbf{x'}
\Vert_2^2}{2\sigma_x^2}}
\right)
\\
\sigma_a = \text{1.6e-2},\sigma_x=\text{1e-2}.
\end{gathered}
\label{eq:semantic-loss}
\end{equation}
By replacing $\mathbf{w}(\mathbf{x})$ with $\mathbf{w}(\mathbf{x,x}')$ and changing the regularization weight to $\lambda_p=\text{1e-3}$,
we can still have reasonable roughness-metallic estimation even with semantic segmentation (Fig.~\ref{fig:roughness-regularization}).
\section{Experiments}
\label{sec:experiments}

\begin{figure*}
    \centering
    \setlength{\tabcolsep}{0.5pt}

\resizebox{0.99\textwidth}{!}{
\begin{tabular}{cccc @{\hskip 0.04in}|@{\hskip 0.04in} cccc}
\multicolumn{8}{c}{\textbf{BRDF and emission estimation}}\\[-4ex]
\multicolumn{6}{c}{}&
\multicolumn{2}{r}{
\begin{tabular}{rr}
\includegraphics[width=.07\textwidth]{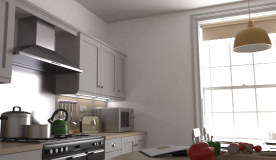}&
\includegraphics[width=.07\textwidth]{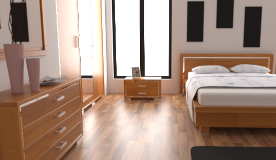}
\end{tabular}
}\\[-0.75ex]

\multicolumn{1}{c}{\lieccv}&
\multicolumn{1}{c}{\ipt}&
\multicolumn{1}{c}{\ours}&
\multicolumn{1}{c}{Ground truth}&

\multicolumn{1}{c}{\lieccv}&
\multicolumn{1}{c}{\ipt}&
\multicolumn{1}{c}{\ours}&
\multicolumn{1}{c}{Ground truth}\\
 
\includegraphics[width=.13\textwidth]{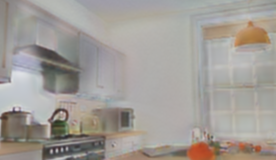} &
\includegraphics[width=.13\textwidth]{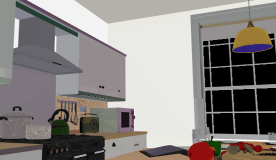} &
\includegraphics[width=.13\textwidth]{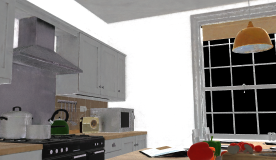} &
\includegraphics[width=.13\textwidth]{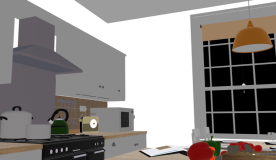} &

\includegraphics[width=.13\textwidth]{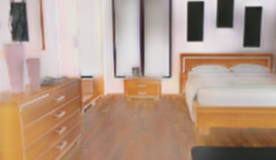} &
\includegraphics[width=.13\textwidth]{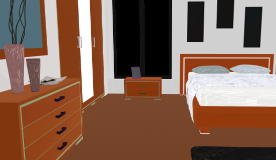} &
\includegraphics[width=.13\textwidth]{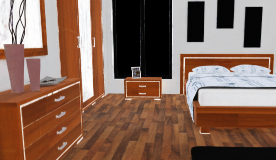} &
\includegraphics[width=.13\textwidth]{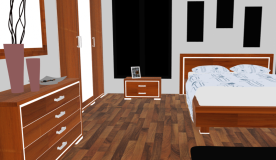} \\[-0.75ex]

\includegraphics[width=.13\textwidth]{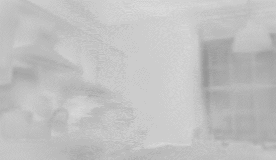} &
\includegraphics[width=.13\textwidth]{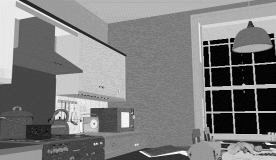} &
\includegraphics[width=.13\textwidth]{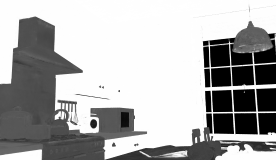} &
\includegraphics[width=.13\textwidth]{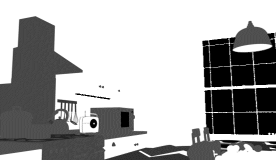} &

\includegraphics[width=.13\textwidth]{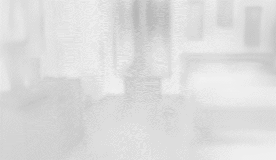} &
\includegraphics[width=.13\textwidth]{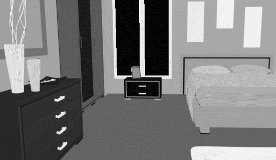} &
\includegraphics[width=.13\textwidth]{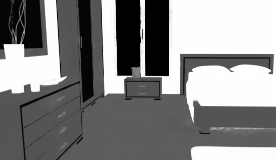} &
\includegraphics[width=.13\textwidth]{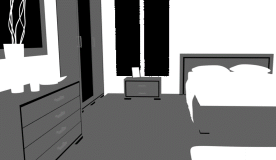} \\[-0.75ex]

\includegraphics[width=.13\textwidth]{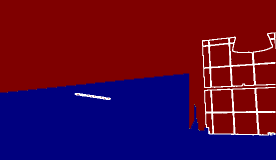} &
\includegraphics[width=.13\textwidth]{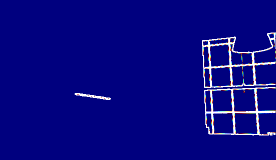} &
\includegraphics[width=.13\textwidth]{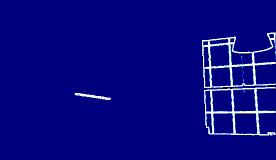} &
\includegraphics[width=.13\textwidth]{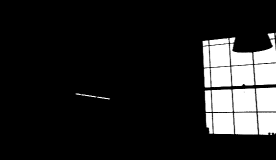} &

\includegraphics[width=.13\textwidth]{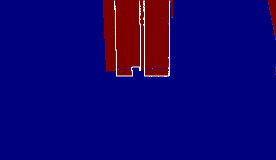} &
\includegraphics[width=.13\textwidth]{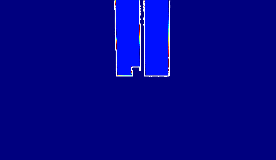} &
\includegraphics[width=.13\textwidth]{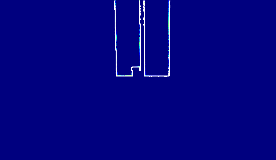} &
\includegraphics[width=.13\textwidth]{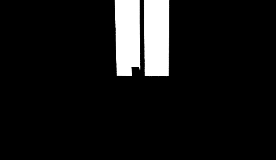}\\

\end{tabular}
}

\caption{
\textbf{BRDF and emission estimation results on 2 synthetic scenes} 
shows our method successfully reconstructs material reflectance (1st row), roughness (2nd row), and emission (3rd row) with high frequency details and less ambiguity. Emission estimation is shown as error heatmaps (warmer colors indicate higher emission error; GT emitter boundary is marked in white lines). Input views are shown in upper-right corner.
}
\label{fig:synthetic_BRDF}
\end{figure*}
\begin{figure*}
    \centering
    \setlength{\tabcolsep}{0.5pt}
\resizebox{0.99\textwidth}{!}{

\begin{tabular}{cccc @{\hskip 0.04in}|@{\hskip 0.04in} cccc}
\multicolumn{4}{c}{\textbf{View synthesis}}&
\multicolumn{4}{c}{\textbf{Relighting}}\\

\fvp & \ipt & \ours & Ground truth&
\fvp & \ipt & \ours & Ground truth\\

\includegraphics[width=.13\textwidth]{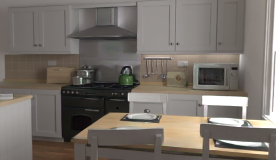}&
\includegraphics[width=.13\textwidth]{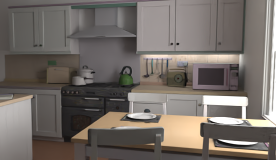} &
\includegraphics[width=.13\textwidth]{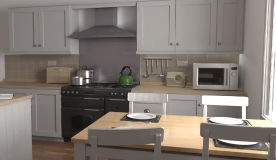} &
\includegraphics[width=.13\textwidth]{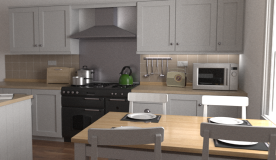}&

\includegraphics[width=.13\textwidth]{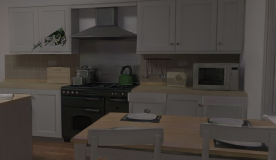}&
\includegraphics[width=.13\textwidth]{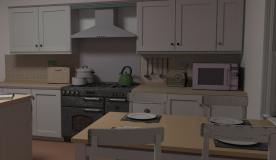} &
\includegraphics[width=.13\textwidth]{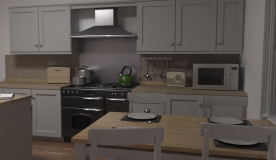} &
\includegraphics[width=.13\textwidth]{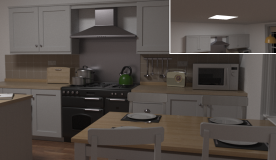}\\[-0.7ex]

\includegraphics[width=.13\textwidth]{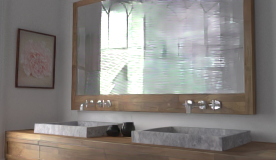}&
\includegraphics[width=.13\textwidth]{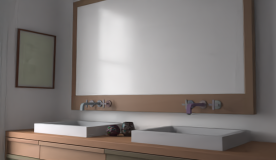} &
\includegraphics[width=.13\textwidth]{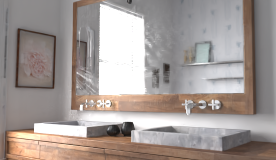} &
\includegraphics[width=.13\textwidth]{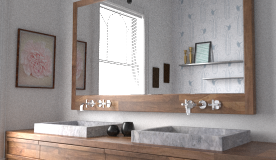}&

\includegraphics[width=.13\textwidth]{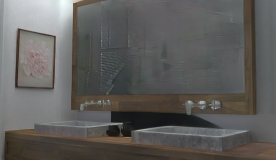}&
\includegraphics[width=.13\textwidth]{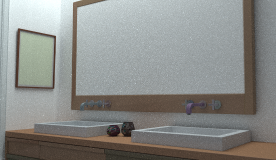} &
\includegraphics[width=.13\textwidth]{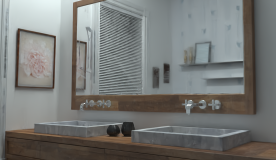} &
\includegraphics[width=.13\textwidth]{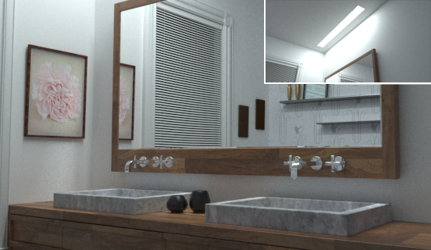}\\
\end{tabular}
}
\caption{
\textbf{Qualitative results of view synthesis (left) and relighting (right) on 2 synthetic scenes} 
demonstrate accurate light transport can be simulated with our estimated BRDF and emission even for very specular surfaces.
The reflection of the chair from the microwave oven can be seen in kitchen scene on top,
and mirrors are correctly rendered for bathroom (bottom).
}
\label{fig:synthetic_render}
\end{figure*}

We evaluate our method on 4 synthetic and 2 real indoor scenes,
where the synthetic scenes are obtained from Bitterli's rendering resources~\cite{resources16} with large glass objects being removed (as we do not model transmission),
and the real scenes are captured by us.
Each synthetic scene contains around 200 posed HDR images generated by Mitsuba3~\cite{jakob2022mitsuba3},
per-camera view BRDF-emission maps generated by Blender~\cite{blender22},
and ground truth geometry. For synthetic scenes, we show our method with both part segmentation (\ours) and semantic segmentation mask (\ours-sem).

The real scenes (Conference room and Classroom) are captured by a Sony A7M3 camera with around 200 HDR images reconstructed by 5-stop exposure bracketing.
The camera poses are estimated from COLMAP~\cite{schoenberger2016mvs} and the geometry is reconstructed using MonoSDF~\cite{Yu2022MonoSDF}. Please refer to the supplementary material for details on real scene capturing and additional results.

\subsection{Synthetic: BRDF-emission estimation}
\label{subsec:synthetic-brdf}
While synthetic scenes allow us to directly compare with the ground truth BRDF-emission without noise from geometry or image captures,
BRDF parameterizations can vary across different baselines.
For fair comparison, 
we empirically found 
diffuse reflectance $\mathbf{k}_d$ for diffuse surfaces, roughness $\sigma$, 
and the material reflectance defined by $\mathbf{a}'=\int_{\Omega^+} fd\bm{\omega}_i$
are very close across different BRDF models~\cite{burley2012physically,Karis:2013}.
We therefore measure the PSNR for these metrics in image space for BRDF comparison.
The $\mathbf{k}_d$ is compared only for diffuse surfaces,
and $\mathbf{a}'$ is estimated using Monte Carlo integration of 128 samples per pixel.
For emission, we estimate the IoU of emission mask and $\log \text{L2}$ error of emission map.

\vspace{-8pt}
\paragraph{Baselines.}
We compare with the original inverse path tracing (IPT)~\cite{azinovic2019inverse} and its extension MILO~\cite{yu2023milo}
that also parameterizes spatially varying BRDF with neural networks.
IPT assumes BRDF parameters to be constant inside each mesh triangle,
and MILO takes manual input of number of emitters.
Both IPT and MILO are evaluated by their original authors due to non-public code, and the MILO training is stopped after 10 hours.
Meanwhile, we also compare NeILF~\cite{yao2022neilf} that models illumination in an unconstrained way and the learning based approach~\cite{li2022physically} (Li22) for single-view inverse rendering.

\begin{table}[t]
    \centering
    \setlength\tabcolsep{3.0pt}
    \resizebox{0.99\linewidth}{!}{
\begin{tabular}{clccccc}
\toprule
& \multirow{2}{*}{Method} & 
$\mathbf{k}_d$ & $\mathbf{a}'$ & $\sigma$ &
\multicolumn{2}{c}{$\mathbf{L}_e$} \\
&
&\multicolumn{3}{c}{PSNR$\uparrow$} & 
IoU$\uparrow$ & logL2$\downarrow$\\
\midrule
\multirow{5}{*}{Bathroom} &
\lieccv & 19.92 & 15.78 & 13.77 & 0.45 & 1.35\\
&\neilf & 10.12 & 9.01 & 14.82 & - & - \\
& \ipt & 22.43 & 18.59 & 14.69 & 0.33 & 1.09e-1\\
&\milo & 11.83 & 9.80 & 5.56 & 0.05 & 5.60e-1\\
&\ours & \textbf{30.13} & \textbf{25.28} & \textbf{28.79} & \textbf{0.63} & \textbf{3.18e-2}\\
&\ours-sem & 27.81 & 24.00 & 21.84 & \textbf{0.63} & \textbf{3.18e-2}\\

\midrule
\multirow{5}{*}{Bedroom} &
\lieccv & 21.87 & 17.18 & 12.12 & 0.34 & 2.78\\
&\neilf & 14.88 & 12.42 & 11.30 & - & - \\
& \ipt & 29.39 & 22.46 & 13.33 & 0.92 & 4.01e-3\\
&\milo & 23.65 & 15.16 & 15.42 & 0.08 & 1.59e-2\\
&\ours & \textbf{31.10} & \textbf{29.41} & 23.19 & \textbf{0.96} & 4.95e-4\\
&\ours-sem & 31.00 & 28.45 & \textbf{25.23} & \textbf{0.96} & \textbf{4.93e-4}\\

\midrule
\multirow{5}{*}{Livingroom} &
\lieccv & 17.25 & 15.32 & 12.72 & 0.17 & 3.61\\
&\neilf & 12.34 & 10.97 & 13.45 & - & - \\
& \ipt & 21.24 & 19.01 & 11.77 & 0.90 & 6.08e-3\\
&\milo & 22.88 & 18.39 & 13.98 & 0.06 & 1.39e-2\\
&\ours & 28.86 & \textbf{28.70} & \textbf{32.48} & \textbf{0.95} & \textbf{8.06e-4}\\
&\ours-sem & \textbf{29.09} & 28.62 & 25.15 & \textbf{0.95} & 8.09e-4\\

\midrule
\multirow{5}{*}{Kitchen} &
\lieccv & 18.14 & 14.54 & 10.82 & 0.43 & 1.41\\
&\neilf & 12.63 & 9.96 & 10.64 & - & - \\
&\ipt & 25.68 & 21.61 & 11.84 & 0.83 & 1.08e-2\\
&\milo & 18.25 & 13.86 & 12.56 & 0.10 & 8.28e-2\\
&\ours & 33.07 & \textbf{27.53} & \textbf{29.24} & \textbf{0.91} & \textbf{1.54e-3}\\
&\ours-sem & \textbf{33.25} & 27.38 & 21.70 & \textbf{0.91} & \textbf{1.54e-3}\\
\bottomrule
\end{tabular}
}
    \caption{\textbf{BRDF-emission comparison on synthetic scenes} 
    shows that our method gives the overall best reconstruction.
    The results are similar even if only semantic segmentation is provided (\ours-sem).
    NeILF does not estimate emitters. The best method is marked in bold.
    }
    \label{tab:synthetic-brdf}
\end{table}
\begin{table}[t]
\centering
\setlength\tabcolsep{2.0pt}
\resizebox{0.99\linewidth}{!}{
\begin{tabular}{cc}
\begin{tabular}{lccc}
\toprule
\multicolumn{4}{c}{Our per-stage profiling}\\
& Stage 1 & Stage 2 & Stage 3\\
\midrule
Memory & 3.2GB & 2.8GB & 3.4GB\\
Time & 6min & 2min & 16min\\
\bottomrule
\end{tabular}&
\begin{tabular}{lc}
\toprule
Method & training time$\downarrow$\\
\midrule
\neilf & 1h38min\\
\ipt & $\approx$3hr\\
\milo &  $\approx$10hr\\
\ours & \textbf{44min}\\
\bottomrule
\end{tabular}
\end{tabular}
}
\caption{
\textbf{Averaged training speed comparison} 
suggests our method is very efficient (right table).
The per-stage profiling is shown on the left with Stage 2 and 3 being repeated twice.
The comparison is made on a 3090Ti GPU.
}
\label{tab:profiling}
\end{table}

\vspace{-8pt}
\paragraph{Results.}
As is shown in Tab.~\ref{tab:synthetic-brdf}, 
our method gives the best BRDF and emission estimation with the fastest training speed (Tab.~\ref{tab:profiling}) even when only semantic level segmentation is provided (\ours-sem).
The learning-based approach (Li22) fails to generate reasonable reconstruction as it does not utilize multi-view cues,
while unconstrained optimization (NeILF) suffers from the ambiguity between material and lighting.
While IPT converges, its accuracy is limited by a piece-wise constant constraint to reduce the variance.
MILO also fails to reconstruct high frequency details because of the Monte Carlo noise from path tracing, and it requires manual specification of the number of emitters to constrain the emission optimization.
In contrast, our method requires no human input during optimization,
which allows more stable and faster convergence with results that match the ground truth well (Fig.~\ref{fig:synthetic_BRDF}).

\subsection{Synthetic: view synthesis and relighting}
\label{subsec:synthetic-relight}
To demonstrate the applications of inverse rendering outputs, we compare the rendered scenes under novel views and novel lighting using estimated BRDF and emission.
For quantitative comparison, we tone-map the rendered images with $\gamma=1/2.2$ then calculate their PSNR with respect to the ground truth.

\vspace{-8pt}
\paragraph{Baselines.}
Besides IPT, MILO, and Li22, we also consider FVP~\cite{philip2021free} that performs view synthesis and relighting in a learning-based way. 
FVP assumes emissions come from saturated regions on the images,
which may wrongly classify surfaces with strong reflection as emitters.
So we offer ground truth emission to FVP instead as oracle.
The renderings for IPT, MILO, and our method are obtained by path tracing with 1024 samples per pixel, which is further denoised by the Optix denoiser~\cite{parker2010optix}.

\begin{table}[t]
\centering
\setlength\tabcolsep{2.0pt}
\resizebox{0.99\linewidth}{!}{
\begin{tabular}{clcccc}
\toprule
&Method&
Bathroom & Bedroom & Livingroom & Kitchen\\
\midrule
\multirow{5}{*}{\makecell{View \\ synthesis}}
&\fvp & 23.38 & 20.49 & 24.63 & 20.77\\
&\ipt & 14.76 & 21.85 & 23.87 & 19.94\\
&\milo & 20.62 & 20.25 & 24.47 & 18.09\\
&\ours & 25.42 & 29.84 & \textbf{30.86} & \textbf{25.38}\\
&\ours-sem & \textbf{25.76} & \textbf{29.89} & 30.84 & 25.27\\
\midrule
\multirow{5}{*}{Relight}
&\lieccv & 22.86 & 23.20 & 19.83 & 21.76 \\
&\fvp & 23.72 & 24.11 & 19.51 & 23.31\\
&\ipt & 20.61 & 28.16 & 27.26 & 27.28\\
&\milo & 14.97 & 23.39 & 22.10 & 19.62\\
&\ours & \textbf{31.28} & 36.64 & \textbf{31.56} & \textbf{29.13}\\
&\ours-sem & 31.03 & \textbf{36.69} & 30.82 & 28.79\\
\bottomrule
\end{tabular}
}
\caption{
\textbf{Quantitative results (PSNR) of view synthesis and relighting on synthetic scenes}
show our estimation yields very consistent rendering under novel views and lighting.
View synthesis is unavailable for Li22.
}
\label{tab:synthetic-relight}
\end{table}

\begin{figure*}[ht!]
    \centering
    \setlength{\tabcolsep}{0.5pt}
    \resizebox{0.99\linewidth}{!}{
    \begin{tabular}{ccc @{\hskip 0.04in}|@{\hskip 0.04in} ccc @{\hskip 0.04in}|@{\hskip 0.04in} c}
    \milo & \lieccv & \ours  &
        \milo & \lieccv & \ours  & Input views\\
        
    \includegraphics[width=.15\linewidth]{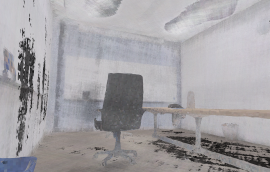}&
    \includegraphics[width=.15\linewidth]{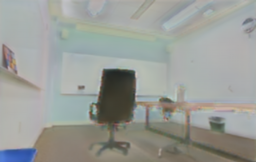}&
    \includegraphics[width=.15\linewidth]{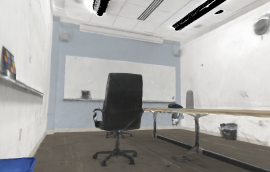}&

    \includegraphics[width=.15\linewidth]{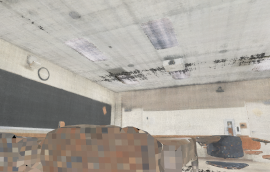}&
    \includegraphics[width=.15\linewidth]{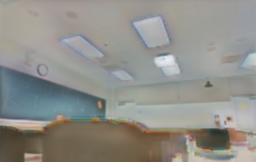}&
    \includegraphics[width=.15\linewidth]{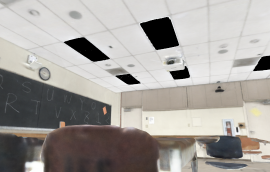}& 

    \includegraphics[width=0.15\linewidth]{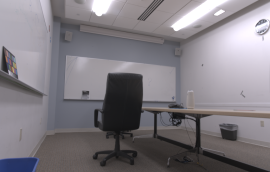}\\[-0.65ex]

    \includegraphics[width=.15\linewidth]{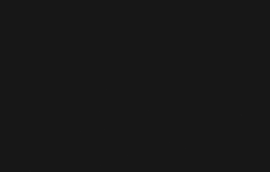}&
    \includegraphics[width=.15\linewidth]{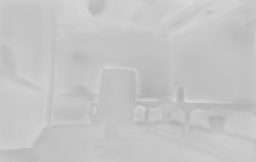}&
    \includegraphics[width=.15\linewidth]{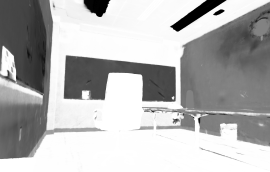}&

    \includegraphics[width=.15\linewidth]{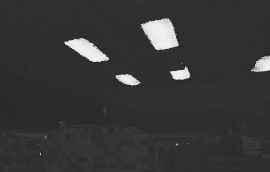}&
    \includegraphics[width=.15\linewidth]{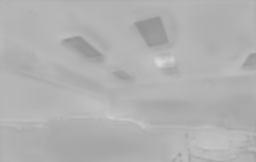}&
    \includegraphics[width=.15\linewidth]{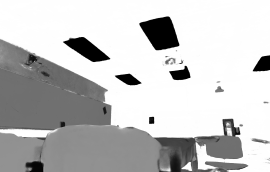}&
    
    \includegraphics[width=0.15\linewidth]{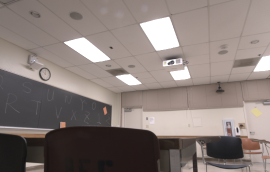}\\[-0.65ex]

    \includegraphics[width=.15\linewidth]{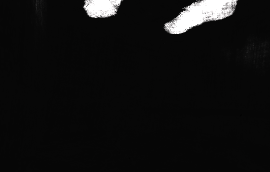}&
    \includegraphics[width=.15\linewidth]{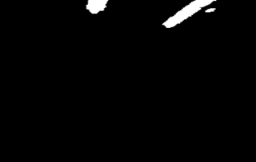}&
    \includegraphics[width=.15\linewidth]{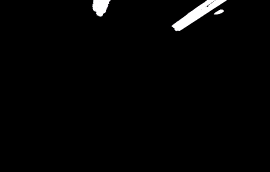}&

    \includegraphics[width=.15\linewidth]{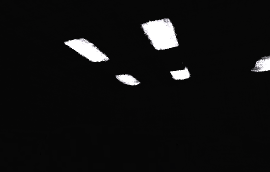}&
    \includegraphics[width=.15\linewidth]{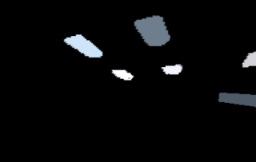}&
    \includegraphics[width=.15\linewidth]{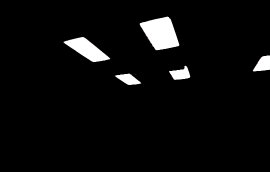}
    
    \end{tabular}
    }
    \caption{
    \textbf{BRDF and emission estimation results on 2 real world scenes}
    demonstrate our method gives reasonable estimation of BRDF and emission.
    The albedo and roughness preserve details without noticeable artifacts (row 1-2),
    and the emitters are correctly identified (3rd row).
    }
    \label{fig:real-world}
\end{figure*}
\begin{figure*}[hbt!]
    \centering
    \setlength{\tabcolsep}{0.5pt}
    \resizebox{0.99\linewidth}{!}{
    \begin{tabular}{ccccc @{\hskip 0.04in}|@{\hskip 0.04in} cccc}
        \multicolumn{5}{c}{\textbf{Rerendering}}&
        \multicolumn{4}{c}{\textbf{Relighting}}\\
         \milo & \lieccv & \fvp & \ours & Ground truth&
         \milo & \lieccv & \fvp & \ours\\
         
        \includegraphics[width=0.12\linewidth]{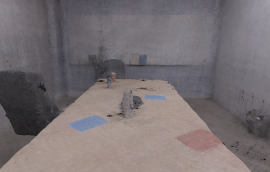}&
         \includegraphics[width=0.12\linewidth]{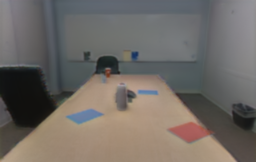}&
         \includegraphics[width=0.12\linewidth]{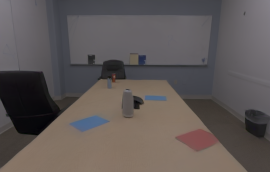}&
         \includegraphics[width=0.12\linewidth]{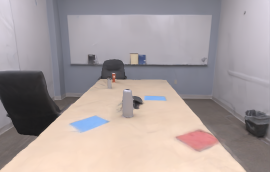}&
         \includegraphics[width=0.12\linewidth]{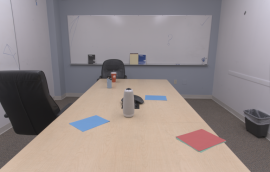}&

        \includegraphics[width=0.12\linewidth]{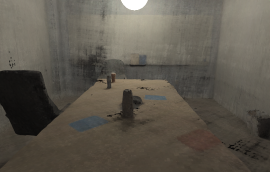}&
         \includegraphics[width=0.12\linewidth]{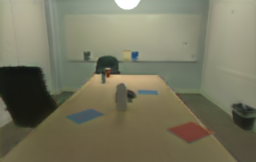}&
         \includegraphics[width=0.12\linewidth]{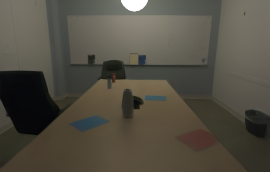}&
         \includegraphics[width=0.12\linewidth]{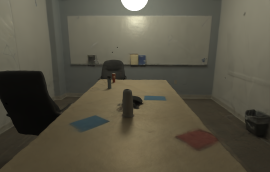}\\[-0.7ex]         
         
         \includegraphics[width=0.12\linewidth]{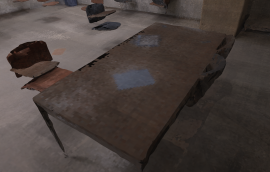}&
         \includegraphics[width=0.12\linewidth]{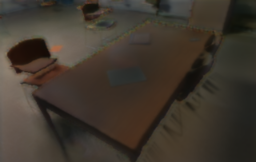}&
         \includegraphics[width=0.12\linewidth]{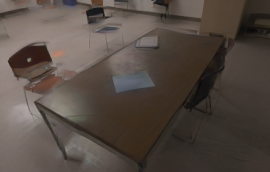}&
         \includegraphics[width=0.12\linewidth]{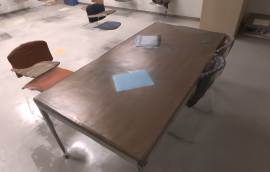}&
         \includegraphics[width=0.12\linewidth]{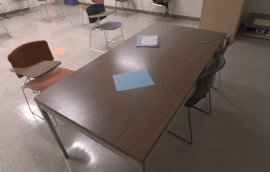}&

         \includegraphics[width=0.12\linewidth]{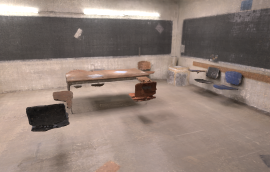}&
         \includegraphics[width=0.12\linewidth]{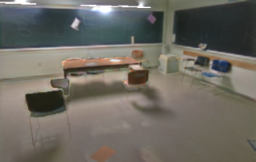}&
         \includegraphics[width=0.12\linewidth]{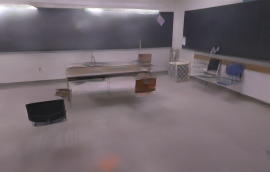}&
         \includegraphics[width=0.12\linewidth]{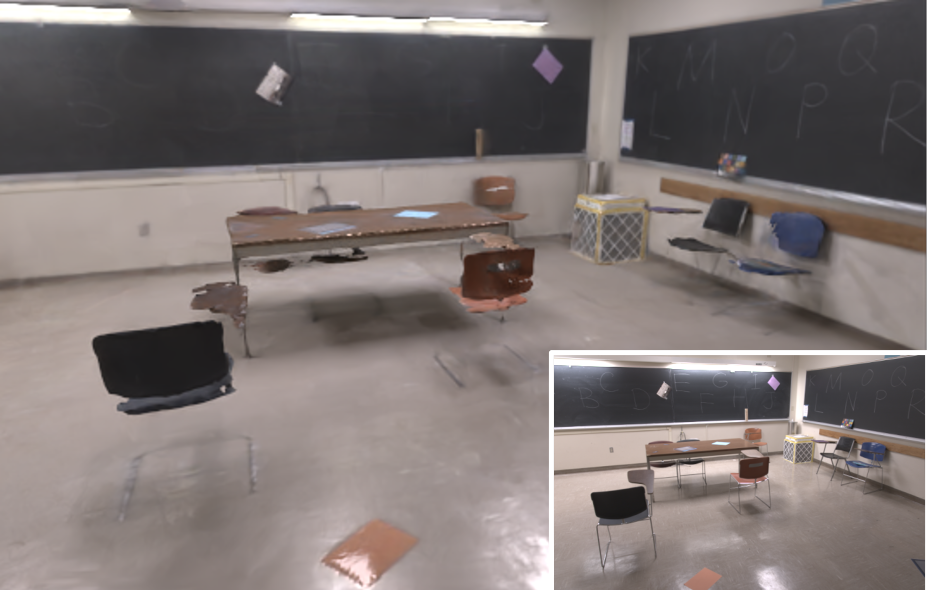}
    \end{tabular}
    }
    \caption{
    \textbf{Rerendering and relighting on 2 real world scenes}
    show our estimation fits the ground truth well (column 1-5)
    and gives good rendering under novel light (column 6-9).
    The inset in lower-right sub-figure shows the reference relighting for the Classroom  scene.
    }
    \label{fig:real-world-relight}
\end{figure*}

\vspace{-8pt}
\paragraph{Results.}
As is shown in Tab.~\ref{tab:synthetic-relight},
our (\ours~and \ours-sem) estimated BRDF-emission gives the most accurate view synthesis and relighting results.
While results from FVP are seemingly visually appealing, the method is not guaranteed to be physically plausible and fails to match the ground truth.
As shown in Fig.~\ref{fig:synthetic_render}, 
our method handles specular reflections and even mirror reflection well, 
which is difficult to be modeled by standard inverse path tracing owing to its high variance.

\subsection{Real-world scenes}
\label{subsec:realworld}
\vspace{-10pt}
Considering that it is not possible to obtain ground truth BRDF and emission from just RGB captures,
we only showcase qualitative comparison with MILO, FVP, and Li22 in terms of material-lighting estimation, rerendering, and relighting.
Only semantic segmentation is used for the real scenes
as ground truth part segmentation is unavailable.

\vspace{-8pt}
\paragraph{Results.}
As is shown in Fig.~\ref{fig:real-world}, our method visually produces more reasonable material reflectance and roughness with emission masks closer to the actual emitters.
The rerendering and relighting in Fig.~\ref{fig:real-world-relight} further serve as indirect measurements of the reconstruction quality,
where our method is capable of reconstructing the reflection of the emitters on the Conference room wall and the Classroom table,
and our relighting is visually closer to reference photos (obtained by switching lights; see supplementary) than the baselines
with good reproduction of specular highlights and shadows. 

\subsection{Ablation study}
\label{subsec:ablation}
\paragraph{Training strategy.}
Tab.~\ref{tab:training-strategy} shows the effect of different training strategies on the kitchen scene.
If we jointly optimize the emission and BRDF with the regularization term in IPT~\cite{azinovic2019inverse}, 
the BRDF optimization can still converge,
but the emission estimation does not converge given the same amount of time (2 epochs).
Since the majority of the scene receives incident light from nearby diffuse surfaces, 
the reconstruction result is still reasonable without shading refinement (stage 3),
but further refining the BRDF estimation helps to correct light transport for specular surfaces.
If we simply path-trace the BRDF-emission without using the radiance cache from stage 1, the refined shadings will accumulate too much estimation error causing the subsequent BRDF estimation to deviate from ground truth. 

\begin{table}[t]
\centering
\setlength\tabcolsep{3.0pt}
\resizebox{0.9\linewidth}{!}{
\begin{tabular}{lccccc}
\toprule
\multirow{2}{*}{Training strategy} & 
$\mathbf{k}_d$ & $\mathbf{a}'$ & $\sigma$ &
\multicolumn{2}{c}{$\mathbf{L}_e$} \\
&\multicolumn{3}{c}{PSNR$\uparrow$} & 
IoU$\uparrow$ & logL2$\downarrow$\\
\midrule
Joint $\mathbf{L}_e$ opt.
& 32.31 & 25.50 & 22.77 & 0.10 & 6.62e-1
\\
w/o stage 3
& 32.20 & 25.18 & 23.37 & \textbf{0.91} & \textbf{1.54e-3}
\\
w/o rad. cache
& 19.35 & 16.82 & 25.65 & \textbf{0.91} & \textbf{1.54e-3}
\\
Full model
& \textbf{33.07} & \textbf{27.53} & \textbf{29.24} & \textbf{0.91} & \textbf{1.54e-3}
\\
\bottomrule
\end{tabular}
}
\caption{
\textbf{Ablation study on training strategy} shows joint BRDF-emission optimization can lead to slow convergence of the emission,
and shading refinement helps further improve the reconstruction quality.
}
\label{tab:training-strategy}
\end{table}
\begin{figure}[t]
    \centering
    \setlength\tabcolsep{0.5pt}
    \resizebox{0.99\linewidth}{!}{
    \begin{tabular}{cccc}
    \multicolumn{2}{c}{Segmentation artifacts} & \multicolumn{2}{c}{Geometry artifacts}\\
    \includegraphics[width=0.3\linewidth]{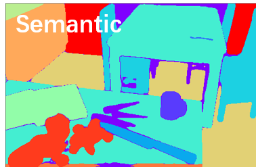}&
    \includegraphics[width=0.3\linewidth]{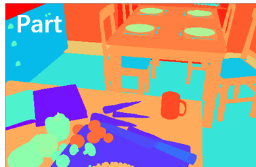}&
    \includegraphics[width=0.3\linewidth]{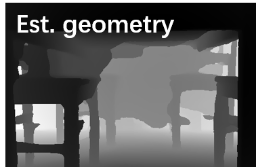}&
    \includegraphics[width=0.3\linewidth]{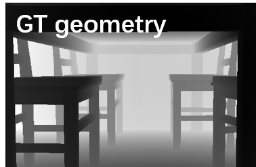}\\[-0.75ex]
    \end{tabular}
    }
    \caption{
    \textbf{Example of input noise} 
    introduced by semantic segmentation and estimated geometry.
    }
    \label{fig:different-masks}
\end{figure}
\begin{table}[t]
\centering
\setlength\tabcolsep{3.0pt}
\resizebox{0.99\linewidth}{!}{
\begin{tabular}{lcccccc}
\toprule
\multirow{2}{*}{Training input} & 
$\mathbf{k}_d$ & $\mathbf{a}'$ & $\sigma$ &
\multicolumn{2}{c}{$\mathbf{L}_e$} 
& Relight\\
&\multicolumn{3}{c}{PSNR$\uparrow$} & 
IoU$\uparrow$ & logL2$\downarrow$ 
& PSNR$\downarrow$\\
\midrule
Fewer views&
32.64 & 26.7 & 28.80 & \textbf{0.91} & 1.57e-3 & 
28.90\\
Est. geometry &
27.01 & 22.57 & 21.33 & 0.78 & 0.1142 &
27.90\\
Semantic seg. &
\textbf{33.25} & 27.38 & 21.70 & \textbf{0.91} & \textbf{1.54e-3} & 
28.79\\
Part seg. &
33.07 & \textbf{27.53} & \textbf{29.24} & \textbf{0.91} & \textbf{1.54e-3} & 
\textbf{29.13}\\
\bottomrule
\end{tabular}
}
\caption{
\textbf{Ablation study on training input} 
shows our method gives similar performance with fewer training images.
If the input geometry or the segmentation is not perfect (semantic segmentation),
the reconstruction quality downgrades but can still give reasonable relighting.
}
\label{tab:training-input}
\end{table}
\vspace{-8pt}
\paragraph{Sensitive analysis on training inputs.}
We run our algorithm on the kitchen scene with different setups:
(1) using 60 training views instead of 200;
(2) using geometry from MonoSDF~\cite{Yu2022MonoSDF} instead of the ground truth;
(3) and replacing part segmentation by semantic segmentation.
As is shown in Tab.~\ref{tab:training-input}, training with fewer views shows almost equal quality as long as they cover most of the scene.
However, it still requires around 200 images in capturing to get good geometry reconstruction.
Both estimated geometry and semantic segmentation introduce inaccuracy for detailed objects (Fig.~\ref{fig:different-masks}),
which disrupts roughness estimation for regions that see weak highlights.
However, given the lighting for those regions is mostly ambiguous,
it does not affect the overall reconstruction quality or relighting too much.

\begin{figure}[t]
    \centering
    \setlength\tabcolsep{1.0pt}
    \resizebox{0.99\linewidth}{!}{
    \begin{tabular}{cc}
    Complex geometry &
    Environment light\\
    \includegraphics[width=0.6\linewidth]{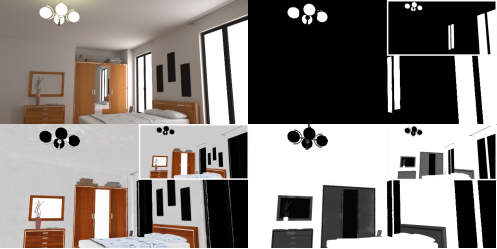}&
    \includegraphics[width=0.6\linewidth]{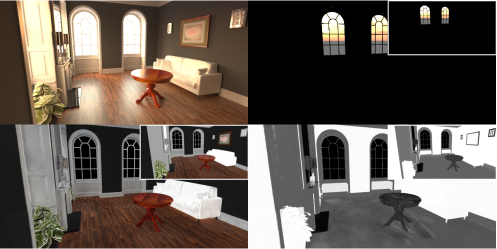}
    \end{tabular}
    }
    \caption{\textbf{Reconstruction with complex emitters.}
    Our method can also correctly identify complex lamps (first 2 columns) or environment lighting (last 2 columns) with reasonable BRDF reconstruction.
    Ground truth is shown in the insets.
    }
    \label{fig:complex-emitter}
\end{figure}
\vspace{-8pt}
\paragraph{Complex emitters.} 
Our method also works with 
lamp-like emitters with complex geometry (Fig.~\ref{fig:complex-emitter}).
For windows with hollow geometry, 
we can model the window light as a directional light source
and let rays that fail to hit any indoor geometry to query an outdoor environment map.
Each pixel of the environment map is estimated similarly as the emission of an emitter triangle.
While environment lighting may not be fully observed and consequently causes artifacts near windows (see Sec.~\ref{sec:conclution}),
a majority of the surfaces can still be reconstructed well owing to the diffuse radiance cache.
\section{Limitations and Future Work}
\label{sec:conclution}
\begin{figure}
    \centering
    \setlength\tabcolsep{1.0pt}
    \resizebox{0.99\linewidth}{!}{
    \begin{tabular}{cccc}
    \multicolumn{2}{c}{Bad input geometry}&
    \multicolumn{2}{c}{Insufficient emission observation}\\
    \includegraphics[width=0.3\linewidth]{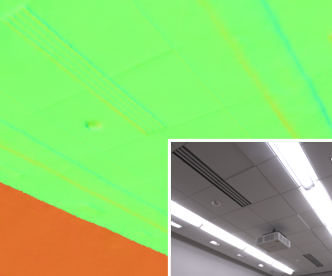}&
    \includegraphics[width=0.3\linewidth]{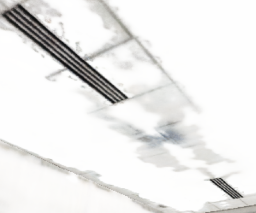}&
    \includegraphics[width=0.3\linewidth]{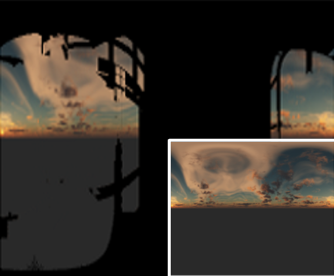}&
    \includegraphics[width=0.3\linewidth]{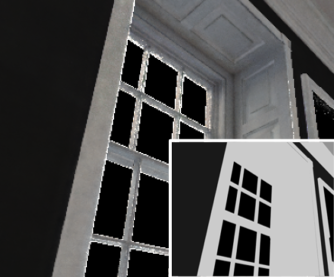}
    \end{tabular}
    }
    \caption{\textbf{Limitations.}
    Bad geometry estimation of an emitter (1st image) will lead to incorrect BRDF reconstruction of its nearby regions (2nd image). 
    Incomplete observation of the environment light (3rd image) can cause artifacts in BRDF estimation (4th image).
    }
    \label{fig:limitation}
\end{figure}
Our method shares certain limitations with standard inverse path tracing.
The framework does not optimize geometry,
so that the BRDF and emission estimations can be inaccurate if the input geometry (especially for emitters) is extremely bad (Fig~\ref{fig:limitation}, left).
Combining differentiable geometry optimization~\cite{Bangaru2022DifferentiableRO,vicini2022differentiable}
may help improve the robustness.
Meanwhile, the BRDF estimation fails if the dominant light source (\eg the sun) is not directly observed,
which can happen very often with environment emitters whose observations are blocked by the windows (Fig.~\ref{fig:limitation}, right).
Incorporating learning based methods may help.
Lastly, the optimization relies on photometric observations,
which means it cannot remove ambient occlusion effects out of the BRDF maps (as radiance there is near zero) and our model does not model transparent objects. 
\paragraph{Acknowledgements.}
This work was supported in part by NSF grants 1751365, 2100237, 2105806, 2110409, 2120019, 2127544, ONR
grant N000142012529, N000142312526, a Sony research Award, gifts from Qualcomm, Adobe and Google, the Ronald
L. Graham Chair and the UC San Diego Center for Visual Computing.

Additionally, we would like to thank Bohan Yu, Dejan Azinovic, 
Julien Philip, and Zhengqin Li for generous assistance in evaluation of their methods, David Forsyth, Shenlong Wang, and Merlin Nimier-David for insightful discussions,
as well as Jiaer Zhang for assistance in implementation.

{\small
\bibliographystyle{ieee_fullname}
\bibliography{egbib}
}
\newpage
\begin{appendix}
\vspace{20pt}

\section*{Supplementary Material Overview}
In Sec.~\ref{sec:implementation-details},
we provide the details of our pipeline implementation and data pre-processing.

In Sec.~\ref{sec:experiment-details}, we present additional details of our experiments,
including: 
(1) the setup of real world scenes (Sec.~\ref{subsec:real-world-scene});
(2) detailed ablation study (Sec.~\ref{subsec:ablation-details});
(2) additional results (Sec.~\ref{subsec:synthetic-scene});
and (4) the schemes for evaluating the baselines (Sec.~\ref{supp:baselines}).
\begin{figure}
    \centering
    \setlength\tabcolsep{2.0pt}
    \resizebox{0.95\linewidth}{!}{
        \begin{tabular}{ccc}
             \includegraphics[width=0.35\linewidth]{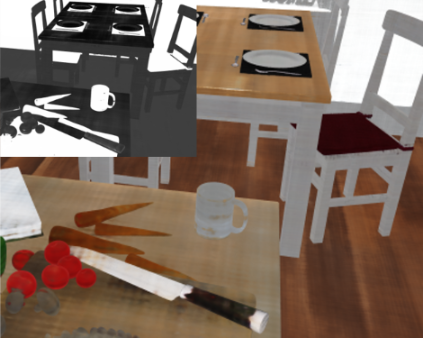}&
             \includegraphics[width=0.35\linewidth]{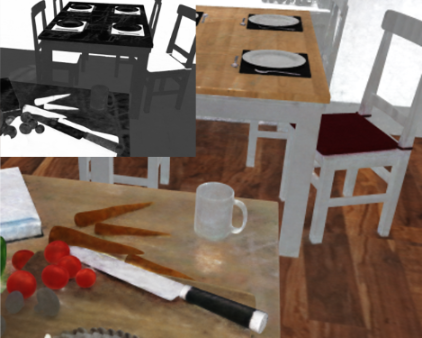}&
             \includegraphics[width=0.35\linewidth]{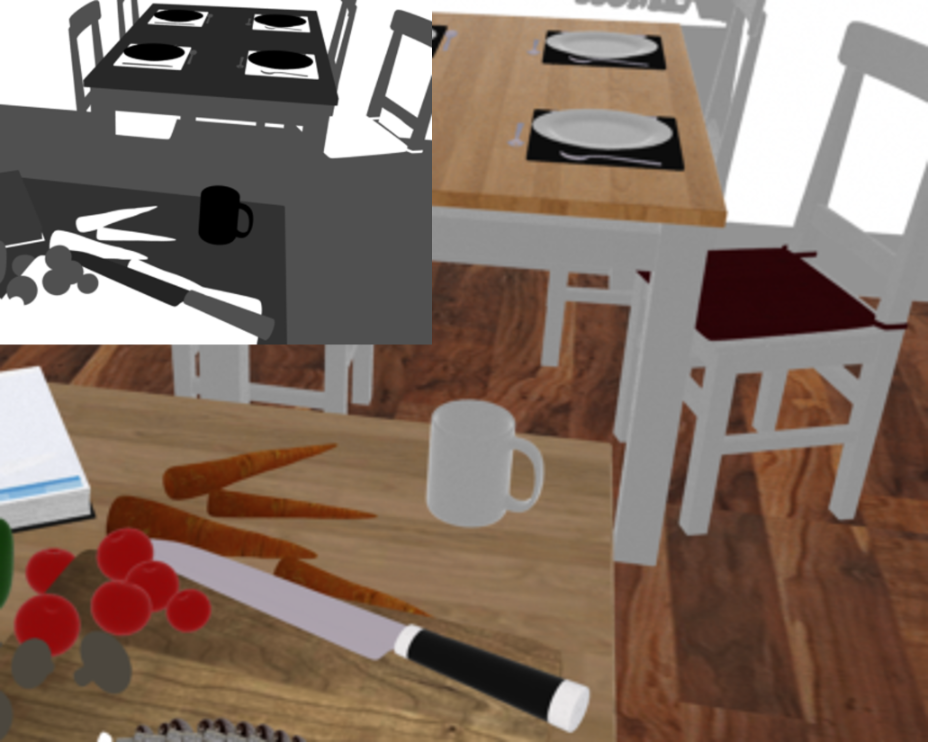}\\
             Positional encoding &
             Hash grid & Ground truth
        \end{tabular}
    }
    \caption{\textbf{Qualitative comparison of different input encoding}
    shows a hash grid can better model the detailed texture on the floor.}
    \label{fig:network-ablation}
\end{figure}
\begin{figure}
    \centering
    \setlength\tabcolsep{0.5pt}
\resizebox{0.99\linewidth}{!}{
    \begin{tabular}{cccc}
    \multicolumn{2}{c}{Origin semantic segmentation} & \multicolumn{2}{c}{Fused semantic segmentation}\\
    \includegraphics[width=0.28\linewidth]{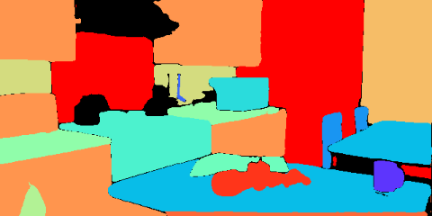}&
    \includegraphics[width=0.28\linewidth]{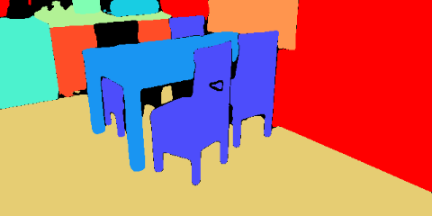}&
    \includegraphics[width=0.28\linewidth]{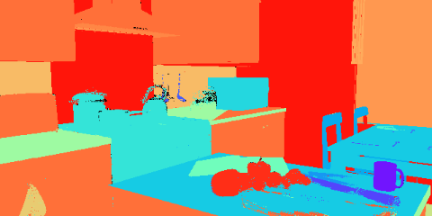}&
    \includegraphics[width=0.28\linewidth]{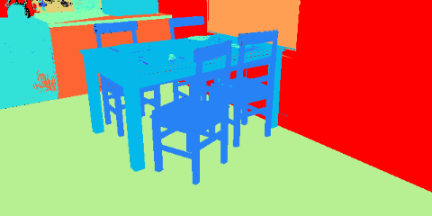}
    \end{tabular}
    }
    \caption{\textbf{Fusing segmentation on raw images} 
    onto the mesh produces multi-view consistent segmentation.
    }
    \label{fig:segmentation-fused}
\end{figure}
\begin{figure*}[hbt!]
    \centering
    \setlength\tabcolsep{1.0pt}
    \resizebox{0.99\linewidth}{!}{
    \begin{tabular}{cc}
         \begin{tabular}{c}
         \includegraphics[width=0.18\linewidth]{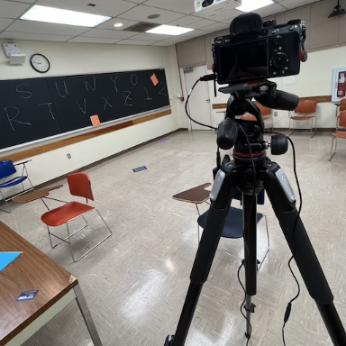}
         \end{tabular}
         &
         \begin{tabular}{|@{\hskip 0.05in}ccc @{\hskip 0.05in}|@{\hskip 0.05in}ccc}
         \includegraphics[width=0.13\linewidth]{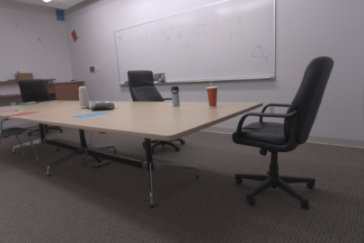}
         &
         \includegraphics[width=0.13\linewidth]{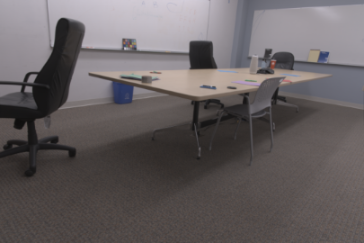}
         &
         \includegraphics[width=0.13\linewidth]{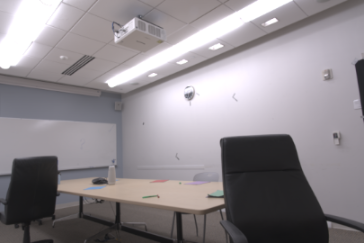}
         & 
         \includegraphics[width=0.13\linewidth]{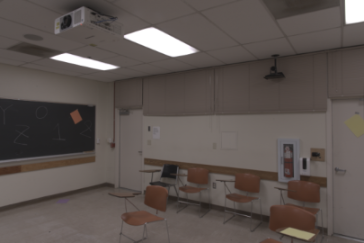}
         &
         \includegraphics[width=0.13\linewidth]{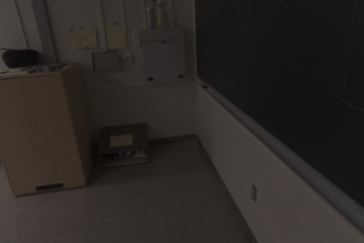}
         &
         \includegraphics[width=0.13\linewidth]{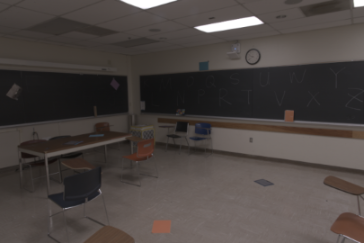}
         \\

         \includegraphics[width=0.13\linewidth]{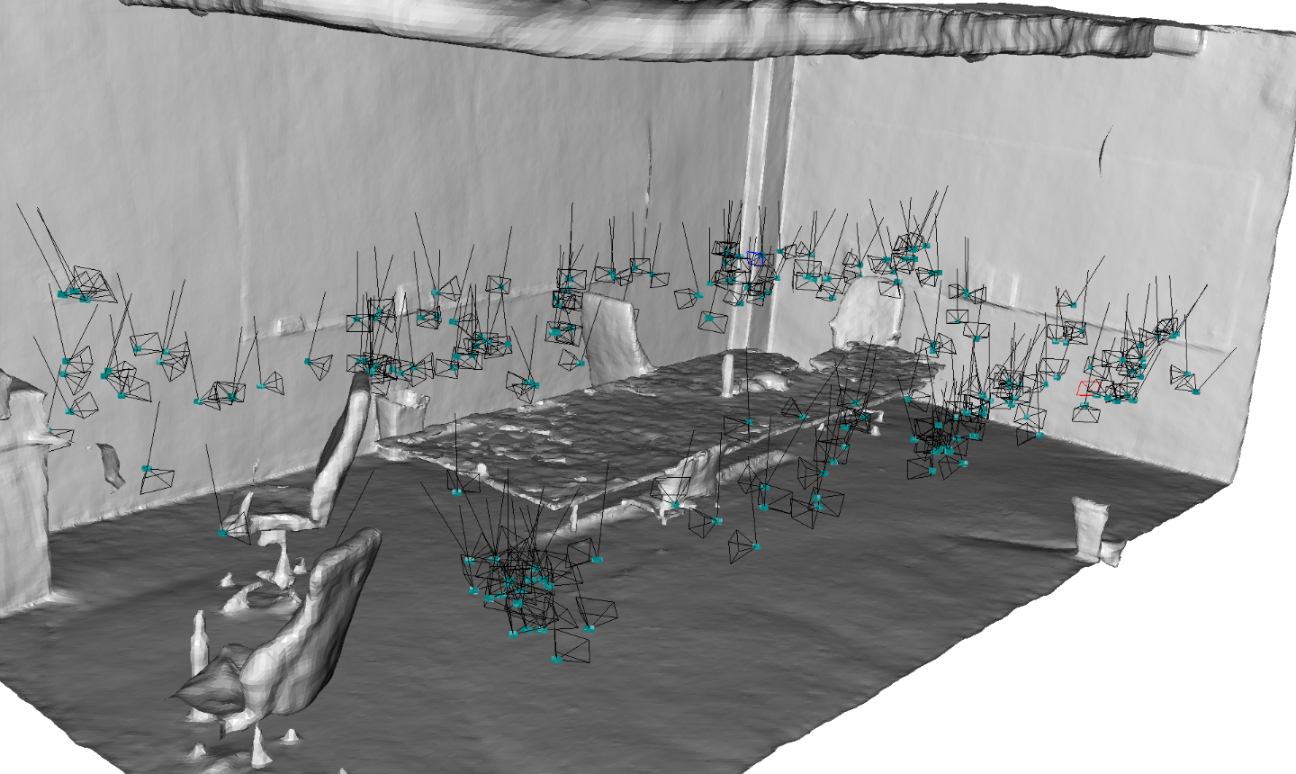}
         &
         \includegraphics[width=0.13\linewidth]{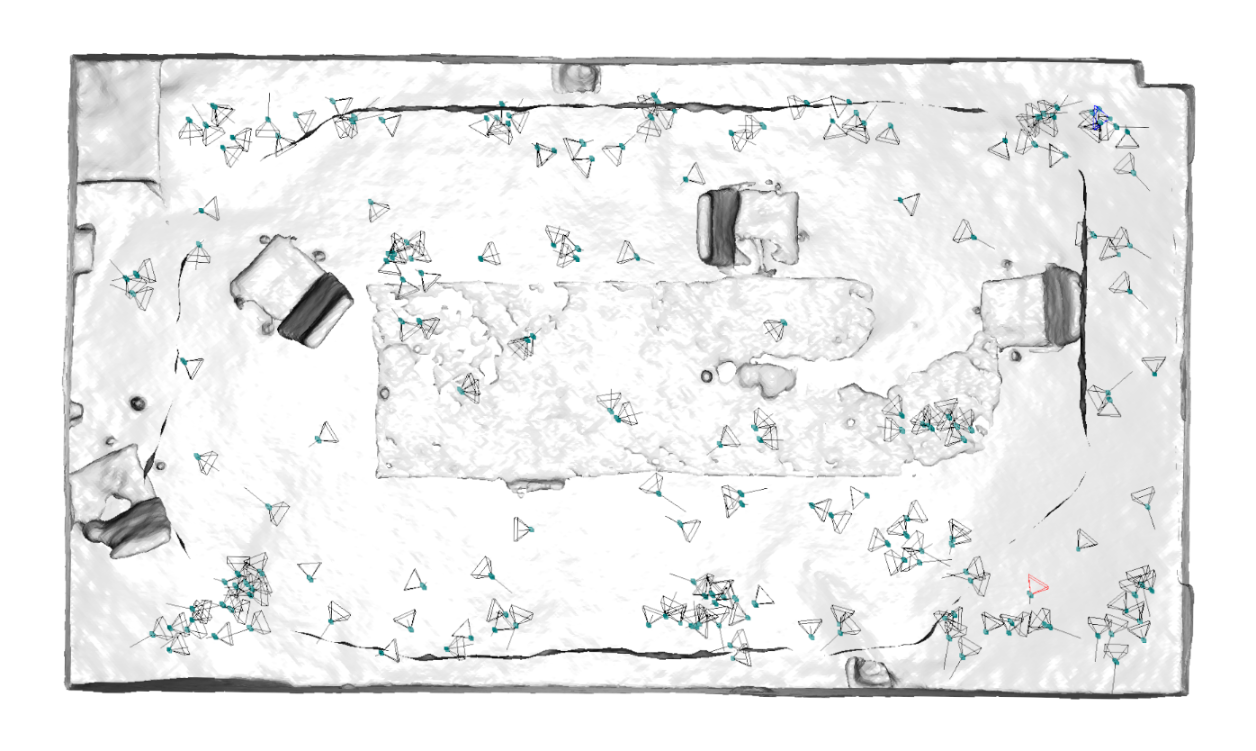}
         &
         \includegraphics[width=0.13\linewidth]{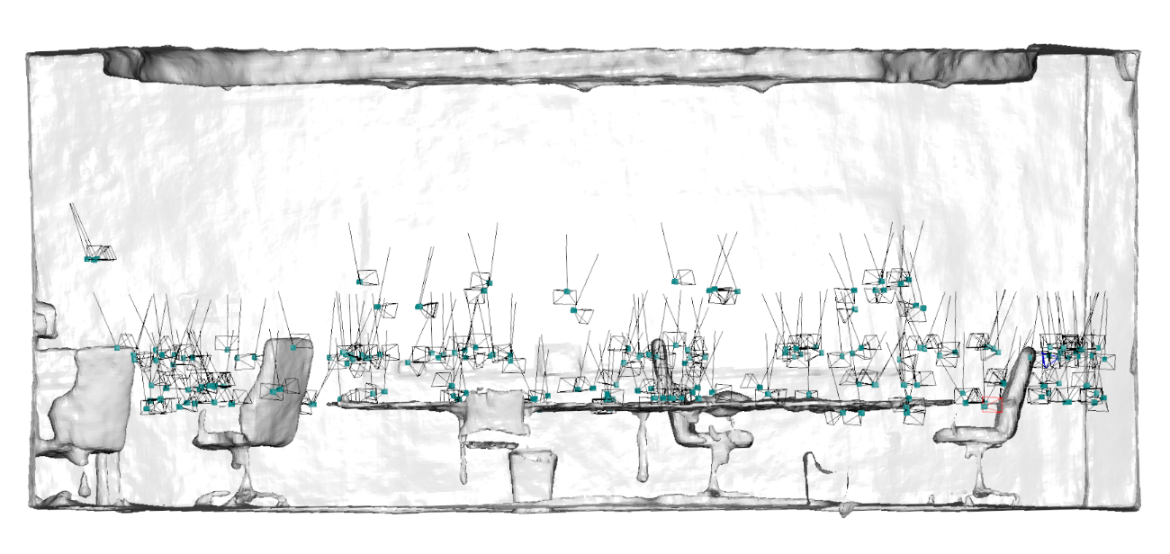} 
         &
         \includegraphics[width=0.13\linewidth]{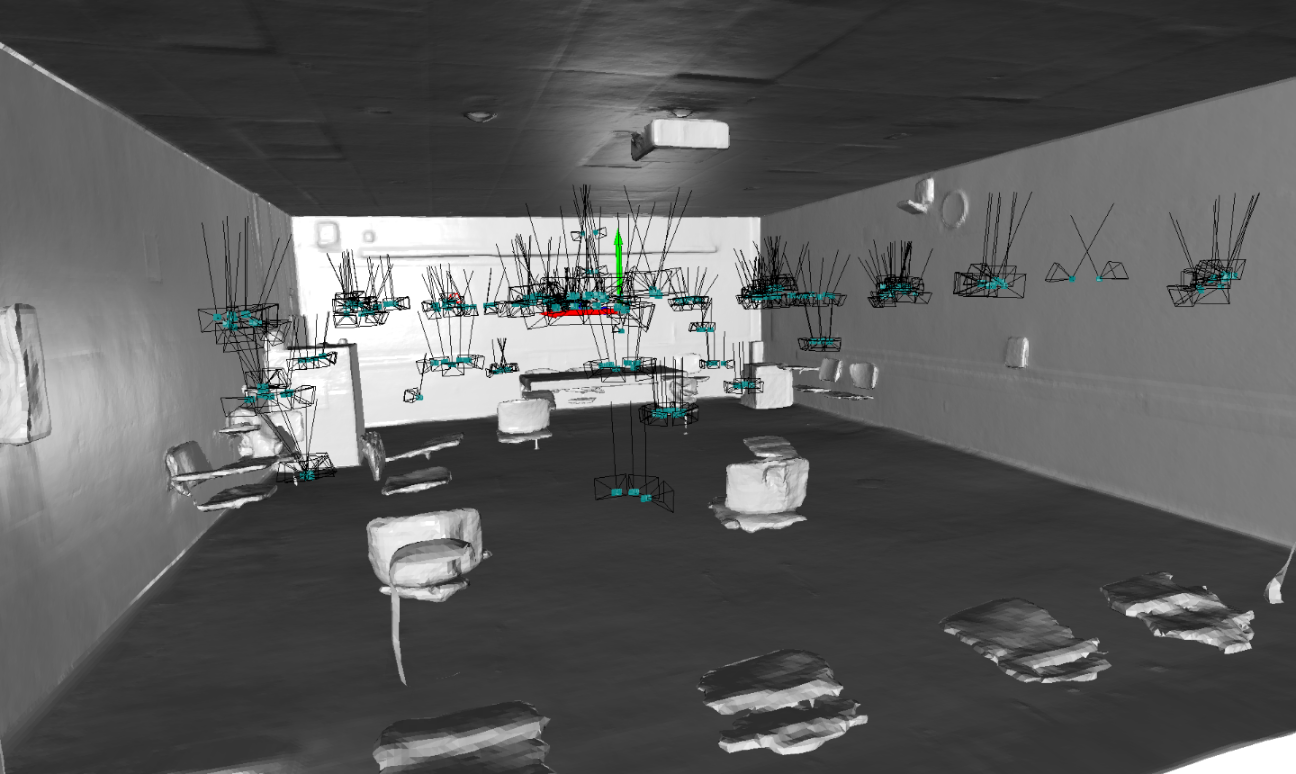}
         &
         \includegraphics[width=0.13\linewidth]{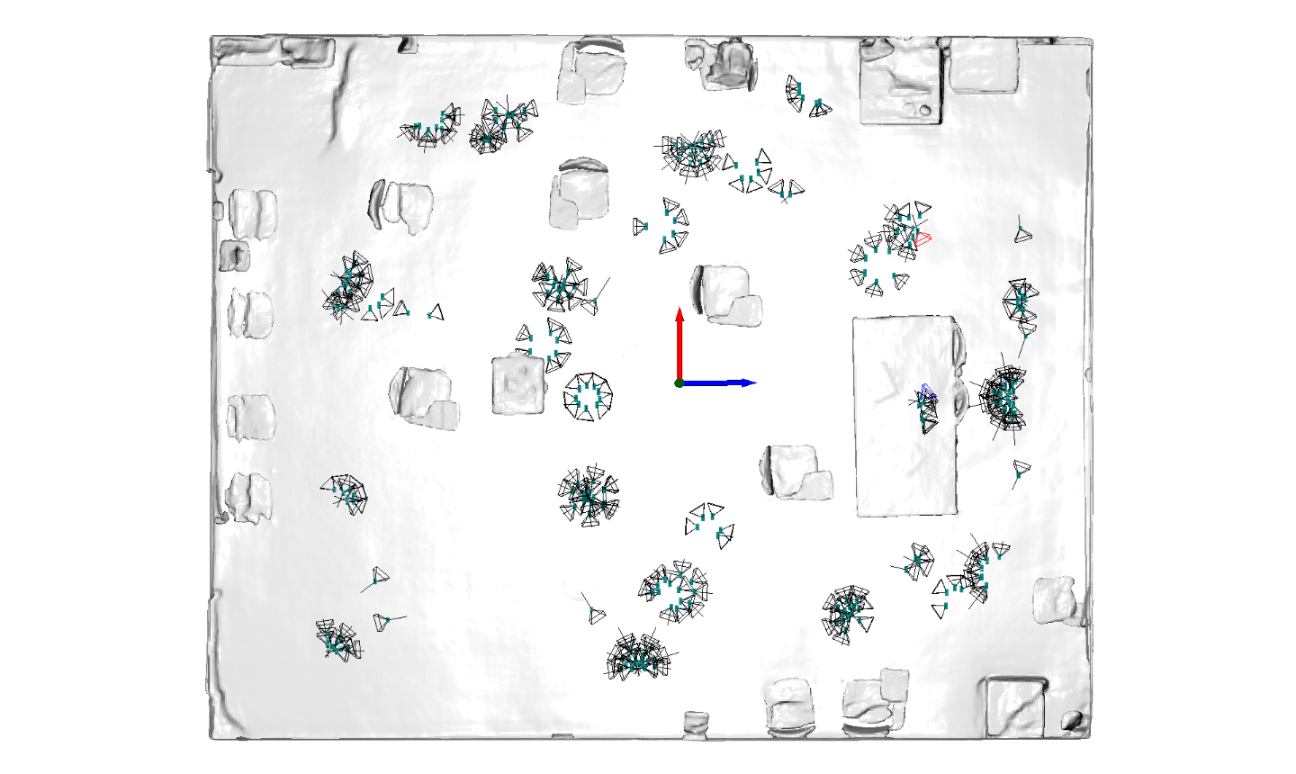}
         &
         \includegraphics[width=0.13\linewidth]{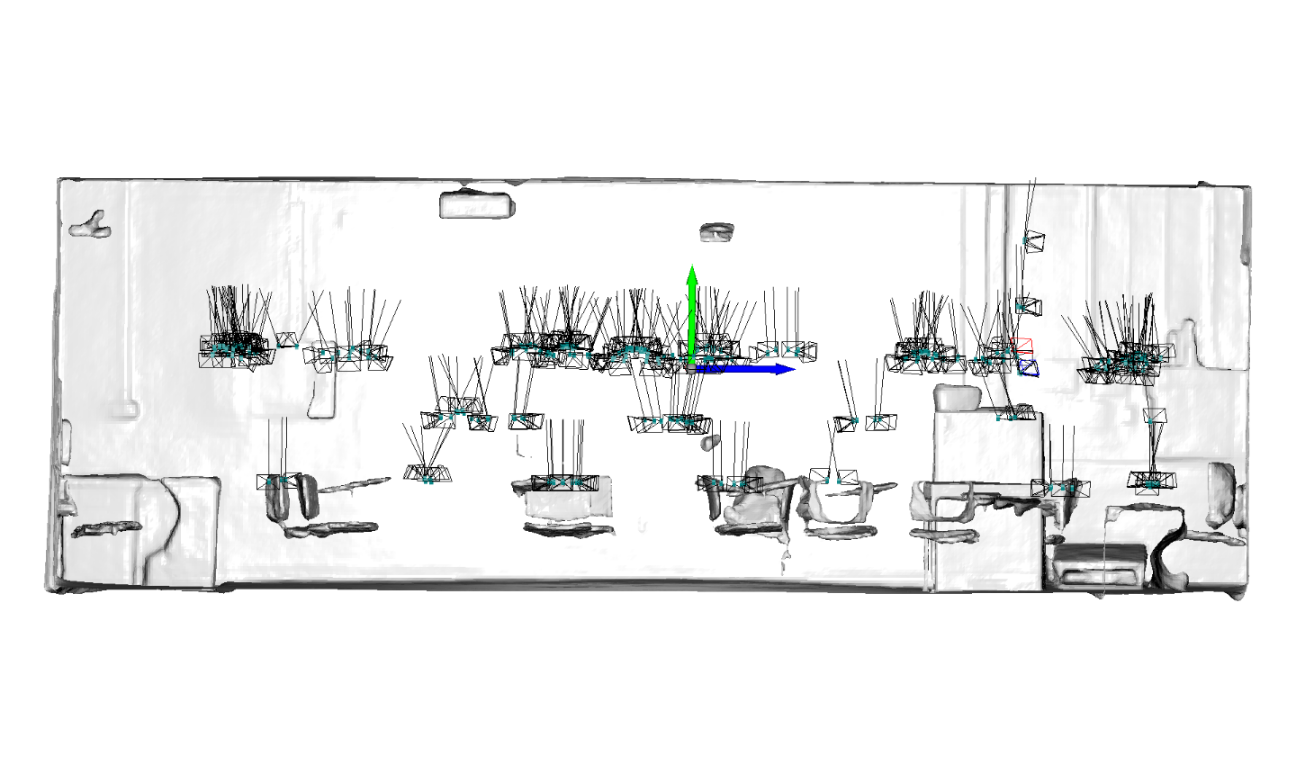} 
        \\
         \multicolumn{3}{c}{\textsf{Conference room}} & \multicolumn{3}{c}{\textsf{Classroom}} \\
         \end{tabular}
    \end{tabular}
    }
    \caption{\textbf{The capture setting (left) and observations of the real world scenes (middle and right).}
    We present two real world scenes (\textsf{Conference room} and \textsf{Classroom}) with samples of captured images, reconstructed geometries in 3 views, and all camera poses.}
    \label{fig:capturing}
\end{figure*}

\section{Implementation Details}
\label{sec:implementation-details}
We implement our method in PyTorch~\cite{paszke2019pytorch} and Mitsuba 3~\cite{jakob2022mitsuba3}.
The diffuse and specular shadings in Eq.~6 are path-traced and denoised by the OptiX denoiser~\cite{parker2010optix},
where we use 128 samples per pixel for diffuse shadings and 64 for specular shadings.
Importance sampling of the BRDF is applied for shading initialization (stage 1),
and multiple importance sampling is applied for shading refinement (stage 3).
For each round of BRDF-emission mask estimation (stage 2), 
the optimization is run over the entire training set for 2 epochs using Adam~\cite{kingma2014adam} optimizer
with a learning rate of 1e-3 and a batch size of 8,192.
Stage 2 and 3 are repeated twice after stage 1,
and all the experiments are run on a single 3090Ti GPU.

\vspace{-8pt}
\paragraph{Network architecture.}
The BRDF network $\text{MLP}_\text{brdf}$ has 2 hidden layers of size 64,
and its hash encoding~\cite{muller2022instant} has 32 levels and 19 $\log_2$ hash map size
with other parameters set to their recommended defaults.
For emission mask network $\text{MLP}_\text{emit}$, we use positional encoding~\cite{mildenhall2020nerf} with 10 frequency bands,
6 hidden layers of size 128, and one residual connection in the middle.
Hash encoding is preferred for the BRDF network as albedo usually demonstrates high frequency pattern, which can be more efficiently modeled by a hash grid (Fig.~\ref{fig:network-ablation}).
Both networks use ReLU activation between the intermediate layers.

\vspace{-8pt}
\paragraph{Semantic segmentation acquisition.}
To obtain semantic segmentation, 
we use Mask2Former~\cite{cheng2021mask2former} pre-trained on the COCO dataset~\cite{lin2014microsoft} with Swin-L backbone. 
The input images are firstly tone-mapped with $\gamma=1/2.2$ then clipped to be in the range $[0,1]$.
Given segmentation from multi-view images, we fuse them onto the mesh and let each mesh triangle take the segmentation ID with the maximum occurrence (Fig.~\ref{fig:segmentation-fused}).

\vspace{-8pt}
\paragraph{Geometry acquisition with MonoSDF~\cite{Yu2022MonoSDF}.} 
We adapt the original code from MonoSDF in the default configuration for ScanNet 
with Multi-Resolutional Feature Grids architecture and the following changes: 
(1) instead of having all rays coming from one image in each training iteration,
we randomly sample over all training pixels,
which is empirically found to yield more stable convergence on noisy inputs especially for real world images;
(2) input images are changed from SDR to HDR to be in the same format as our model input;
accordingly, output activation of MLP is changed to ReLU, 
and re-rendering loss is changed to L1 loss on tone-mapped outputs and labels. 
Considering MonoSDF does not incorporate an outlier rejection algorithm,
we employ a two-step training strategy to deal with the bad camera poses.
We first train for one epoch to acquire a rough mesh
and reproject the mesh onto all frames.
Frames with significant misalignment are then rejected and the model is re-trained.
To extract the mesh, we employ Marching Cubes with a grid size of 512.
In total, the entire process takes around 1 day per-scene.
\section{Experiment Details}
\label{sec:experiment-details}

\subsection{Real world scene capture and relighting}
\label{subsec:real-world-scene}
\paragraph{Need for acquiring new real world data.}

Existing datasets that provide multi-view HDR images and camera poses of real world scenes
may include: Replica~\cite{straub2019replica}
, Matterport3D~\cite{chang2018matterport3d}, and sample scenes from \fvp.
However, each dataset has their own limitations that prohibit usage in our evaluation.
Specifically, HDR images from FVP do not employ exposure bracketing, 
which results in overexposed emission that is not applicable to our physically-based light transport modeling. 
HDR images from Replica are not publicly available, 
thus view-dependent effects cannot be observed.
For Matterport3D, the captured images exhibit artifacts including camera glare and problematic tone-mapping.

Therefore, we capture a few scenes as proof of concept of our method, including a conference room scene presented in the main paper and an additional classroom scene. Fig.~\ref{fig:capturing} demonstrates our capture setting. 
We mount a Sony A7M3 full-frame camera on a tripod
and use a remote control shutter release to capture images with exposure bracketing 
of 5 steps 1EV each or 5 steps 2EV each depending on the dynamic range of the room. 
We take images from multiple locations of the room, 
starting roughly with a direction towards the room center,
then randomizing yaw angles between $-60\degree$ to $60\degree$, 
pitch angles between $-45\degree$ to $45\degree$, 
with minimal roll. 
The camera height is sampled between $0.5m$ to $2.5m$.
For HDR reconstruction,
we process the captured RAW images with
black level subtraction, demosaicing, de-vignetting, and undistortion.
The recovered images are assumed to follow linear camera response
and are combined using a hat function similar to Debevec~\etal~\cite{debevec2008recovering}.

\begin{figure}
    \centering
    \begin{tabular}{c}
         \includegraphics[width=0.85\linewidth]{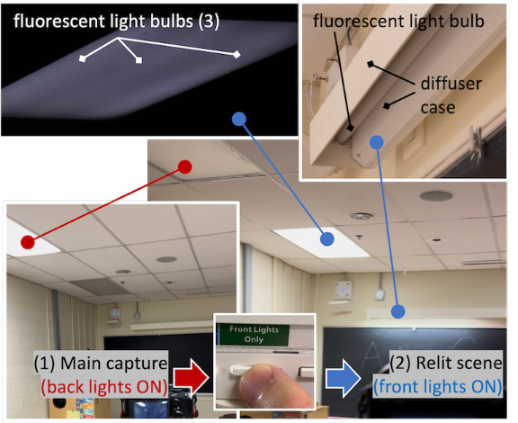}\\
         \includegraphics[width=0.85\linewidth]{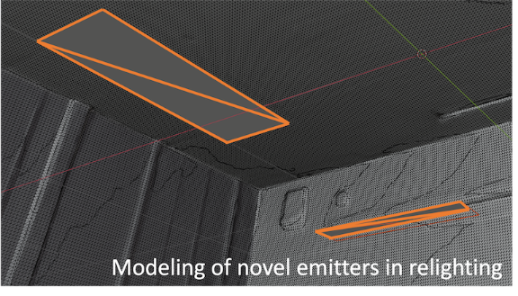} 
    \end{tabular}
    
    \caption{
    \textbf{Relighting of the Classroom scene.}
    The upper figure shows (1) the lighting for main captures with only back lights turned on,
    and (2) the relit scene with only front lights on (as reference for our relighting experiments).
    The lower figure shows our inserted area emitters as approximation of the actual front lights.
    }
    \label{fig:demo_emitter_relighting}
\end{figure}

\vspace{-8pt}
\paragraph{Reference relighting of Classroom.}
As is shown in Fig.~\ref{fig:demo_emitter_relighting}, 
lights in the Classroom can be switched between front and rear light modes. 
We choose the rear lights as original lighting for the main capture, 
and take a few additional photos with only front lights on as reference for relighting.
Given BRDF-emission estimation from the main capture,
we relight the scene by turning the estimated emission off and insert simple novel emitters to roughly match the front lights in their actual locations (see demonstration in Fig.~\ref{fig:demo_emitter_relighting}, bottom). 
Considering it is not possible to have the manually inserted novel emitters to perfectly match the actual complex front lights, we treat the reference relighting photos only as pseudo-ground truth.

\begin{figure}[t]
    \centering
    \setlength\tabcolsep{0.5pt}
\resizebox{0.99\linewidth}{!}{
    \begin{tabular}{cccccc}
    \includegraphics[width=0.2\linewidth]{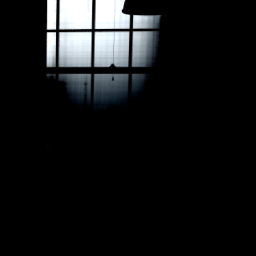}&
    \includegraphics[width=0.2\linewidth]{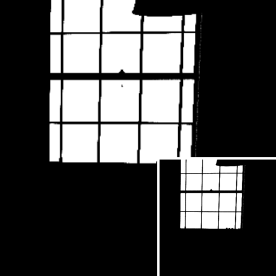}&
    \includegraphics[width=0.2\linewidth]{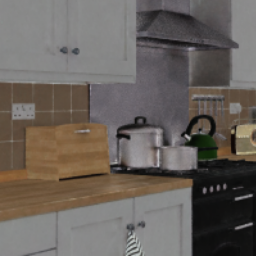}&
    \includegraphics[width=0.2\linewidth]{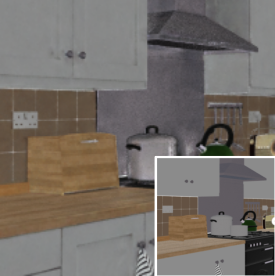}&
    \includegraphics[width=0.2\linewidth]{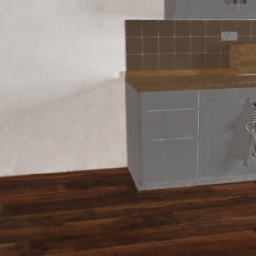}&
    \includegraphics[width=0.2\linewidth]{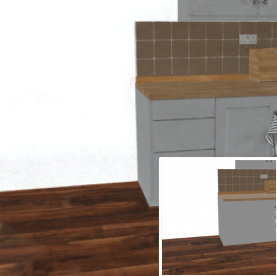}\\
    \multicolumn{2}{c}{Joint vs separate $\mathbf{L}_e$ opt.}&
    \multicolumn{2}{c}{w/o vs w/ stage 3} &
    \multicolumn{2}{c}{w/o vs w/ rad. cache}
    \end{tabular}
    }
    \caption{
    \textbf{Qualitative comparison of different training strategies}
    shows all of the strategies are necessary for efficient and accurate BRDF-emission estimation.
    The insets are the ground truth.
    }
    \label{fig:training-strategy}
\end{figure}
\begin{figure}[t]
    \centering
    \setlength\tabcolsep{0.25pt}
\resizebox{0.99\linewidth}{!}{
    \begin{tabular}{cccc}
    \multicolumn{2}{c}{Estimated geometry}&
    \multicolumn{2}{c}{Ground truth geometry}\\
    \includegraphics[width=0.28\linewidth]{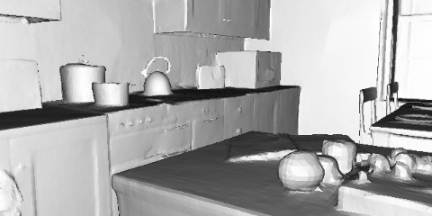}&
    \includegraphics[width=0.28\linewidth]{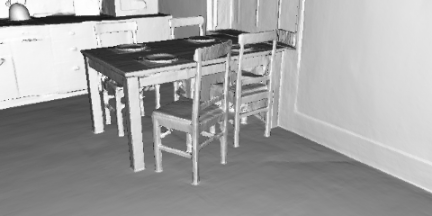}&
    \includegraphics[width=0.28\linewidth]{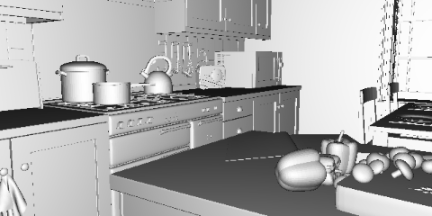}&
    \includegraphics[width=0.28\linewidth]{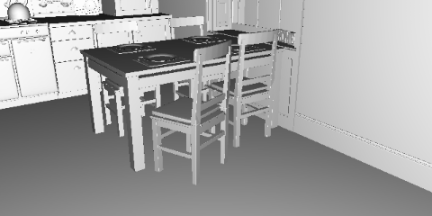}\\[-0.4ex]

    \multicolumn{2}{c}{Semantic segmentation}&
    \multicolumn{2}{c}{Part segmentation}\\

    \includegraphics[width=0.28\linewidth]{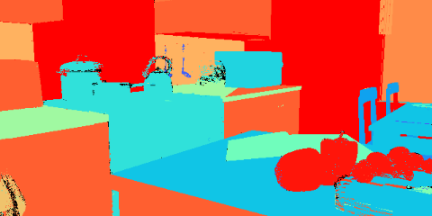}&
    \includegraphics[width=0.28\linewidth]{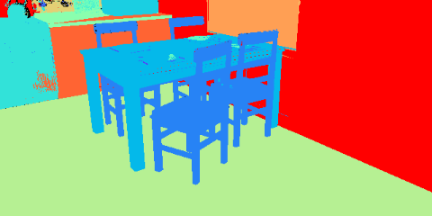}&
    \includegraphics[width=0.28\linewidth]{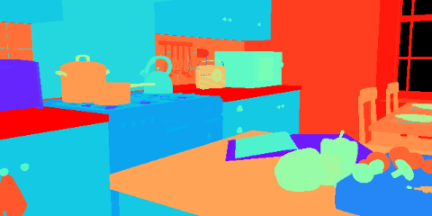}&
    \includegraphics[width=0.28\linewidth]{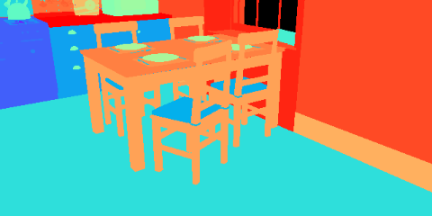}
    \\[-0.4ex]
    \multicolumn{4}{c}{w/ estimated geometry}\\
    \includegraphics[width=0.28\linewidth]{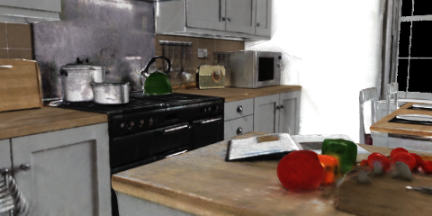}&
    \includegraphics[width=0.28\linewidth]{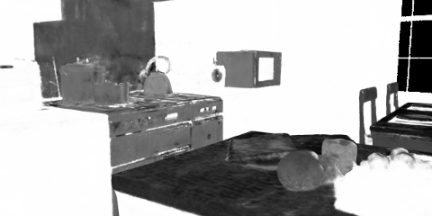}&
    \includegraphics[width=0.28\linewidth]{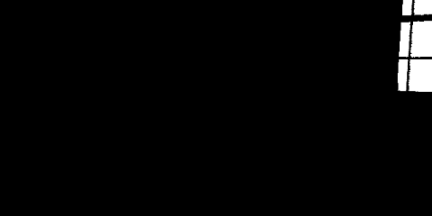}&
    \includegraphics[width=0.28\linewidth]{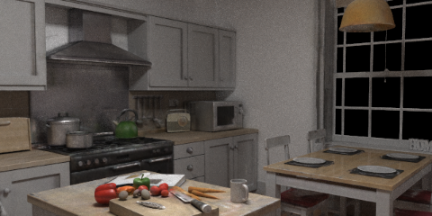}\\[-0.4ex]

    \multicolumn{4}{c}{w/ semantic segmentation}\\
    \includegraphics[width=0.28\linewidth]{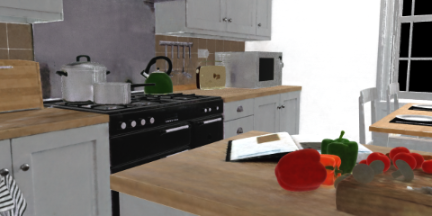}&
    \includegraphics[width=0.28\linewidth]{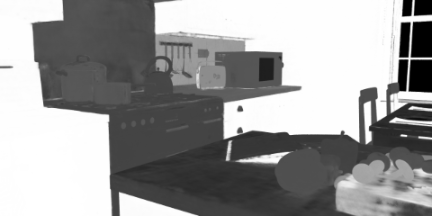}&
    \includegraphics[width=0.28\linewidth]{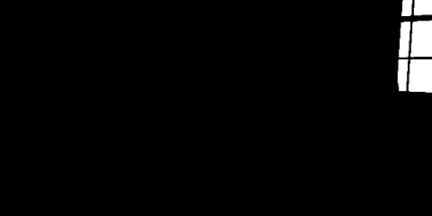}&
    \includegraphics[width=0.28\linewidth]{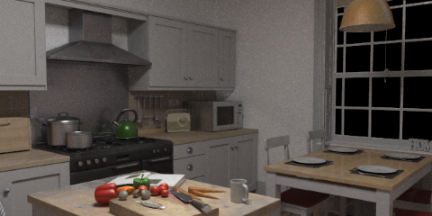}
    \\[-0.4ex]
    \multicolumn{4}{c}{w/ ground truth geometry and part segmentation}\\
    \includegraphics[width=0.28\linewidth]{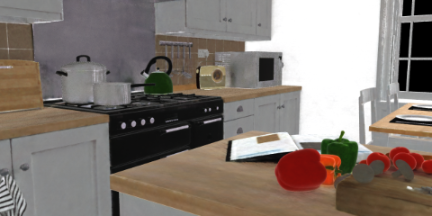}&
    \includegraphics[width=0.28\linewidth]{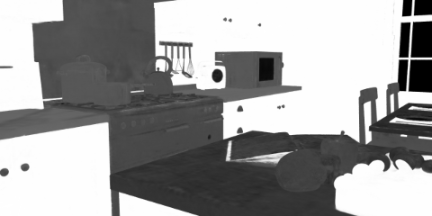}&
    \includegraphics[width=0.28\linewidth]{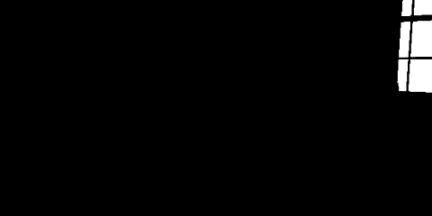}&
    \includegraphics[width=0.28\linewidth]{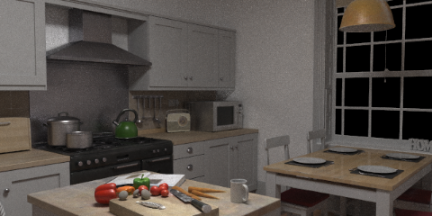}\\[-0.4ex]

    \multicolumn{4}{c}{Ground truth}\\
    \includegraphics[width=0.28\linewidth]{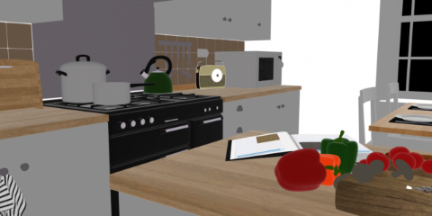}&
    \includegraphics[width=0.28\linewidth]{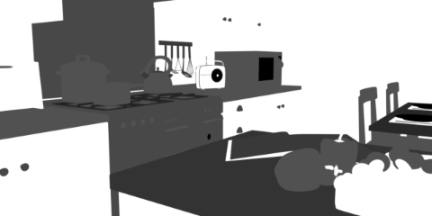}&
    \includegraphics[width=0.28\linewidth]{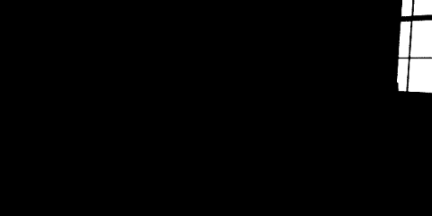}&
    \includegraphics[width=0.28\linewidth]{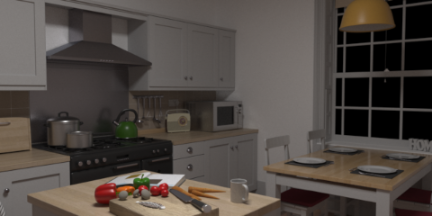}\\
    \end{tabular}
    }
    \caption{
    \textbf{Sensitivity analysis on training inputs.}
    Imperfect geometry and the usage of semantic segmentation instead of fine-grained part segmentation (row 1-2) can be acceptable for our BRDF-emission estimation (row 3-4, column 1-2).
    Ambiguity in roughness increases as geometry is imperfect or coarser segmentation is used (row 3-4, column 2),
    but they do not significantly affect applications like relighting (row 3-4, column 4).
    }
    \label{fig:training-input-geometry}
\end{figure}

\begin{figure}[t]
    \centering
\resizebox{0.99\linewidth}{!}{
    \begin{tabular}{c}
    \includegraphics[width=0.99\linewidth]{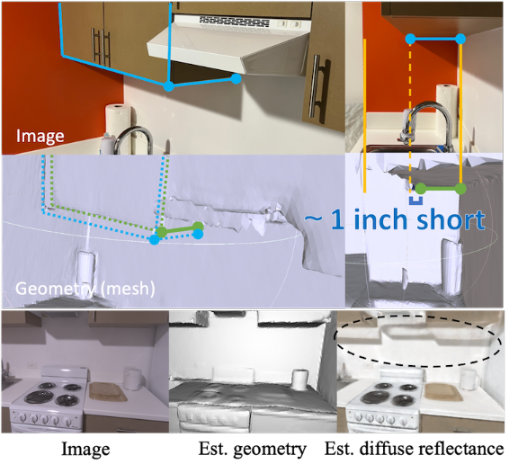}\\
    [-1.5ex]
    \includegraphics[width=0.99\linewidth]{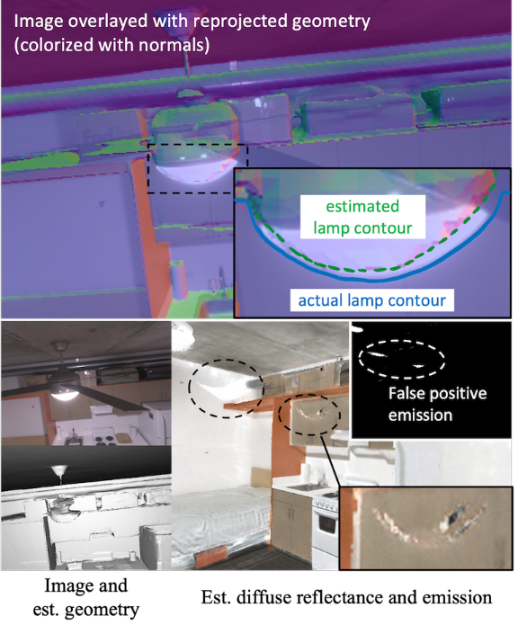} 
    \end{tabular}
    }
    \caption{\textbf{Failure cases of our method due to bad geometry.} 
    Top: indented cupboard 
    (actual boundary in blue; estimated in green which has a geometry error of around 1 inch)  
    results in incorrect light-surface intersection
    and boundary artifacts on the diffuse reflectance map (circled).
    Bottom: emission and bright artifacts (for diffuse reflectance) get erroneously baked onto the wall and cupboard (circled) because of the missing geometry on the lamp.
    }
    \label{fig:supp-failure-case}
\end{figure}

\subsection{Ablation study details}
\label{subsec:ablation-details}
Fig.~\ref{fig:training-strategy} shows the effect of different training strategies on BRDF-emission estimation as discussed in Sec.~5.4.
To demonstrate the impact of noisy inputs,
Fig.~\ref{fig:training-input-geometry} shows the quality of estimated geometry and semantic segmentation with respect to their ground truth together with the corresponding reconstruction results.
It can be seen that surface roughness for regions with weak highlights can be very sensitive to inputs,
while emission and material reflectance estimation are robust as long as the noise stays in a reasonable range.

\vspace{-8pt}
\paragraph{Failure cases.}
As discussed in the limitation section (Sec.~6),
broken geometry can lead to large artifacts in our BRDF-emission reconstruction.
A dormitory scene capture is shown in Fig.~\ref{fig:supp-failure-case} to demonstrate the problem,
where the front face of the reconstructed wall cabinet fails to align with the actual geometry (because of insufficient view coverage),
causing the shadow boundary to be baked into the reflectance map.
Meanwhile, geometry of the lamp on the ceiling fan is partly missing,
which causes the emission to be incorrectly projected to the background wall and cabinet surface,
creating bright artifacts on the reflectance map
and phantom emitters on the wall.

\subsection{Additional results}
\label{subsec:synthetic-scene}
\begin{figure*}[ht]
    \begin{center}
    \setlength{\tabcolsep}{0.05em}

\resizebox{0.99\linewidth}{!}{%

\begin{tabular}{c c c c c c c c}
\multicolumn{7}{c}{\textsf{\large Kitchen}}\\
[-6.5ex]
\multicolumn{7}{r}{
\begin{tabular}{cc}
 \includegraphics[width=0.13\linewidth]{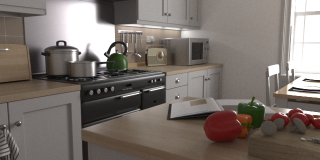}&
     \includegraphics[width=0.13\linewidth]{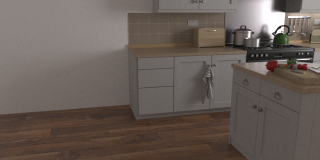}
\end{tabular}
}\\
\multicolumn{1}{c}{\milo} & \multicolumn{1}{c}{\ipt} & \multicolumn{1}{c}{\lieccv} & \multicolumn{1}{c}{\neilf} & \multicolumn{1}{c}{\ours~(Ours)} & \multicolumn{1}{c}{\ours-sem~(Ours)} & \multicolumn{1}{c}{Ground truth}\\

\includegraphics[width=.15\linewidth]{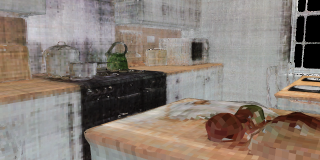} &
\includegraphics[width=.15\linewidth]{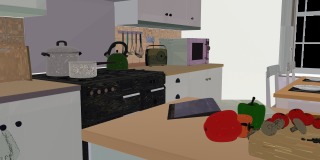} &
\includegraphics[width=.15\linewidth]{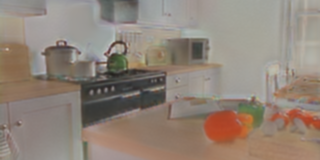} &
\includegraphics[width=.15\linewidth]{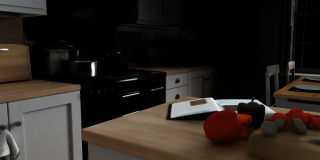} &
\includegraphics[width=.15\linewidth]{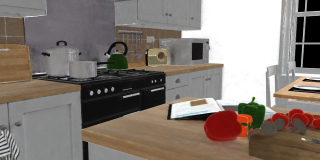} &
\includegraphics[width=.15\linewidth]{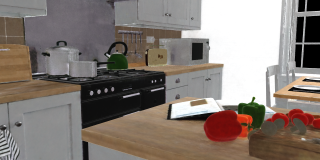} &
\includegraphics[width=.15\linewidth]{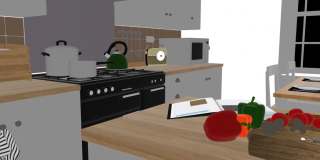} \\ [-0.65ex] 

\includegraphics[width=.15\linewidth]{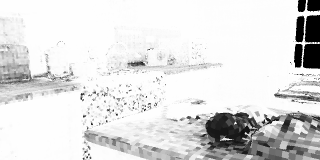} &
\includegraphics[width=.15\linewidth]{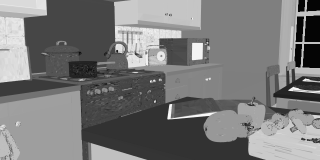} &
\includegraphics[width=.15\linewidth]{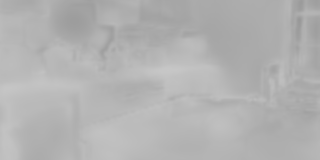} &
\includegraphics[width=.15\linewidth]{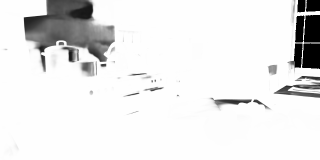} &
\includegraphics[width=.15\linewidth]{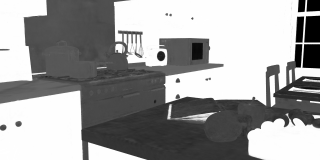} &
\includegraphics[width=.15\linewidth]{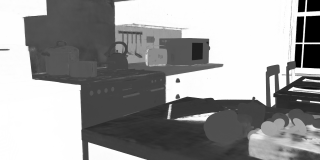} &
\includegraphics[width=.15\linewidth]{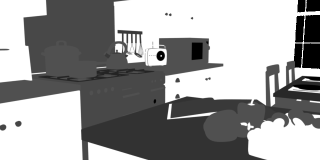} \\ [-0.65ex]
 
\includegraphics[width=.15\linewidth]{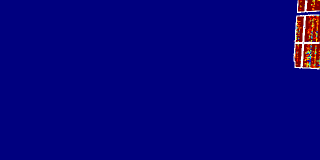} &
\includegraphics[width=.15\linewidth]{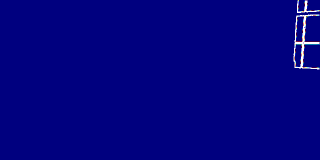} &
\includegraphics[width=.15\linewidth]{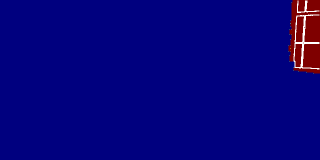} &
\includegraphics[width=.15\linewidth]{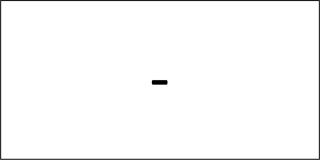} &
\includegraphics[width=.15\linewidth]{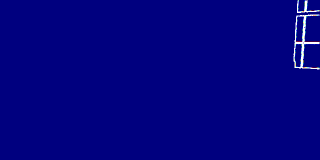} &
\includegraphics[width=.15\linewidth]{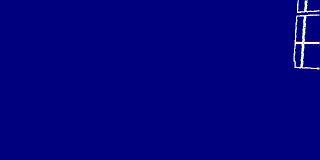} &
\includegraphics[width=.15\linewidth]{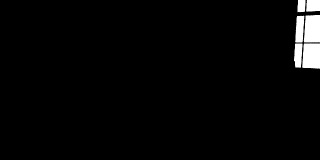} \\ 

\includegraphics[width=.15\linewidth]{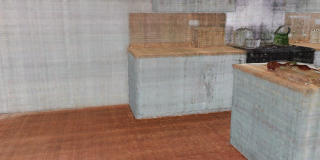} &
\includegraphics[width=.15\linewidth]{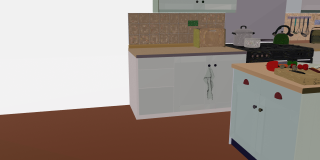} &
\includegraphics[width=.15\linewidth]{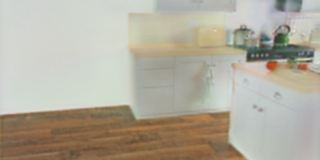} &
\includegraphics[width=.15\linewidth]{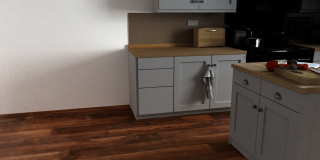} &
\includegraphics[width=.15\linewidth]{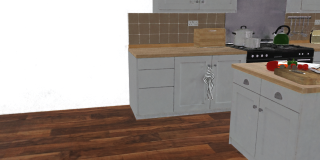} &
\includegraphics[width=.15\linewidth]{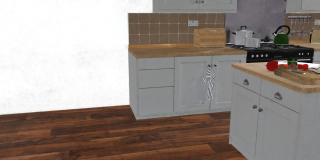} &
\includegraphics[width=.15\linewidth]{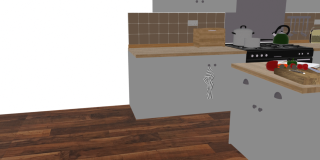} \\ [-0.65ex] 

\includegraphics[width=.15\linewidth]{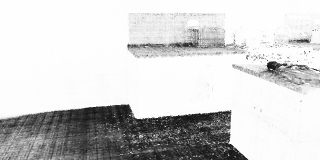} &
\includegraphics[width=.15\linewidth]{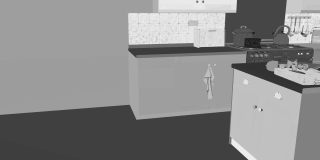} &
\includegraphics[width=.15\linewidth]{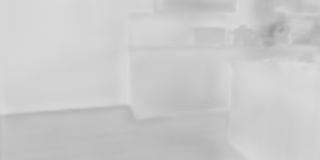} &
\includegraphics[width=.15\linewidth]{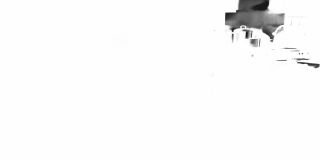} &
\includegraphics[width=.15\linewidth]{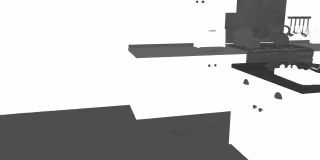} &
\includegraphics[width=.15\linewidth]{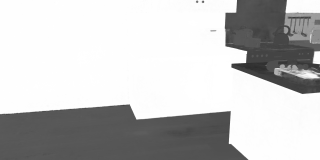} &
\includegraphics[width=.15\linewidth]{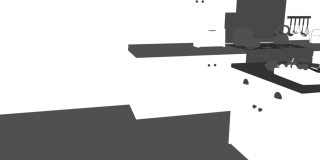} \\ [-0.65ex]
 
\includegraphics[width=.15\linewidth]{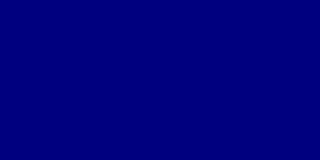} &
\includegraphics[width=.15\linewidth]{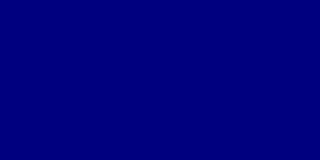} &
\includegraphics[width=.15\linewidth]{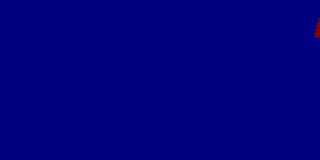} &
\includegraphics[width=.15\linewidth]{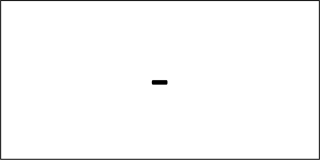} &
\includegraphics[width=.15\linewidth]{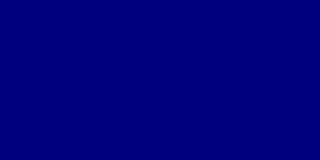} &
\includegraphics[width=.15\linewidth]{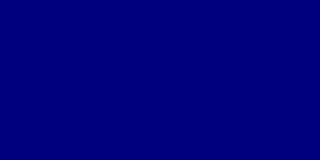} &
\includegraphics[width=.15\linewidth]{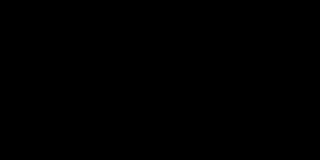} \\[4ex]
\multicolumn{7}{c}{\textsf{\large Bathroom}}\\[-6.5ex]
\multicolumn{7}{r}{
\begin{tabular}{cc}
     \includegraphics[width=0.13\linewidth]{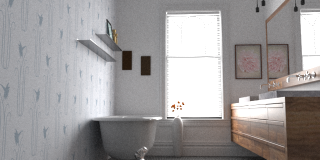}&
     \includegraphics[width=0.13\linewidth]{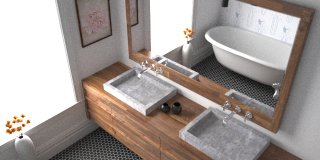}
\end{tabular}
}\\

\includegraphics[width=.15\linewidth]{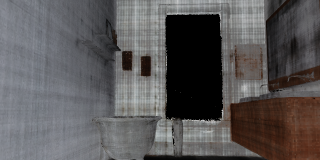} &
\includegraphics[width=.15\linewidth]{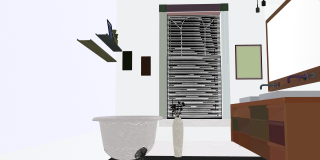} &
\includegraphics[width=.15\linewidth]{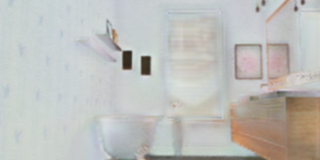} &
\includegraphics[width=.15\linewidth]{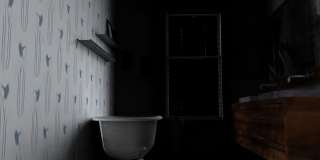} &
\includegraphics[width=.15\linewidth]{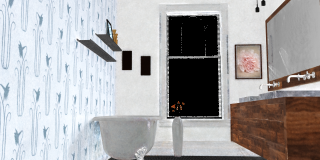} &
\includegraphics[width=.15\linewidth]{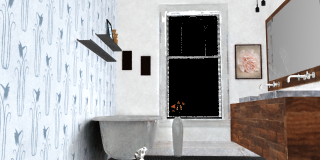} &
\includegraphics[width=.15\linewidth]{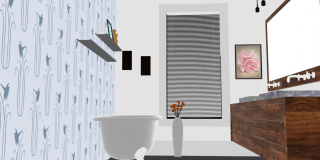} \\ [-0.65ex] 

\includegraphics[width=.15\linewidth]{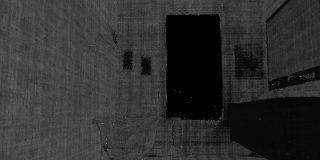} &
\includegraphics[width=.15\linewidth]{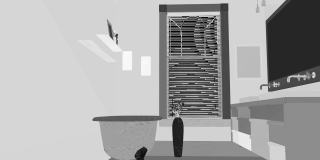} &
\includegraphics[width=.15\linewidth]{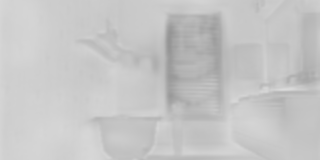} &
\includegraphics[width=.15\linewidth]{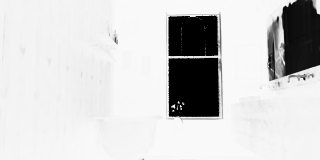} &
\includegraphics[width=.15\linewidth]{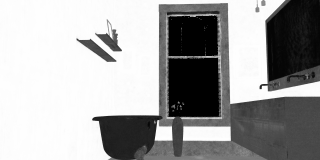} &
\includegraphics[width=.15\linewidth]{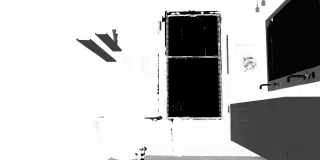} &
\includegraphics[width=.15\linewidth]{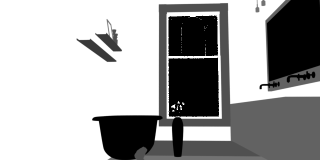} \\ [-0.65ex]

\includegraphics[width=.15\linewidth]{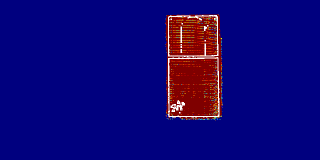} &
\includegraphics[width=.15\linewidth]{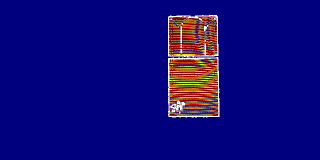} &
\includegraphics[width=.15\linewidth]{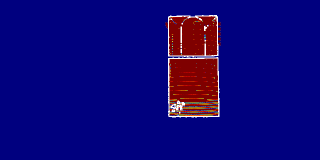} &
\includegraphics[width=.15\linewidth]{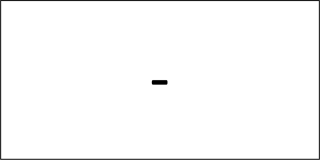} &
\includegraphics[width=.15\linewidth]{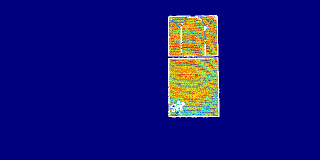} &
\includegraphics[width=.15\linewidth]{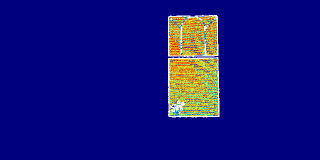} &
\includegraphics[width=.15\linewidth]{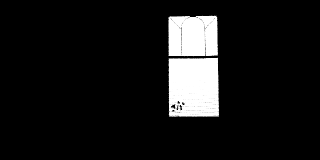} \\ 

\includegraphics[width=.15\linewidth]{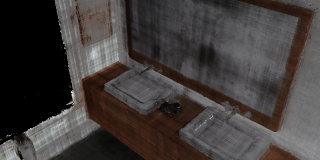} &
\includegraphics[width=.15\linewidth]{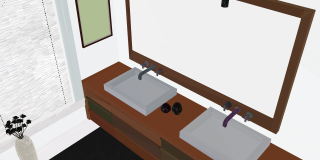} &
\includegraphics[width=.15\linewidth]{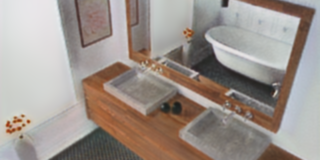} &
\includegraphics[width=.15\linewidth]{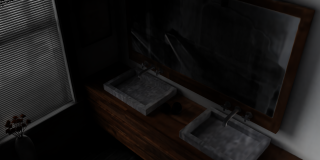} &
\includegraphics[width=.15\linewidth]{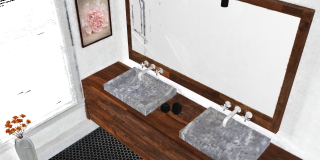} &
\includegraphics[width=.15\linewidth]{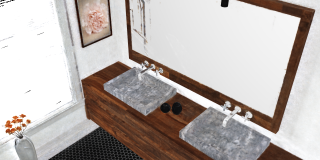} &
\includegraphics[width=.15\linewidth]{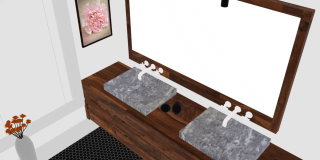} \\ [-0.65ex] 

\includegraphics[width=.15\linewidth]{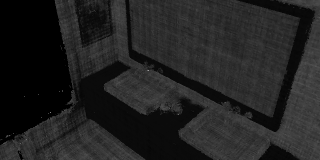} &
\includegraphics[width=.15\linewidth]{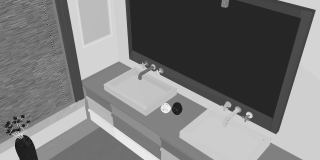} &
\includegraphics[width=.15\linewidth]{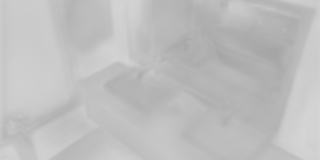} &
\includegraphics[width=.15\linewidth]{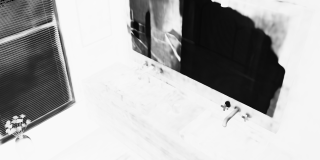} &
\includegraphics[width=.15\linewidth]{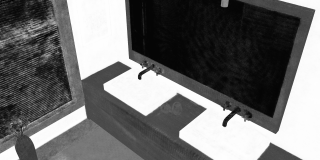} &
\includegraphics[width=.15\linewidth]{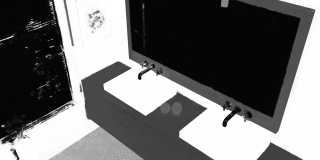} &
\includegraphics[width=.15\linewidth]{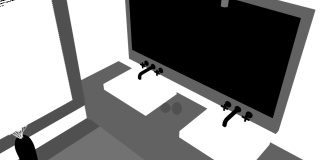} \\ [-0.65ex]
 
\includegraphics[width=.15\linewidth]{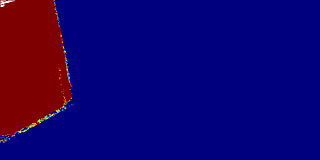} &
\includegraphics[width=.15\linewidth]{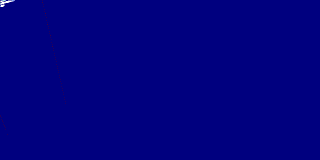} &
\includegraphics[width=.15\linewidth]{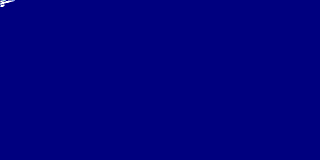} &
\includegraphics[width=.15\linewidth]{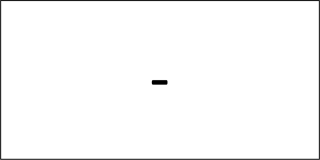} &
\includegraphics[width=.15\linewidth]{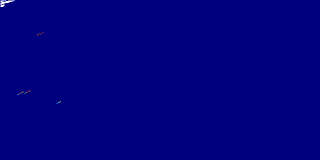} &
\includegraphics[width=.15\linewidth]{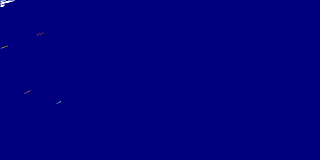} &
\includegraphics[width=.15\linewidth]{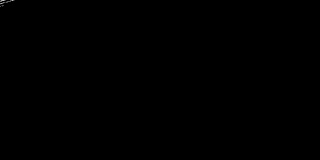}

\end{tabular}
}
\end{center}

\caption{\textbf{BRDF and emission estimation on synthetic Kitchen and Bathroom for all methods.} Input views are shown in the upper-right corner of each scene.}
\label{fig:synthetic_BRDF_supp_1}
\end{figure*}

\begin{figure*}[hbt!]
    \begin{center}
    \setlength{\tabcolsep}{0.05em}
    
\resizebox{0.99\linewidth}{!}{%
\begin{tabular}{c c c c c c c c}
\multicolumn{7}{c}{\textsf{\large Bedroom}}\\[-6.5ex]
\multicolumn{7}{r}{
    \begin{tabular}{cc}
         \includegraphics[width=0.13\linewidth]{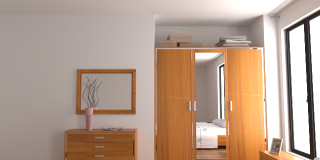}&
     \includegraphics[width=0.13\linewidth]{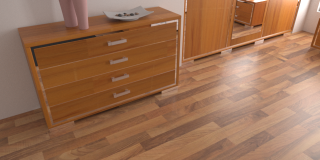}
    \end{tabular}
}\\

\multicolumn{1}{c}{\milo} & \multicolumn{1}{c}{\ipt} & \multicolumn{1}{c}{\lieccv} & \multicolumn{1}{c}{\neilf} & \multicolumn{1}{c}{\ours~(Ours)} & \multicolumn{1}{c}{\ours-sem~(Ours)} & \multicolumn{1}{c}{Ground truth}\\ 

\includegraphics[width=.15\linewidth]{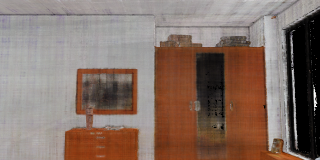} &
\includegraphics[width=.15\linewidth]{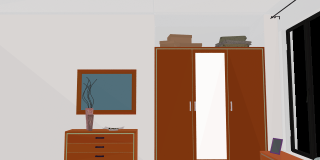} &
\includegraphics[width=.15\linewidth]{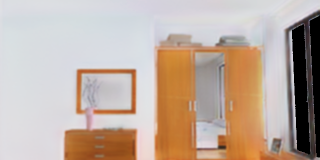} &
\includegraphics[width=.15\linewidth]{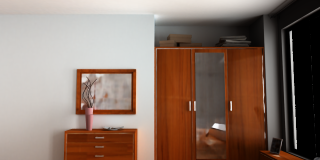} &
\includegraphics[width=.15\linewidth]{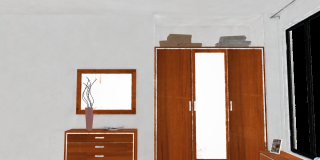} &
\includegraphics[width=.15\linewidth]{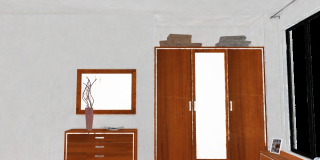} &
\includegraphics[width=.15\linewidth]{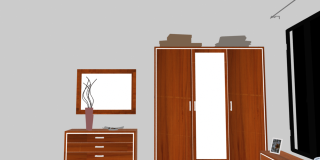} \\ [-0.65ex] 

\includegraphics[width=.15\linewidth]{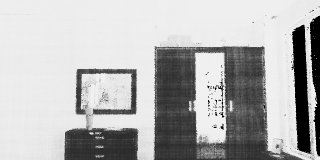} &
\includegraphics[width=.15\linewidth]{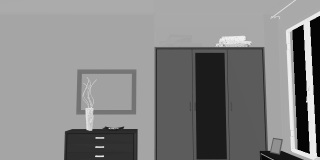} &
\includegraphics[width=.15\linewidth]{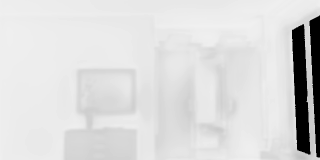} &
\includegraphics[width=.15\linewidth]{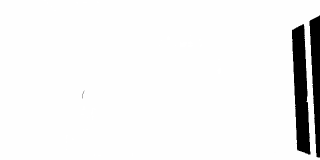} &
\includegraphics[width=.15\linewidth]{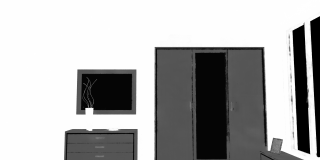} &
\includegraphics[width=.15\linewidth]{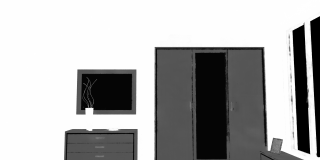} &
\includegraphics[width=.15\linewidth]{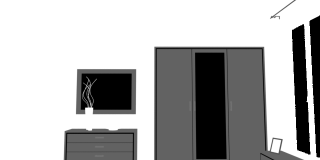} \\ [-0.65ex]
 
\includegraphics[width=.15\linewidth]{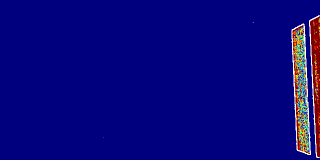} &
\includegraphics[width=.15\linewidth]{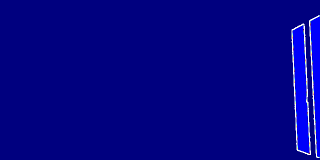} &
\includegraphics[width=.15\linewidth]{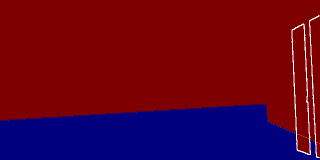} &
\includegraphics[width=.15\linewidth]{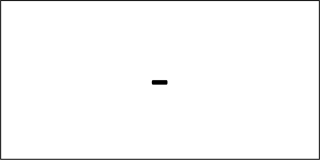} &
\includegraphics[width=.15\linewidth]{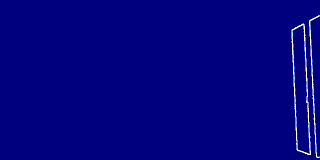} &
\includegraphics[width=.15\linewidth]{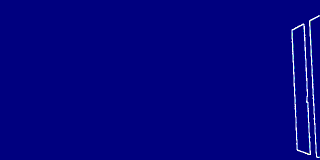} &
\includegraphics[width=.15\linewidth]{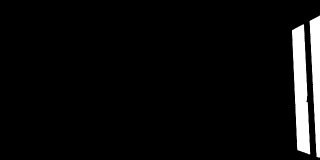} \\

\includegraphics[width=.15\linewidth]{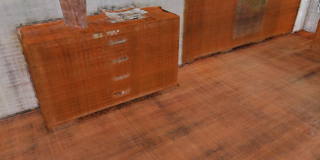} &
\includegraphics[width=.15\linewidth]{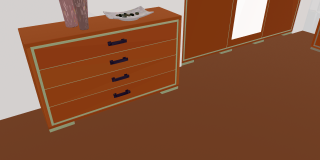} &
\includegraphics[width=.15\linewidth]{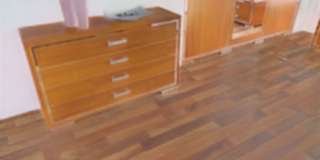} &
\includegraphics[width=.15\linewidth]{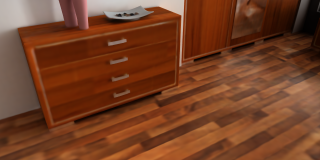} &
\includegraphics[width=.15\linewidth]{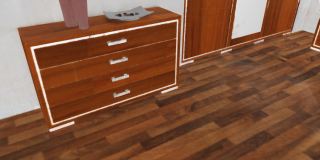} &
\includegraphics[width=.15\linewidth]{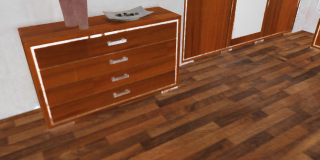} &
\includegraphics[width=.15\linewidth]{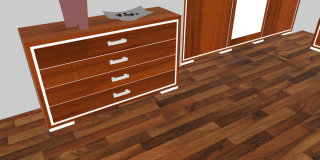} \\ [-0.65ex] 

\includegraphics[width=.15\linewidth]{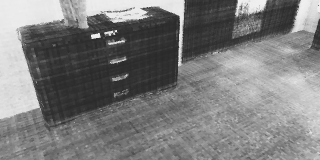} &
\includegraphics[width=.15\linewidth]{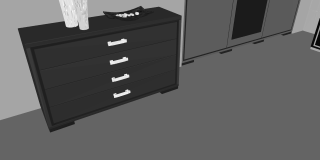} &
\includegraphics[width=.15\linewidth]{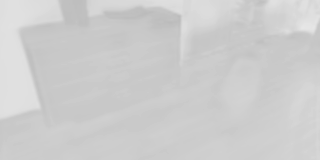} &
\includegraphics[width=.15\linewidth]{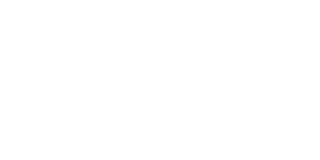} &
\includegraphics[width=.15\linewidth]{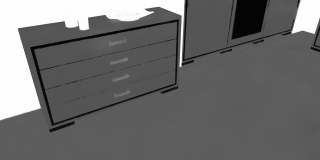} &
\includegraphics[width=.15\linewidth]{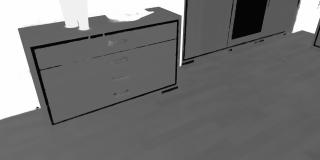} &
\includegraphics[width=.15\linewidth]{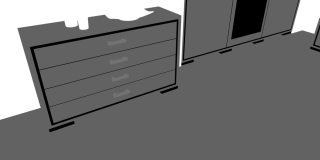} \\ [-0.65ex]
 
\includegraphics[width=.15\linewidth]{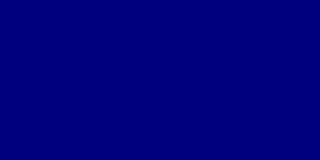} &
\includegraphics[width=.15\linewidth]{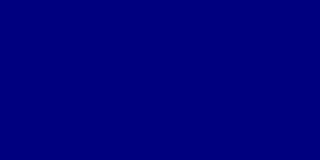} &
\includegraphics[width=.15\linewidth]{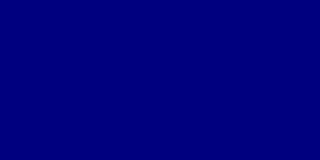} &
\includegraphics[width=.15\linewidth]{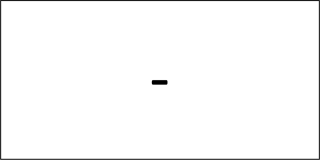} &
\includegraphics[width=.15\linewidth]{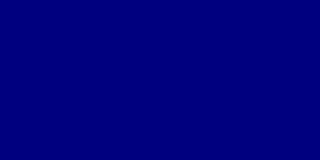} &
\includegraphics[width=.15\linewidth]{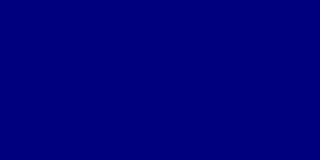} &
\includegraphics[width=.15\linewidth]{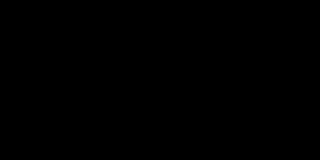} \\
[4ex]

\multicolumn{7}{c}{\textsf{\large Livingroom}}\\[-6.5ex]
\multicolumn{7}{r}{
\begin{tabular}{cc}
     \includegraphics[width=0.13\linewidth]{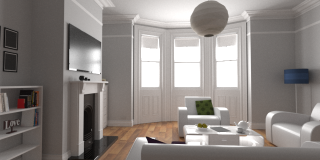}&
     \includegraphics[width=0.13\linewidth]{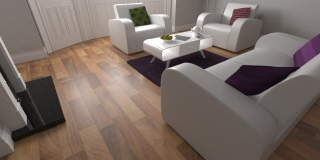}
\end{tabular}
}\\

\includegraphics[width=.15\linewidth]{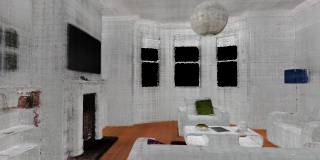} &
\includegraphics[width=.15\linewidth]{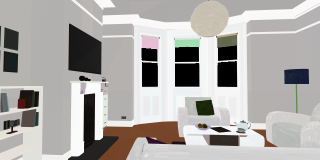} &
\includegraphics[width=.15\linewidth]{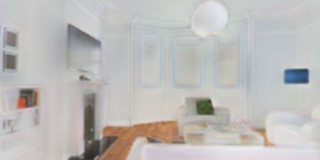} &
\includegraphics[width=.15\linewidth]{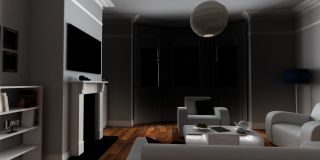} &
\includegraphics[width=.15\linewidth]{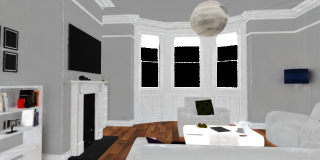} &
\includegraphics[width=.15\linewidth]{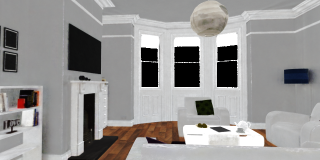} &
\includegraphics[width=.15\linewidth]{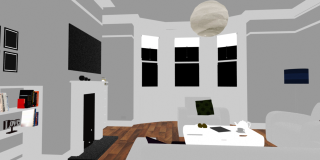} \\ [-0.65ex] 

\includegraphics[width=.15\linewidth]{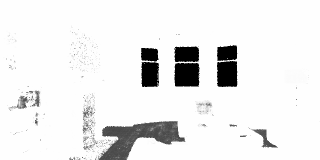} &
\includegraphics[width=.15\linewidth]{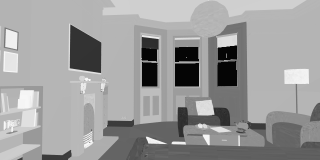} &
\includegraphics[width=.15\linewidth]{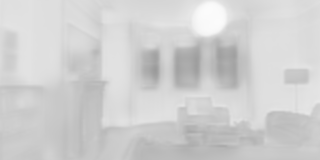} &
\includegraphics[width=.15\linewidth]{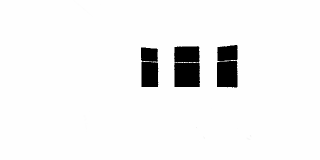} &
\includegraphics[width=.15\linewidth]{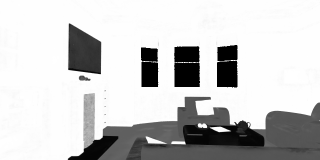} &
\includegraphics[width=.15\linewidth]{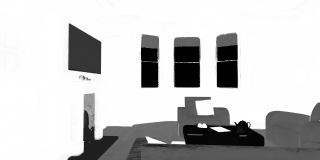} &
\includegraphics[width=.15\linewidth]{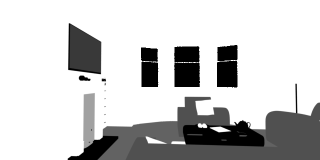} \\ [-0.65ex]
 
\includegraphics[width=.15\linewidth]{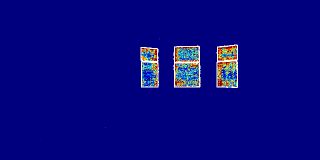} &
\includegraphics[width=.15\linewidth]{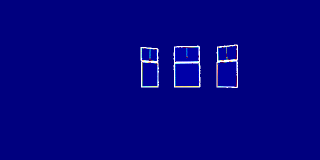} &
\includegraphics[width=.15\linewidth]{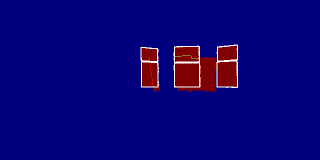} &
\includegraphics[width=.15\linewidth]{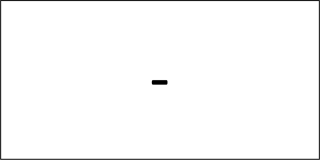} &
\includegraphics[width=.15\linewidth]{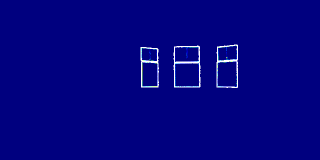} &
\includegraphics[width=.15\linewidth]{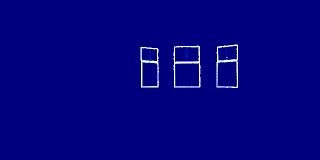} &
\includegraphics[width=.15\linewidth]{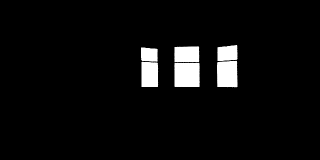} \\ 

\includegraphics[width=.15\linewidth]{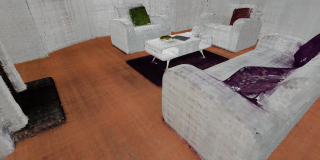} &
\includegraphics[width=.15\linewidth]{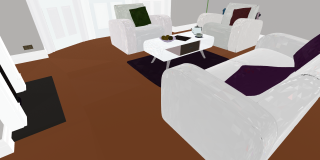} &
\includegraphics[width=.15\linewidth]{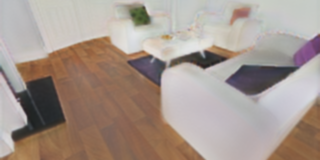} &
\includegraphics[width=.15\linewidth]{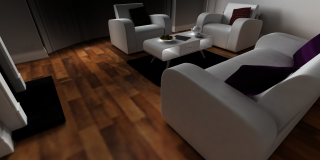} &
\includegraphics[width=.15\linewidth]{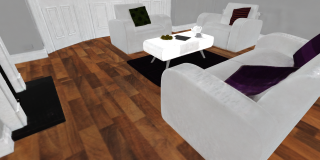} &
\includegraphics[width=.15\linewidth]{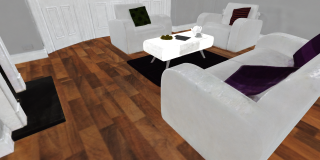} &
\includegraphics[width=.15\linewidth]{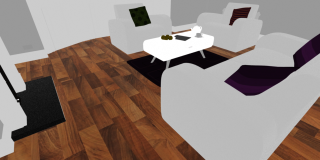} \\ [-0.65ex] 

\includegraphics[width=.15\linewidth]{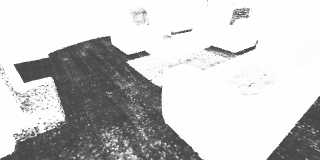} &
\includegraphics[width=.15\linewidth]{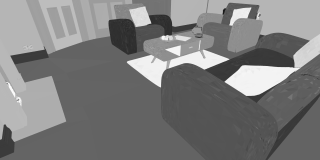} &
\includegraphics[width=.15\linewidth]{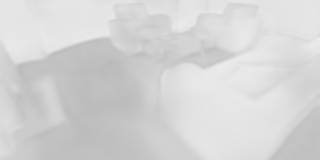} &
\includegraphics[width=.15\linewidth]{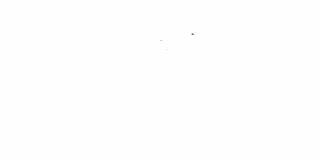} &
\includegraphics[width=.15\linewidth]{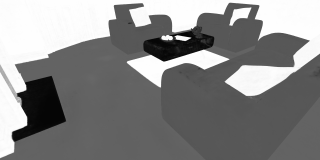} &
\includegraphics[width=.15\linewidth]{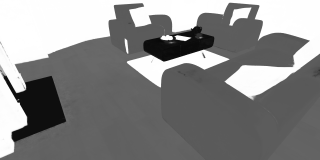} &
\includegraphics[width=.15\linewidth]{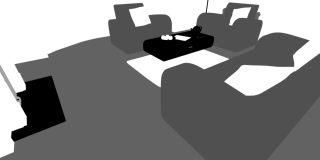} \\ [-0.65ex]

\includegraphics[width=.15\linewidth]{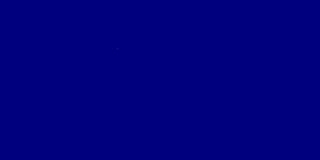} &
\includegraphics[width=.15\linewidth]{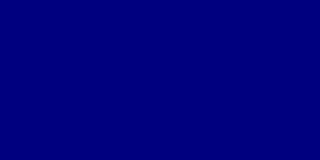} &
\includegraphics[width=.15\linewidth]{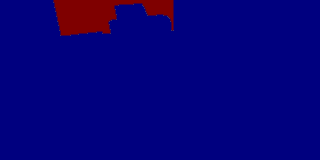} &
\includegraphics[width=.15\linewidth]{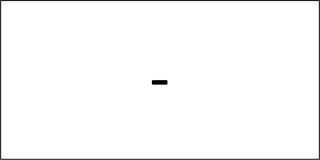} &
\includegraphics[width=.15\linewidth]{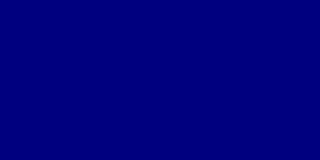} &
\includegraphics[width=.15\linewidth]{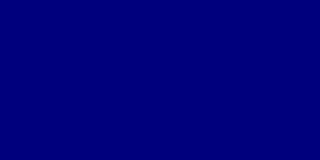} &
\includegraphics[width=.15\linewidth]{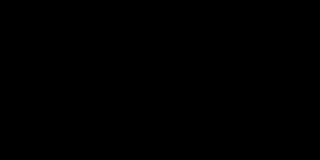}

\end{tabular}
}
\end{center}

\caption{\textbf{BRDF and emission estimation results on synthetic Bedroom and Livingroom for all methods}, showing 2 views per-scene. Input views are shown in the upper-right corner of each scene.
}
\label{fig:synthetic_BRDF_supp_2}
\end{figure*}

\begin{figure*}[hbt!]
    \begin{center}
    \setlength{\tabcolsep}{0.1em}

\resizebox{0.99\linewidth}{!}{%
\begin{tabular}{c c c c c c}

\multicolumn{6}{c}{\textsf{\large Kitchen}}\\[1ex]
\multicolumn{1}{c}{\milo} & \multicolumn{1}{c}{\ipt} & \multicolumn{1}{c}{\fvp} & \multicolumn{1}{c}{\ours~(Ours)} & \multicolumn{1}{c}{\ours-sem~(Ours)} & \multicolumn{1}{c}{Ground truth}\\ 

\includegraphics[width=.18\textwidth]{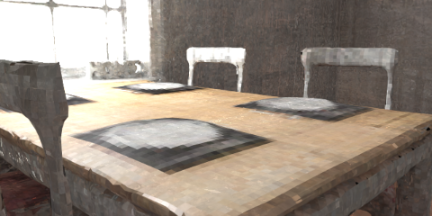} &
\includegraphics[width=.18\textwidth]{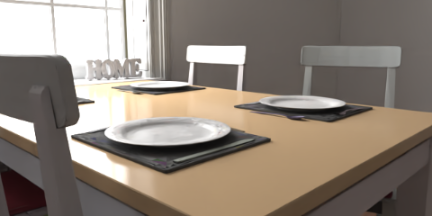} &
\includegraphics[width=.18\textwidth]{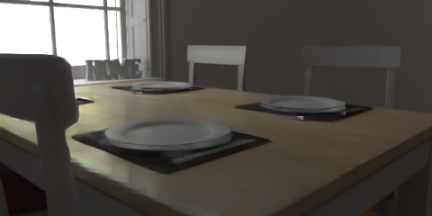} &
\includegraphics[width=.18\textwidth]{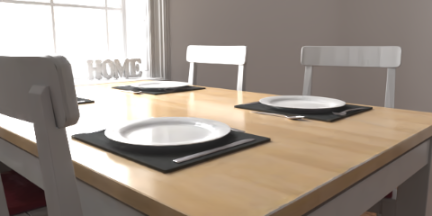} &
\includegraphics[width=.18\textwidth]{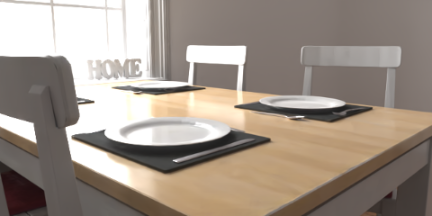} &
\includegraphics[width=.18\textwidth]{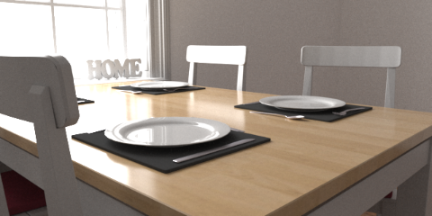} \\ [-0.65ex] 

\includegraphics[width=.18\textwidth]{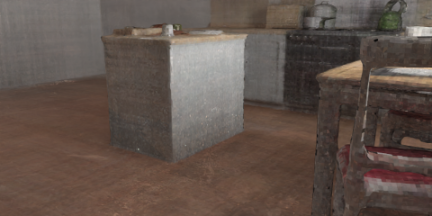} &
\includegraphics[width=.18\textwidth]{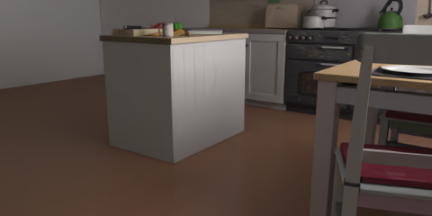} &
\includegraphics[width=.18\textwidth]{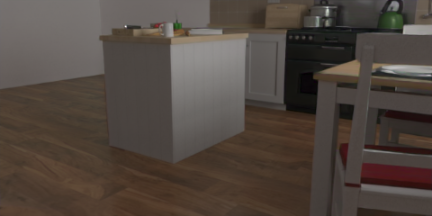} &
\includegraphics[width=.18\textwidth]{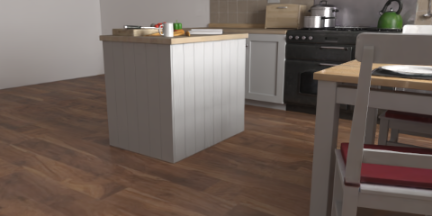} &
\includegraphics[width=.18\textwidth]{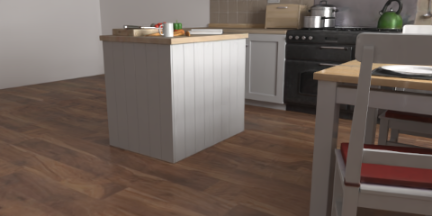} &
\includegraphics[width=.18\textwidth]{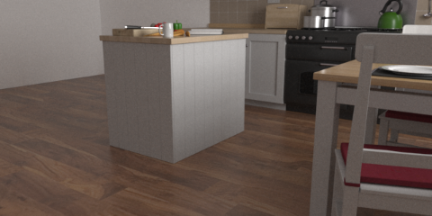} \\ [1ex] 

\multicolumn{6}{c}{\textsf{\large Bathroom}}\\[1ex]

\includegraphics[width=.18\textwidth]{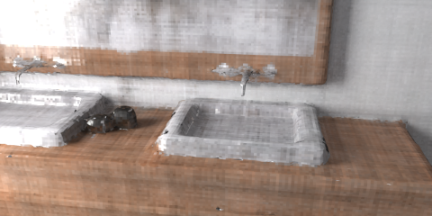} &
\includegraphics[width=.18\textwidth]{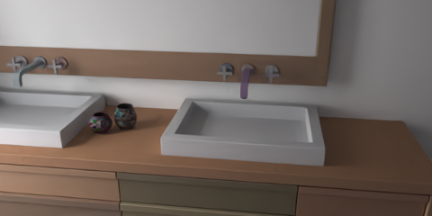} &
\includegraphics[width=.18\textwidth]{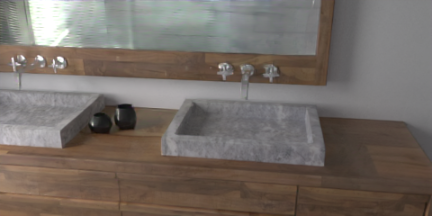} &
\includegraphics[width=.18\textwidth]{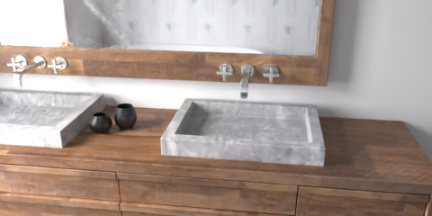} &
\includegraphics[width=.18\textwidth]{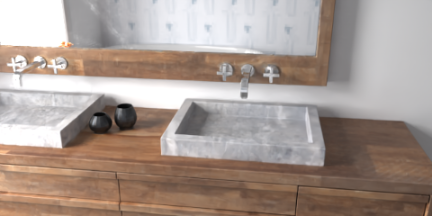} &
\includegraphics[width=.18\textwidth]{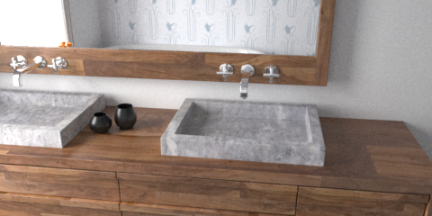} \\ [-0.65ex] 

\includegraphics[width=.18\textwidth]{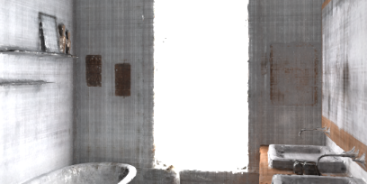} &
\includegraphics[width=.18\textwidth]{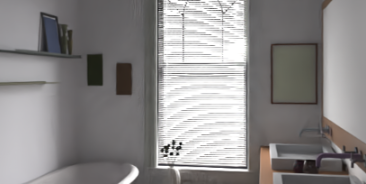} &
\includegraphics[width=.18\textwidth]{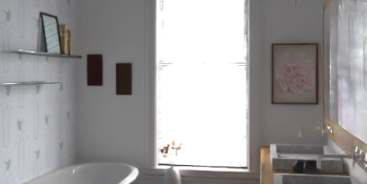} &
\includegraphics[width=.18\textwidth]{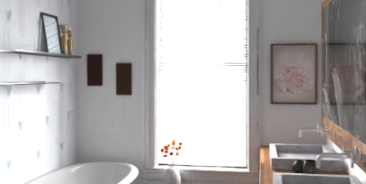} &
\includegraphics[width=.18\textwidth]{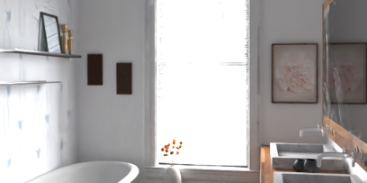} &
\includegraphics[width=.18\textwidth]{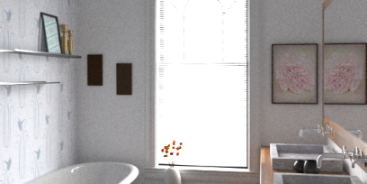} \\ [1ex] 

\multicolumn{6}{c}{\textsf{\large Bedroom}}\\[1ex]

\includegraphics[width=.18\textwidth]{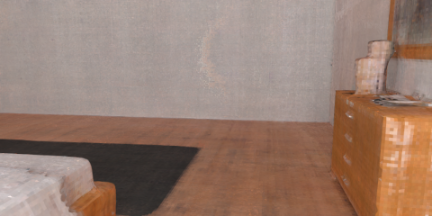} &
\includegraphics[width=.18\textwidth]{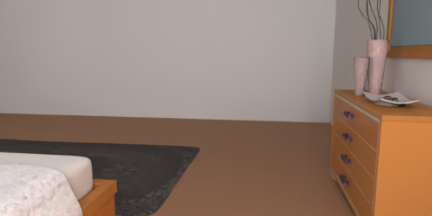} &
\includegraphics[width=.18\textwidth]{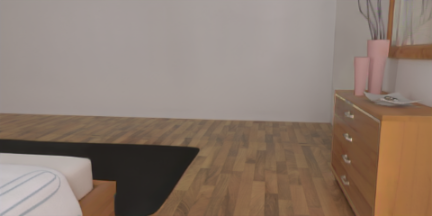} &
\includegraphics[width=.18\textwidth]{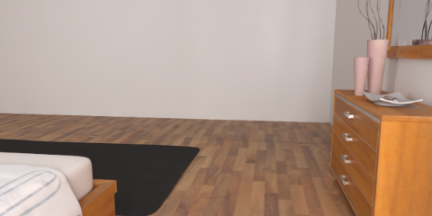} &
\includegraphics[width=.18\textwidth]{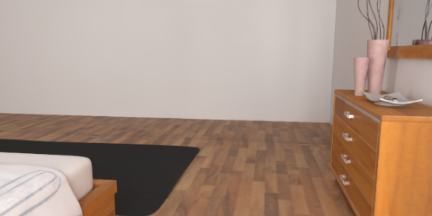} &
\includegraphics[width=.18\textwidth]{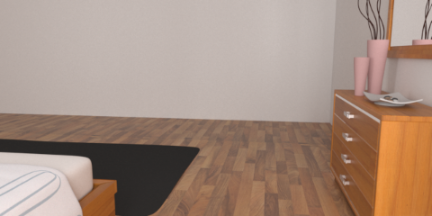} \\ [-0.65ex] 

\includegraphics[width=.18\textwidth]{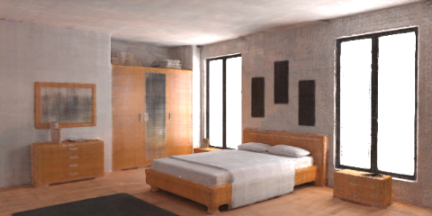} &
\includegraphics[width=.18\textwidth]{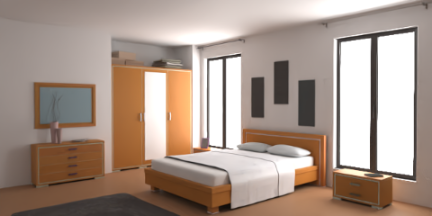} &
\includegraphics[width=.18\textwidth]{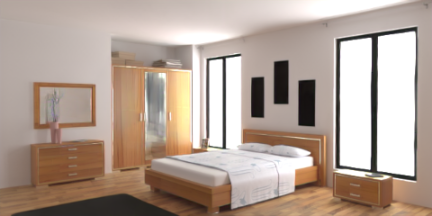} &
\includegraphics[width=.18\textwidth]{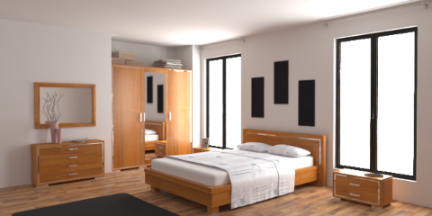} &
\includegraphics[width=.18\textwidth]{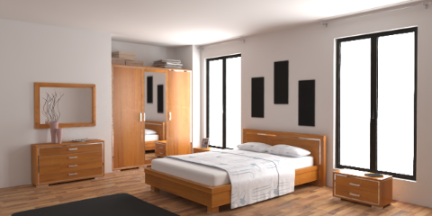} &
\includegraphics[width=.18\textwidth]{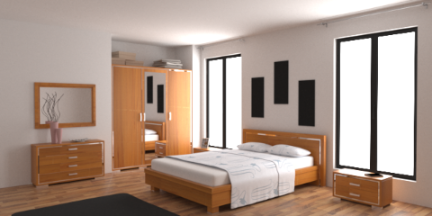} \\ [1ex] 

\multicolumn{6}{c}{\textsf{\large Livingroom}}\\[1ex]

\includegraphics[width=.18\textwidth]{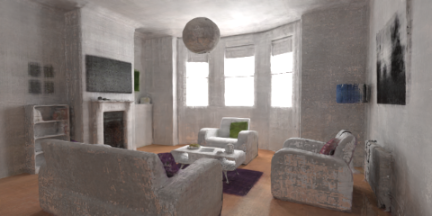} &
\includegraphics[width=.18\textwidth]{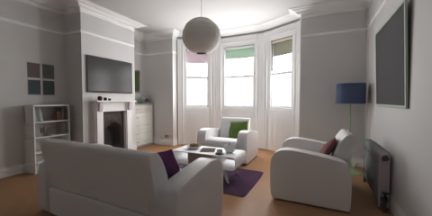} &
\includegraphics[width=.18\textwidth]{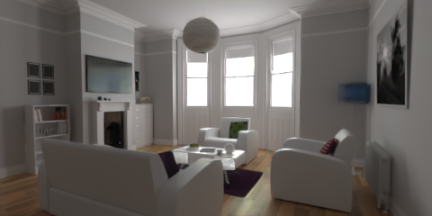} &
\includegraphics[width=.18\textwidth]{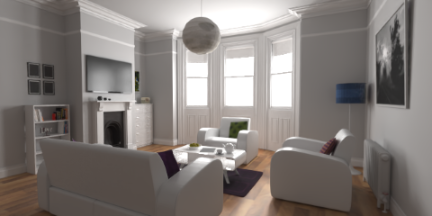} &
\includegraphics[width=.18\textwidth]{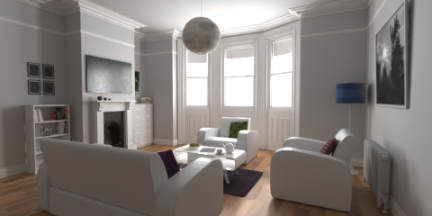} &
\includegraphics[width=.18\textwidth]{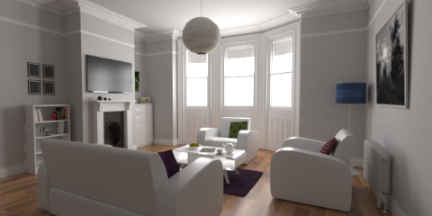} \\ [-0.65ex] 

\includegraphics[width=.18\textwidth]{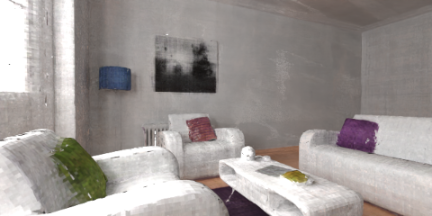} &
\includegraphics[width=.18\textwidth]{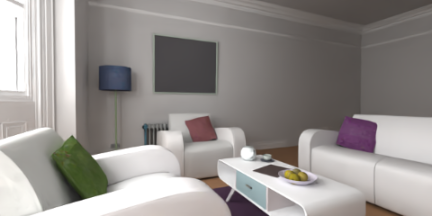} &
\includegraphics[width=.18\textwidth]{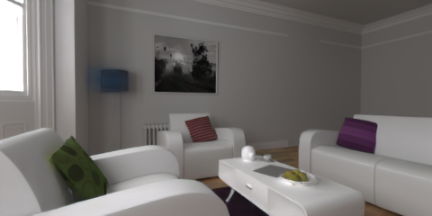} &
\includegraphics[width=.18\textwidth]{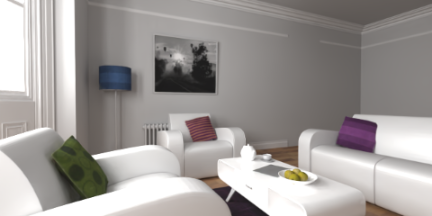} &
\includegraphics[width=.18\textwidth]{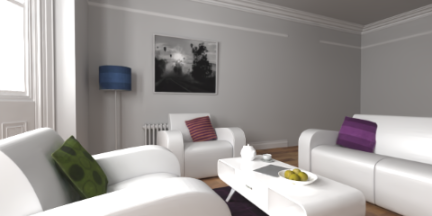} &
\includegraphics[width=.18\textwidth]{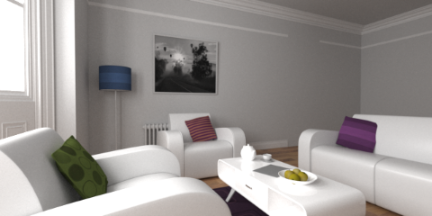} \\ 

\end{tabular}
}
\end{center}

\caption{\textbf{View synthesis results on synthetic scenes for all methods}, showing 2 views per-scene. 
}
\label{fig:synthetic_synthesis_more}
\end{figure*}

\begin{figure*}
    \begin{center}
    \setlength{\tabcolsep}{0.1em}

\resizebox{0.99\linewidth}{!}{%
\begin{tabular}{c c c c c c c}

\multicolumn{7}{c}{\textsf{\large Kitchen}}\\[1ex]
\multicolumn{1}{c}{\milo} & \multicolumn{1}{c}{\ipt} & \multicolumn{1}{c}{\lieccv} & \multicolumn{1}{c}{\fvp} & \multicolumn{1}{c}{\ours~(Ours)} & \multicolumn{1}{c}{\ours-sem~(Ours)} & \multicolumn{1}{c}{Ground truth}\\ 

\includegraphics[width=.16\textwidth]{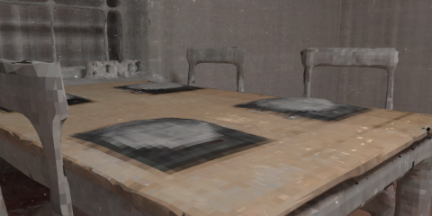} &
\includegraphics[width=.16\textwidth]{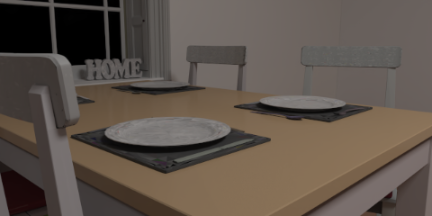} &
\includegraphics[width=.16\textwidth]{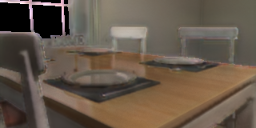} &
\includegraphics[width=.16\textwidth]{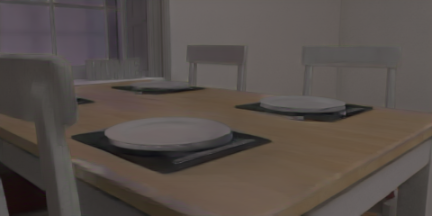} &
\includegraphics[width=.16\textwidth]{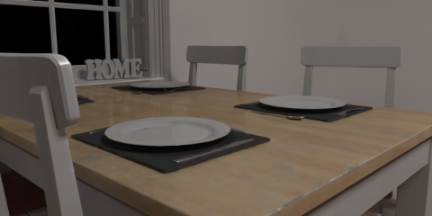} &
\includegraphics[width=.16\textwidth]{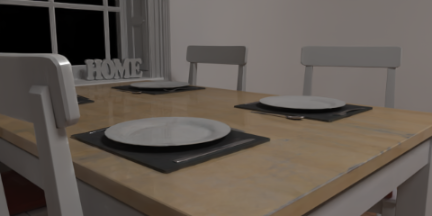} &
\includegraphics[width=.16\textwidth]{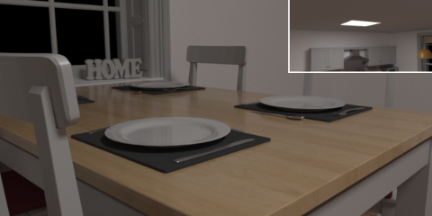} \\ [-0.65ex] 

\includegraphics[width=.16\textwidth]{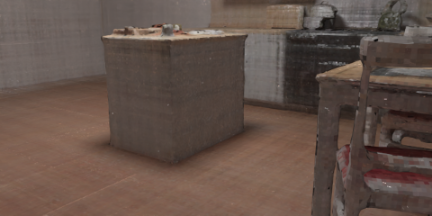} &
\includegraphics[width=.16\textwidth]{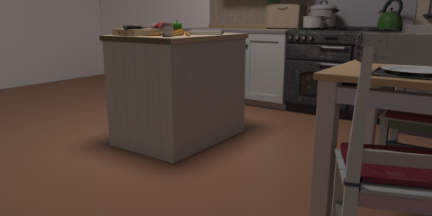} &
\includegraphics[width=.16\textwidth]{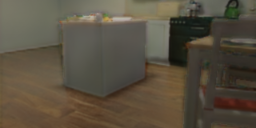} &
\includegraphics[width=.16\textwidth]{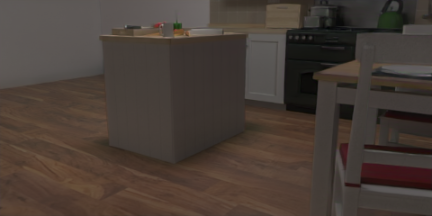} &
\includegraphics[width=.16\textwidth]{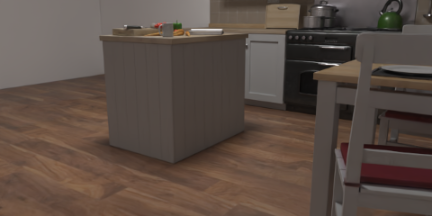} &
\includegraphics[width=.16\textwidth]{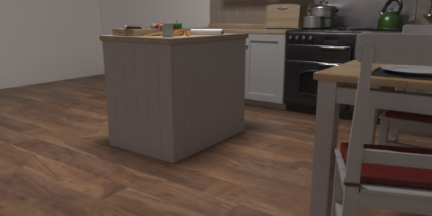} &
\includegraphics[width=.16\textwidth]{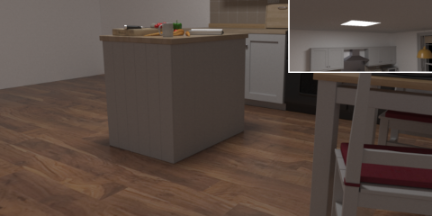} \\ [1ex] 
\multicolumn{7}{c}{\textsf{\large Bathroom}}\\[1ex]

\includegraphics[width=.16\textwidth]{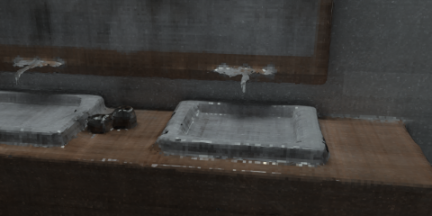} &
\includegraphics[width=.16\textwidth]{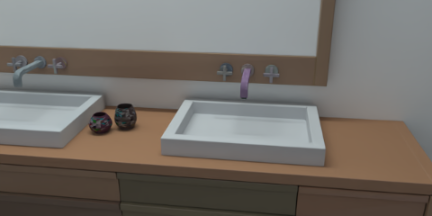} &
\includegraphics[width=.16\textwidth]{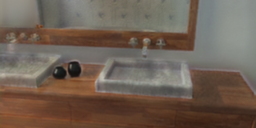} &
\includegraphics[width=.16\textwidth]{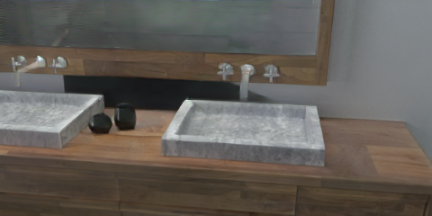} &
\includegraphics[width=.16\textwidth]{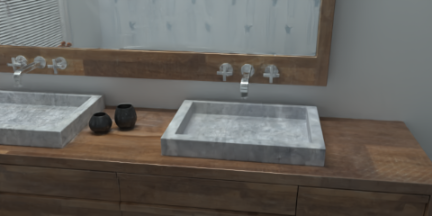} &
\includegraphics[width=.16\textwidth]{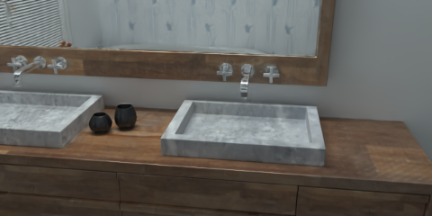} &
\includegraphics[width=.16\textwidth]{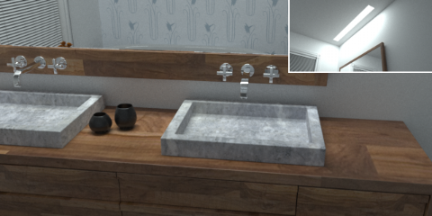} \\ [-0.65ex] 

\includegraphics[width=.16\textwidth]{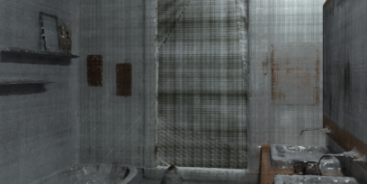} &
\includegraphics[width=.16\textwidth]{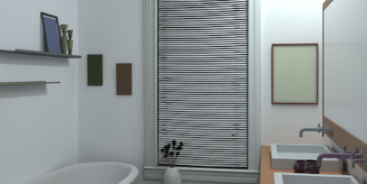} &
\includegraphics[width=.16\textwidth]{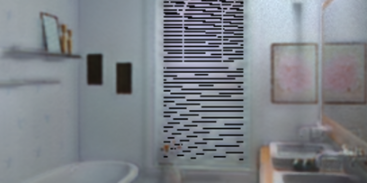} &
\includegraphics[width=.16\textwidth]{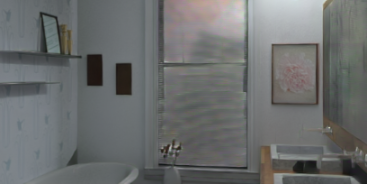} &
\includegraphics[width=.16\textwidth]{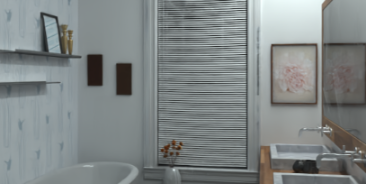} &
\includegraphics[width=.16\textwidth]{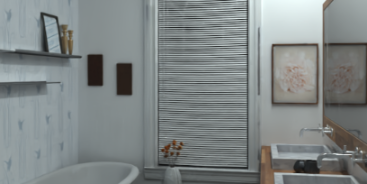} &
\includegraphics[width=.16\textwidth]{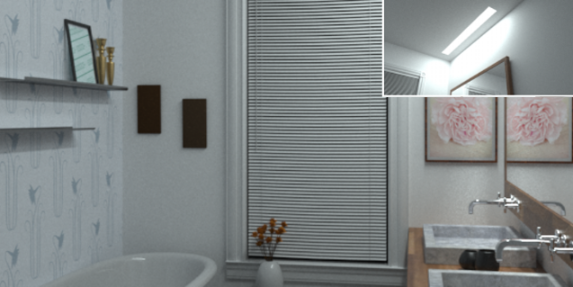} \\ [1ex] 
\multicolumn{7}{c}{\textsf{\large Bedroom}}\\[1ex]

\includegraphics[width=.16\textwidth]{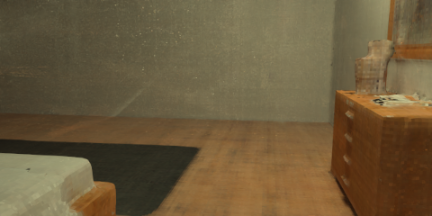} &
\includegraphics[width=.16\textwidth]{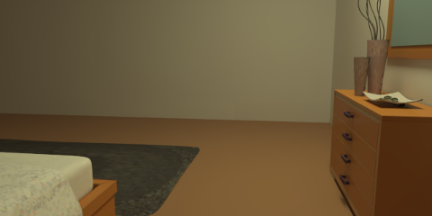} &
\includegraphics[width=.16\textwidth]{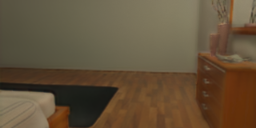} &
\includegraphics[width=.16\textwidth]{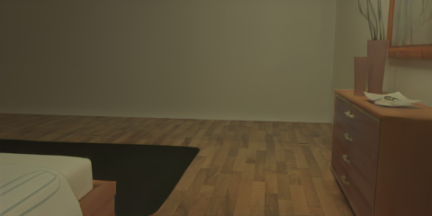} &
\includegraphics[width=.16\textwidth]{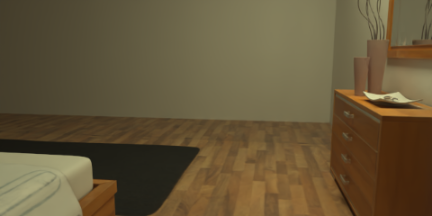} &
\includegraphics[width=.16\textwidth]{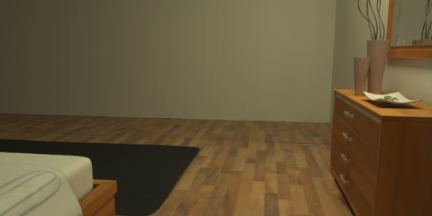} &
\includegraphics[width=.16\textwidth]{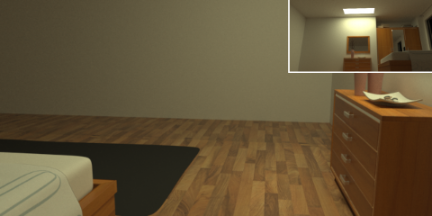} \\ [-0.65ex] 

\includegraphics[width=.16\textwidth]{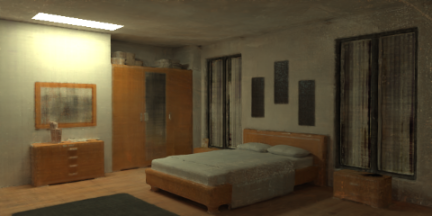} &
\includegraphics[width=.16\textwidth]{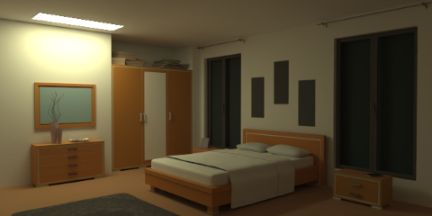} &
\includegraphics[width=.16\textwidth]{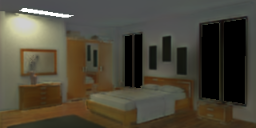} &
\includegraphics[width=.16\textwidth]{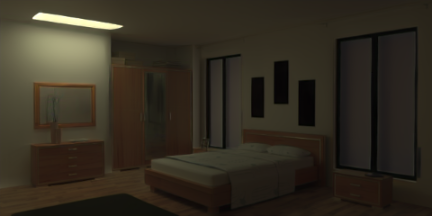} &
\includegraphics[width=.16\textwidth]{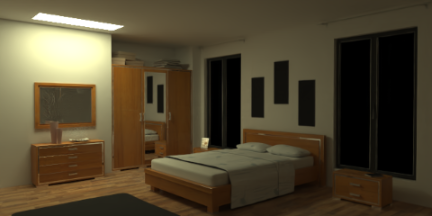} &
\includegraphics[width=.16\textwidth]{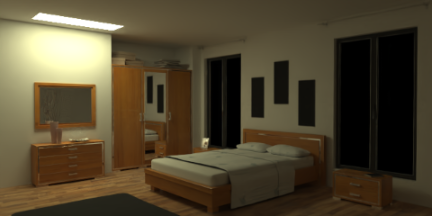} &
\includegraphics[width=.16\textwidth]{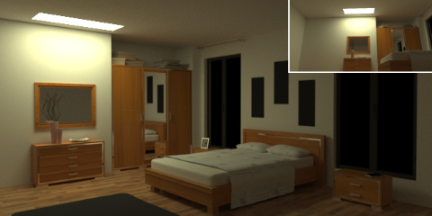} \\ [1ex] 
\multicolumn{7}{c}{\textsf{\large Livingroom}}\\[1ex]

\includegraphics[width=.16\textwidth]{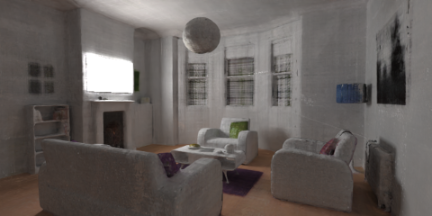} &
\includegraphics[width=.16\textwidth]{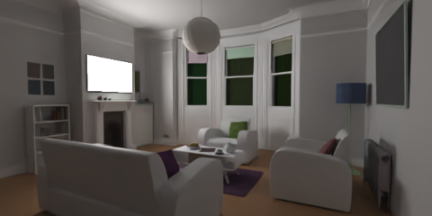} &
\includegraphics[width=.16\textwidth]{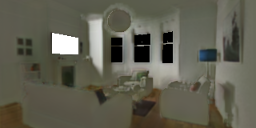} &
\includegraphics[width=.16\textwidth]{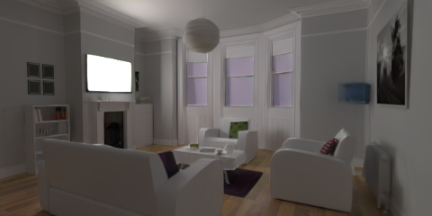} &
\includegraphics[width=.16\textwidth]{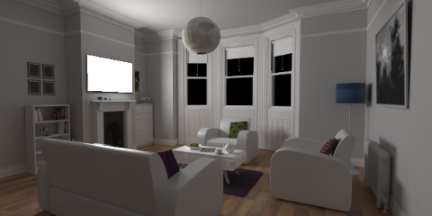} &
\includegraphics[width=.16\textwidth]{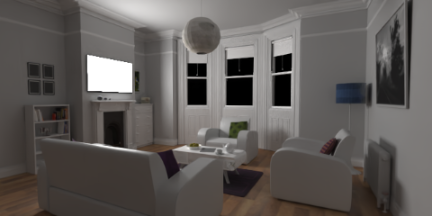} &
\includegraphics[width=.16\textwidth]{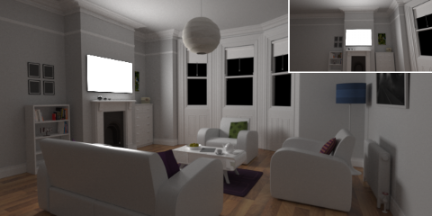} \\ [-0.65ex] 

\includegraphics[width=.16\textwidth]{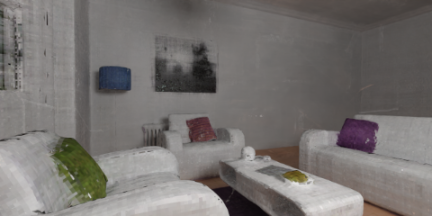} &
\includegraphics[width=.16\textwidth]{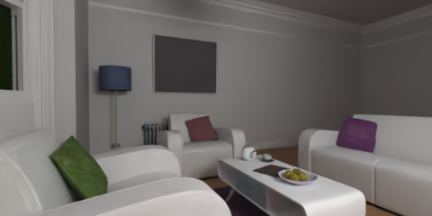} &
\includegraphics[width=.16\textwidth]{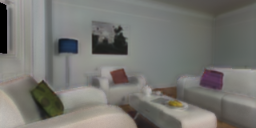} &
\includegraphics[width=.16\textwidth]{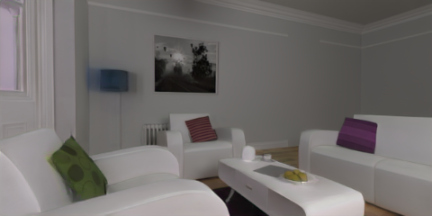} &
\includegraphics[width=.16\textwidth]{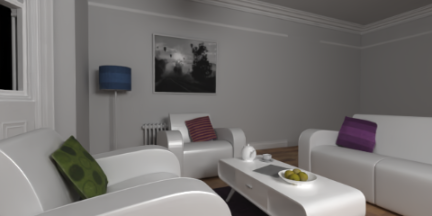} &
\includegraphics[width=.16\textwidth]{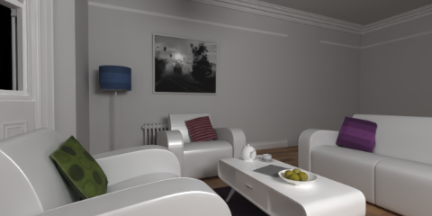} &
\includegraphics[width=.16\textwidth]{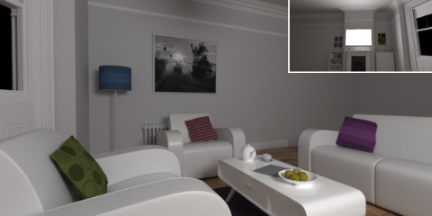}

\end{tabular}
}
\end{center}

\caption{
\textbf{Relighting results on synthetic scenes for all methods}, showing 2 views per-scene. 
}

\label{fig:synthetic_relighting_more}
\end{figure*}

\begin{figure*}[ht!]
    \centering
    \setlength{\tabcolsep}{0.05em}
    \resizebox{0.99\linewidth}{!}{

    \begin{tabular}{ccc @{\hskip 0.04in}|@{\hskip 0.04in} ccc @{\hskip 0.04in}|@{\hskip 0.04in} c}
    \multicolumn{3}{c}{\textsf{\large Conference room}}&
    \multicolumn{3}{c}{\textsf{\large Classroom}}&
    \\
    \milo & \lieccv & \ours~(Ours) & 
    \milo & \lieccv & \ours~(Ours) & Reference\\

    \includegraphics[width=.15\linewidth]{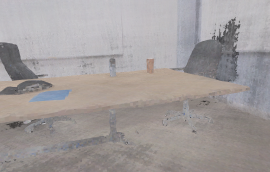}&
    \includegraphics[width=.15\linewidth]{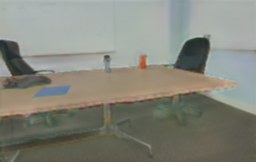}&
    \includegraphics[width=.15\linewidth]{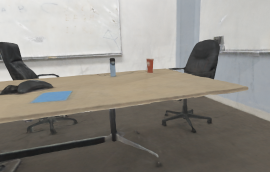}&

    \includegraphics[width=.15\linewidth]{images/real_supp/milo-ClassRoom-1-albedo.png}&
    \includegraphics[width=.15\linewidth]{images/real_supp/li22-ClassRoom-1_albedo.png}&
    \includegraphics[width=.15\linewidth]{images/real_supp/classroom-ours-albedo-1.png}&
    
    \includegraphics[width=0.15\linewidth]{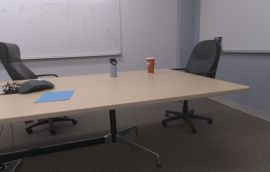}\\[-0.65ex]

    \includegraphics[width=.15\linewidth]{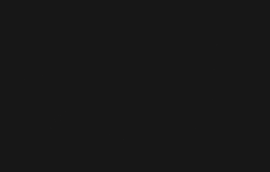}&
    \includegraphics[width=.15\linewidth]{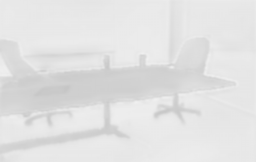}&
    \includegraphics[width=.15\linewidth]{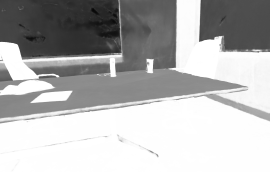}&

    \includegraphics[width=.15\linewidth]{images/real_supp/milo-ClassRoom-1-roughness.png}&
    \includegraphics[width=.15\linewidth]{images/real_supp/li22-ClassRoom-1_roughness.png}&
    \includegraphics[width=.15\linewidth]{images/real_supp/classroom-ours-roughness-1.png}&
    
    \includegraphics[width=0.15\linewidth]{images/real_supp/classroom-reference-0.png}\\[-0.65ex]

    \includegraphics[width=.15\linewidth]{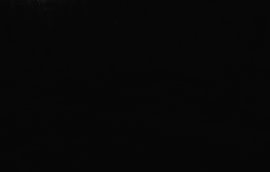}&
    \includegraphics[width=.15\linewidth]{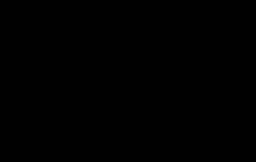}&
    \includegraphics[width=.15\linewidth]{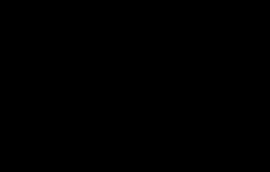}&

    \includegraphics[width=.15\linewidth]{images/real_supp/milo-ClassRoom-1-emission.png}&
    \includegraphics[width=.15\linewidth]{images/real_supp/li22-ClassRoom-1_emission.png}&
    \includegraphics[width=.15\linewidth]{images/real_supp/classroom-ours-emission-1.png}&
    \\
    \end{tabular}
    }
    \caption{
    \textbf{BRDF and emission estimation on real scenes}, showing 1 additional view per-scene besides views shown in the main paper.}
    \label{fig:sup-real}
\end{figure*}

\begin{figure*}[hbt!]
    \centering
    \setlength{\tabcolsep}{0.05em}
    \resizebox{0.7\linewidth}{!}{
    \begin{tabular}{cccc c }
        \multicolumn{5}{c}{\textsf{\large Conference room}} \\
        \milo & \lieccv & \fvp & \ours~(Ours)& Ground Truth\\
        \includegraphics[width=0.15\linewidth]{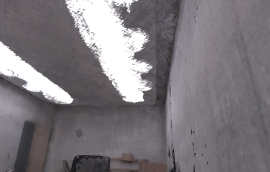}&
         \includegraphics[width=0.15\linewidth]{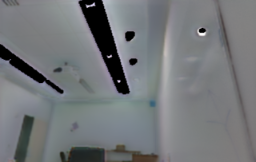}&
         \includegraphics[width=0.15\linewidth]{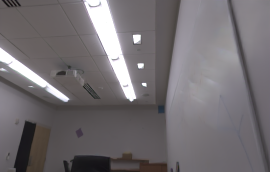}&
         \includegraphics[width=0.15\linewidth]{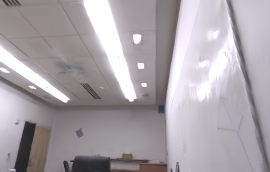}&
         
         \includegraphics[width=0.15\linewidth]{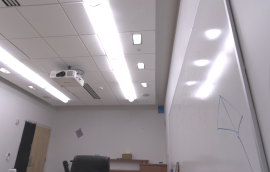} \\

         \multicolumn{5}{c}{\textsf{\large Classroom}} \\
         \includegraphics[width=0.15\linewidth]{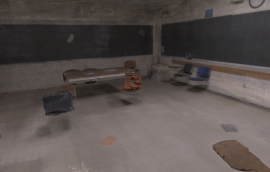}&
         \includegraphics[width=0.15\linewidth]{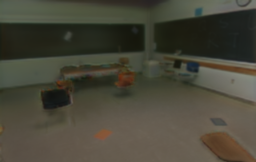}&
         \includegraphics[width=0.15\linewidth]{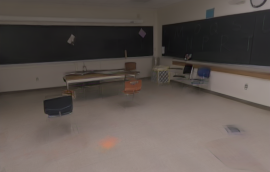}&
         \includegraphics[width=0.15\linewidth]{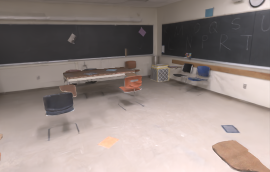}&

        \includegraphics[width=0.15\linewidth]{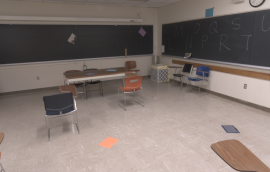}\\
        \multicolumn{5}{c}{\textbf{Rerendering}}\\
        [3ex]
        \multicolumn{5}{c}{\textsf{\large Conference room}} \\
         \milo & \lieccv & \fvp & \ours~(Ours)& Reference\\
         \includegraphics[width=0.15\linewidth]{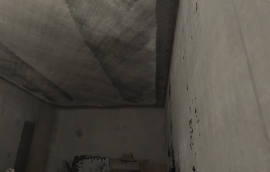}&
         \includegraphics[width=0.15\linewidth]{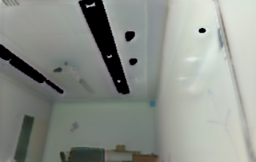}&
         \includegraphics[width=0.15\linewidth]{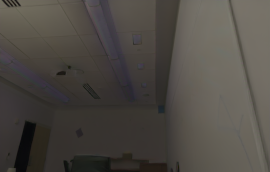}&
         \includegraphics[width=0.15\linewidth]{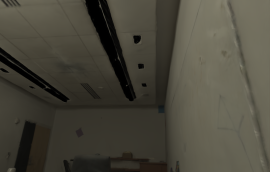}\\
         
         \multicolumn{5}{c}{\textsf{\large Classroom}} \\
         \includegraphics[width=0.15\linewidth]{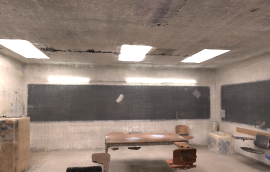}&
         \includegraphics[width=0.15\linewidth]{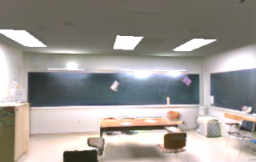}&
         \includegraphics[width=0.15\linewidth]{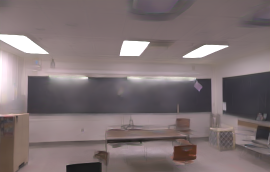}&
         \includegraphics[width=0.15\linewidth]{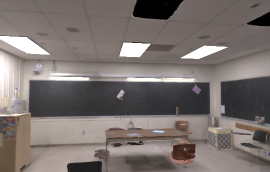}&
         
        \includegraphics[width=.15\linewidth]{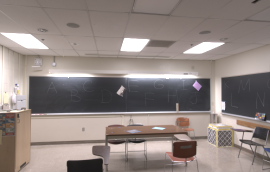}\\
        \multicolumn{5}{c}{\textbf{Relighting}}\\
    \end{tabular}
    }
    \caption{
    \textbf{Rerendering and relighting on real scenes,} showing 1 additional view per-scene for each task besides views shown in the main paper.
    Top two rows show the rerendering with original lighting (Conference room: all ceiling lamps on; Classroom: rear lights on and fronts lights off).
    Bottom two rows show the relighting under novel light with relit Classroom also included as pseudo-ground truth (with rear lights off, and front lights on).
    }
    \label{fig:sup-real-synthesis}
\end{figure*}

In Fig.~\ref{fig:synthetic_BRDF_supp_1},~\ref{fig:synthetic_BRDF_supp_2}, we show the per-scene qualitative comparison of estimated BRDF and emission for all methods on synthetic dataset,
and we compare the view synthesis and relighting results in Fig.~\ref{fig:synthetic_synthesis_more},~\ref{fig:synthetic_relighting_more}.
In Fig.~\ref{fig:sup-real} and Fig.~\ref{fig:sup-real-synthesis},
we provide evaluation on additional views of our real world captures.

\subsection{Evaluation scheme of baseline methods}
\label{supp:baselines}
For FVP~\cite{philip2021free}, We use its original code with the following adaptations: 
FVP relies on thresholding RGB values to locate emitters,
so we pick the threshold that separates emitters from the rest of the scene in our images.
It also involves a step to manually set the exposure of each overexposed emitter,
which in our adaptation is provided as the median radiance within each emitter.
In relighting, FVP assumes the maximum radiance of novel emitters to be 1 so as to yield shading in SDR for input into its network. 
Afterwards, the exposure of novel emitters can be set to arbitrary numbers in FVP's GUI. 
We follow the strategy but set exposure as our desired radiance values for novel emitters, so that relighting results from FVP can be directly comparable.

For evaluation of \ipt~and \milo, since their code is not available, and a re-implementation requires careful design choices, we depend on results provided by the authors of MILO and IPT on our data, where it was possible for them to evaluate. 
Because MILO and FVP use texture-based representations, 
geometry is remeshed to prevent artifacts like UV seam and bleeding, which gives equivalent quality in most of the cases except for thin structures like disks on the kitchen table.

For Li22~\cite{li2022physically}, instead of using predicted depth,
we directly back-project ground truth geometry to obtain a depth image as its input. 
On real scenes, 
considering the method is based on single-view input and does not allow rerendering to novel views under different lighting,
we directly feed the reference relighting image as the input,
replace all estimated emitters by our novel eimtters (Sec.~\ref{subsec:real-world-scene}),
and re-render the scene with estimated materials using its neural rendering pipeline.


\newpage

\end{appendix}
\end{document}